\documentclass{x2lab}

\usepackage{microtype}
\usepackage{graphicx}
\usepackage{subcaption}
\usepackage{tikz}
\usetikzlibrary{arrows.meta,positioning,calc,decorations.pathreplacing,patterns}
\usepackage{csquotes}
\usepackage{afterpage}
\usepackage{placeins}

\usepackage{amsmath}
\usepackage{amssymb}
\usepackage{mathtools}
\usepackage{amsthm}
\usepackage{xspace}
\usepackage{colortbl}
\usepackage{multirow}
\usepackage{enumitem}
\usepackage{wrapfig}
\usepackage{nicefrac}
\usepackage{makecell}

\usepackage{wrapfig}
\usepackage{xcolor}
\usepackage{tikz}
\usetikzlibrary{arrows.meta, decorations.pathreplacing}

\usepackage{pifont}

\definecolor{fbApp}{HTML}{c8e7fa}
\definecolor{fbPurple3}{HTML}{f0ebf5}

\definecolor{citecolor}{HTML}{0071BC}
\definecolor{linkcolor}{HTML}{ED1C24}
\definecolor{efblue}{RGB}{0, 102, 204}

\providecommand{\ours}{\textsc{WALL-WM}\xspace}

\makeatletter
\renewcommand\paragraph{\@startsection{paragraph}{4}{\z@}
  {1.5ex \@plus 1ex \@minus .2ex}
  {-1em}
  {\normalfont\normalsize\bfseries}}
\makeatother

\title{WALL-WM: Carving World Action Modeling\\at the Event Joints}
\author{X Square Robot Team}

\abstract{
WALL-WM is a World Action Model that shifts video-action learning from chunk-centric optimization to event-grounded Vision-Language-Action pretraining, using semantically coherent action events as the atomic unit of learning. Existing WAMs commonly initialize from multimodal or video foundation models and then optimize fixed-length action chunks conditioned directly on the current observation and instruction. Although convenient, this chunk-centric formulation creates a fundamental granularity mismatch. Language describes semantic goals and events, vision evolves through continuous scene dynamics, and actions operate at control-level timescales; forcing all three into the same fixed-length prediction window turns VLA training into short-horizon correlation fitting. This not only underuses the pretrained visual-semantic prior, but can actively overwrite it with chunk-specific action shortcuts, weakening compositionality and long-horizon generalization.
WALL-WM addresses this mismatch by organizing both supervision and data around semantic events. Specifically, it pairs event-grounded VLA pretraining with a data ecosystem built from event-level captions and cluster-balanced sampling, enabling scalable learning over diverse behaviors, scenes, and task structures. From the same event-pretrained backbone, WALL-WM supports two complementary inference modes. The event mode consumes next-event descriptions and enables variable-length execution chunks, while the unified mode uses a VLM with Staircase Decoding to condition conventional fixed-length chunk inference while preserving a gradient-continuous VLA path.
Together with Muon-optimizer-based large-scale pretraining infrastructure, WALL-WM provides a practical scale-up recipe for general-purpose WAMs. Experiments show that WALL-WM generalizes broadly across language, scenes, and tasks, achieving state-of-the-art performance in large-scale real-world generalization evaluation.
\begin{quote}
\centering
\small
\itshape
``Carve nature at its joints.''\\[0.5em]
\normalfont\hfill --- after Plato, \textit{Phaedrus} 265e
\end{quote}
}
  
\date{\today}

\metadata[Code]{\url{https://github.com/X-Square-Robot/wall-wm}}

\begin{document}
\maketitle

\section{Introduction}

Recent progress in embodied foundation models~\cite{zitkovich2023rt,kim2024openvla,black2024pi_0,black2025pi_,bjorck2025gr00t,cheang2025gr,comanici2025gemini} has increasingly been driven by large-scale priors inherited from multimodal understanding models and video foundation models~\cite{du2023learning,jang2025dreamgen,cen2025worldvla,zhu2025unified}. In this report, we use VLA to broadly denote embodied foundation models that predict actions from visual-language inputs, and WAMs to denote world-action models that explicitly couple future observation modeling with action prediction.

Most existing embodied foundation models adapt these priors by predicting fixed-length action chunks from the current observation and language instruction. This chunk-centric formulation is effective and convenient, but it hides a structural mismatch: it cuts embodied dynamics by an external clock, while language, vision, and action evolve at different semantic and temporal scales. Language specifies goals and events; vision evolves through continuous scene dynamics; action operates at control-level time scales and is sensitive to contact, timing, and small perturbations. Thus, adapting pretrained visual-semantic priors to embodied control is not merely a fine-tuning problem. It is first a question of where to place the atomic unit of video-action learning.

This distinction is the starting point of WALL-WM:

\begin{center}
\vspace{-0.25em}
{\small\itshape
Fixed chunks cut by clock;\
semantic events cut by embodied dynamics.
}
\vspace{-0.25em}
\end{center}

\begin{figure}[t]
    \centering
    \begin{center}
    \includegraphics[width=\linewidth]{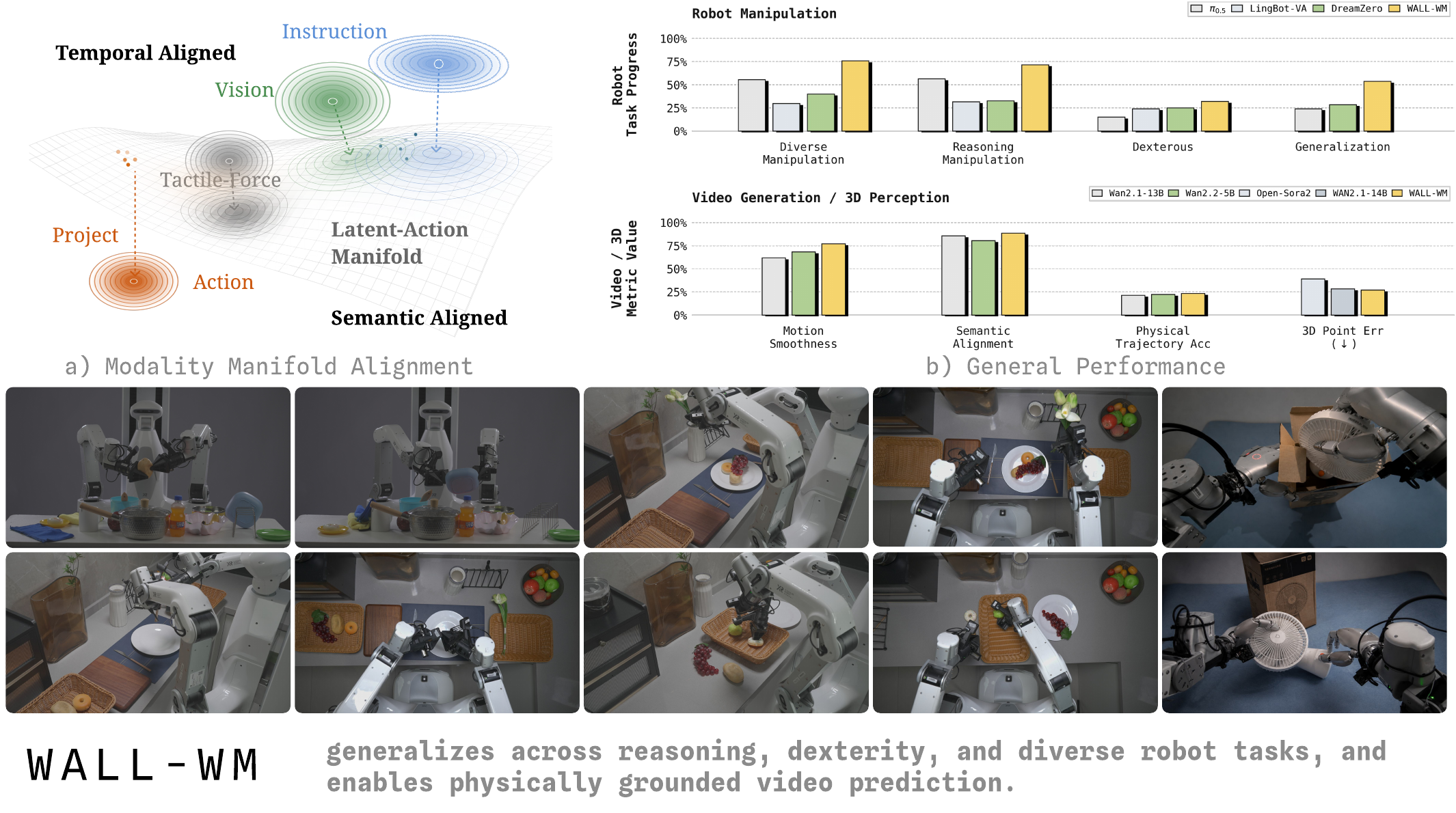}
    \end{center}
    \caption{\textbf{Conceptual illustration of modality hierarchy and WALL-WM's general performance.}
    \emph{Left:} a stylized alignment landscape over semantic abstraction and spatial-temporal precision. Text provides coarse semantic alignment, vision provides denser spatial-temporal grounding, and action requires fine-grained contact-sensitive precision. Tactile-force input, when available, is treated as an optional contact-rich signal rather than a required modality.
    \emph{Right:} WALL-WM shows clear advantages across manipulation-task performance and video-generation metrics, indicating that event-grounded pretraining improves both executable control and future-observation modeling.
    \emph{Bottom:} representative real-robot task snapshots provide illustrative examples of the physical tasks.}
    \label{fig:modality_convergence}
\end{figure}

The rest of the argument follows from this mismatch. The central challenge is not multimodal fusion in the usual sense, but \emph{geometry-preserving alignment}. Text, vision, and action do not share the same notion of neighborhood: semantically similar instructions may induce different visual trajectories, while visually similar states may require incompatible control responses. As illustrated in Fig.\ref{fig:modality_convergence}, text provides coarse semantic alignment, vision supplies denser spatial-temporal grounding, and action requires the finest local precision. A useful WAM must connect these modalities without flattening them into a single undifferentiated embedding space.

Video is the natural scaffold between language and action. Internet-scale video pretraining captures rich visual dynamics that would otherwise have to be learned from embodied interaction alone\cite{jang2025dreamgen,du2023learning}. Video is also semantically structured enough to align with language at event boundaries, yet temporally dense enough to expose the timing, transitions, and state changes needed for action execution. Future video shares causal temporal structure with action, making visual-to-action grounding and inverse dynamics possible~\cite{du2023learning,ko2024learning}. In this sense, video offers an embodiment-light bridge from high-level semantic intent to low-level execution.

Turning this scaffold into a WAM is not a short adaptation stage. It is a \emph{prior-preserving lift}: the model must inherit the semantic and temporal structure learned from large-scale video, while acquiring the controllability, contact sensitivity, and causal grounding required for embodied action. Simply appending an action decoder can attach actions to a visual prior, but it does not determine the unit at which the prior should become executable. Without such a unit, joint optimization can collapse toward the most data-rich modality or overwrite useful visual-semantic structure with short-horizon action correlations.

This lift imposes two requirements. First, a WAM must \emph{preserve the video prior}: video generation models favor semantic invariance, visual plausibility, and temporal smoothness, whereas embodied control requires sensitivity to action-induced divergence and contact transitions~\cite{yuan2026fast,li2026causal}. Second, a WAM must provide \emph{temporal grounding}: language instructions often describe global tasks or semantic events, while observations and actions unfold over many frames and control steps. Fixed-length chunks are poorly matched to both requirements. They can be too short to contain a complete semantic event, yet too long to preserve clean causal separation between context and prediction target.

The same reasoning clarifies what must be lifted from a video prior into a general-purpose WAM:
\begin{enumerate}
    \item[\textbf{T1.}] \textbf{Reasoning:} convert global instructions and task progress into event-structured intent.
    \item[\textbf{T2.}] \textbf{Visual prediction:} preserve the caption-to-video inductive bias while making future observations controllable by executable events.
    \item[\textbf{T3.}] \textbf{Fine manipulation:} expose timing, contact transitions, and local state changes required for action execution.
\end{enumerate}
These layers cannot be obtained by stacking a VLM, a video model, and an action head as independent modules. They require a training regime whose alignment unit is simultaneously meaningful to language, visible in video, and executable through action.

This gives three design principles for the alignment unit:
\begin{itemize}
    \item \textbf{Geometry preservation:} connect language, video, and action without collapsing their native structures into one shared space.
    \item \textbf{Prior preservation:} remain compatible with the caption-to-video structure inherited from video foundation models.
    \item \textbf{Executable causality:} provide a prediction target with clear temporal support, while allowing duration to follow the task rather than a fixed clock.
\end{itemize}

These principles rule out the fixed-length action chunk as the fundamental unit. A chunk is convenient for batching and deployment, but it is not a natural object shared by language, video, and action. It may cut through the middle of a semantic behavior, merge multiple behaviors into one target, or require historical context merely to determine what the chunk is supposed to mean.

WALL-WM therefore replaces the fixed-length chunk with an \emph{action-grounded semantic event}: a temporally coherent segment of executable behavior, such as reaching, grasping, lifting, moving, or placing, that is expressible in language, observable in video, and realizable through action. Unlike a fixed temporal chunk, which follows an external clock, an action-grounded semantic event begins and ends when the underlying executable behavior changes. It therefore satisfies the principles above: language names the event, video grounds its spatial-temporal evolution, and action realizes it through control.

WALL-WM instantiates this principle through event-grounded WAM pretraining. Event captions are paired with corresponding video and action segments, and the model is trained to denoise future video and action over event-aligned intervals. This does not merely use events as auxiliary conditions; it grounds the training problem itself at the event level. The result is a prior-preserving route from video foundation models to executable world-action models.

\begin{figure*}[!htbp]
    \centering
    \includegraphics[width=\linewidth]{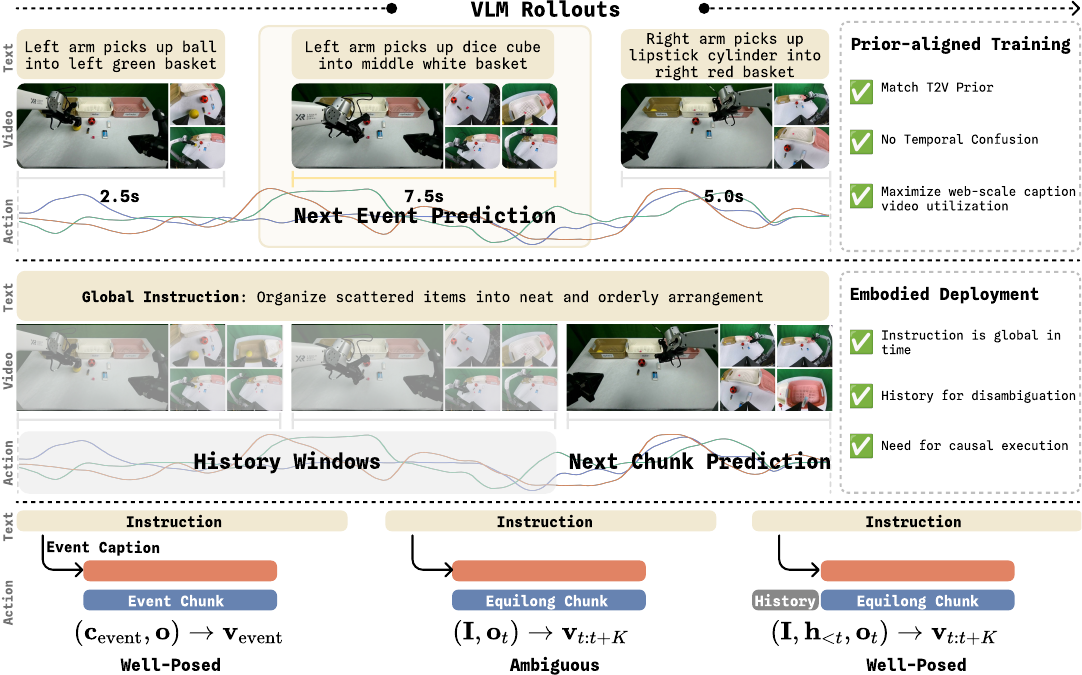}
    \caption{\textbf{Next-event training and equilong-chunk schemes.}
    In prior-aligned training, the event caption, event video, and event action describe the same semantic interval, giving a well-posed caption-to-video/action target.
    In equilong-chunk mode, a global instruction alone is ambiguous for a local chunk; adding history windows restores a well-posed next-chunk prediction problem.}
    \label{fig:vla_alignment}
\end{figure*}

At inference time, the same event-pretrained backbone supports two complementary modes:
\begin{description}
    \item[\textbf{Event mode.}] The system rolls out in event space. A VLM, human, or agent proposes the next-event description, and WALL-WM executes the corresponding variable-length video-action segment before observing the next state. This mode follows the natural duration of the task rather than a fixed control horizon.
    
    \item[\textbf{Unified mode.}] Conventional fixed-length chunk inference is retained, but the chunk is no longer conditioned on a raw global instruction alone. A VLM with Staircase Decoding supplies event-structured latent reasoning over task progress, producing a latent event representation that guides the next local chunk while preserving a gradient-continuous VLA path.
\end{description}
Fig.\ref{fig:vla_alignment} summarizes these two schemes, and Sec.\ref{sec:method} provides the detailed formulation.

This report presents WALL-WM as both a model instance and a training roadmap for prior-preserving scale-up of WAMs. Its main components are:
\begin{itemize}
    \item \textbf{Event-grounded WAM pretraining.} WALL-WM treats action-grounded semantic events as the atomic training unit, pairs them with event captions, and trains a video-action denoiser that preserves the inherited video prior while turning event-aligned visual evidence into executable action.
    
    \item \textbf{Two inference modes from a single event-pretrained backbone.} Event mode performs event-space rollout with variable-length executable segments. Unified mode supports fixed-length chunk prediction by using VLM-based Staircase Decoding to generate event-structured latent reasoning while maintaining a gradient-continuous VLA path.
    
    \item \textbf{Scale-up infrastructure for event-grounded embodied modeling.} WALL-WM combines an event-grounded data ecosystem, cluster-balanced sampling, and Muon-optimizer-based large-scale pretraining infrastructure to support scalable WAM training across diverse behaviors, scenes, and tasks.
\end{itemize}

WALL-WM demonstrates broad generalization across language instructions, scenes, and tasks, achieving strong performance in large-scale real-world generalization evaluation. Overall, WALL-WM connects text, video, and action through action-grounded semantic events. We position WALL-WM less as a short adaptation of a video foundation model and more as a prior-preserving scale-up methodology for the next generation of embodied foundation models.
\section{Related Work}

\subsection{Vision-Language-Action Models}
A growing line of embodied foundation models, commonly referred to as Vision-Language-Action (VLA) policies, extends pretrained vision-language models with action interfaces that map visual observations and natural-language instructions to executable motor commands~\cite{zitkovich2023rt,kim2024openvla,octo2024,black2024pi_0,black2025pi_,bjorck2025gr00t,cheang2025gr,wu2026pragmatic,team2025gigabrain}. By inheriting web-scale semantic priors from VLM pretraining, these models exhibit strong generalization across objects, scenes, and language instructions, and offer a unified alternative to modular perception--planning--control pipelines. Recent work has explored a broad range of design choices within this observation-to-action paradigm, including efficient action tokenization and chunking for long-horizon control~\cite{act,pertsch2025fast,kim2024openvla}, diffusion- or flow-matching-based action experts~\cite{chi2023diffusionpolicy,li2024cogact,wen2025dexvla,black2024pi_0}, lightweight and data-efficient backbones~\cite{wen2025tinyvla}, latent-action pretraining and cross-embodiment routing~\cite{ye2025latent,bu2025univla}, knowledge-insulated fine-tuning that preserves pretrained priors during adaptation~\cite{driess2026knowledge}, spatially enhanced or 3D-aware policy representations~\cite{qu2025spatialvla,nikolov2025spear}, and visual chain-of-thought reasoning interleaved with action generation~\cite{zawalski2024embodiedcot,zhao2025cot}. Despite their broad semantic capabilities, existing VLAs still face systematic limitations. First, their underlying VLMs are pretrained predominantly on static image--text data~\cite{zitkovich2023rt,black2024pi_0}, so even large-scale supervised fine-tuning on teleoperation datasets~\cite{kim2024openvla,bjorck2025gr00t} primarily learns action imitation rather than an explicit model of how the physical world evolves under intervention. Second, most VLAs formulate control as a reactive observation-to-action mapping without action-conditioned future prediction or an explicit temporal-dynamics prior~\cite{li2024cogact,cheang2025gr}. Third, generalization to genuinely novel motions, skills, embodiments, or environments often still requires substantial task- or robot-specific adaptation data~\cite{black2025pi_,comanici2025gemini}. These limitations have motivated a parallel line of research that grounds policies in generative models of how the world evolves, which we review next.

\subsection{Generative Embodied World Models}
Generative world models have recently emerged as a promising paradigm for embodied AI, where agents learn to predict future states of the physical world and use such predictions for planning and control. Early model-based reinforcement learning methods such as PlaNet, Dreamer, and DreamerV3~\cite{hafner2019learning,hafner2019dream,hafner2023mastering} learn compact predictive states for control, while the JEPA family, including V-JEPA and LeWorldModel~\cite{assran2025v,maes2026leworldmodel}, shows that forecasting in feature space can capture meaningful temporal and physical structure without explicit pixel reconstruction. More recent embodied world models further leverage generative video and world-action modeling objectives to learn robot--environment dynamics from large-scale heterogeneous data. These models use predicted future frames or intermediate visual representations as plans, from which executable actions are recovered through inverse-dynamics models~\cite{du2023learning}, intermediate-feature action decoders~\cite{pai2025mimic}, dense correspondence~\cite{ko2024learning}, planning-oriented trajectory decoding~\cite{du2024video,zhou2024robodreamer}, or synthesized demonstrations transferred from human videos and novel scenes~\cite{bharadhwaj2024track2act,bharadhwaj2024gen2act,chen2025moto,jang2025dreamgen,luo2025solving}. Recent work further suggests that video generators encode useful 3D and interaction priors~\cite{vpp2025,kim2026cosmos,zhou2025act2goal,wu2026generation,liang2025video}, motivating their use as embodied dynamics models. Along this direction, LaDi-WM, AdaWorld, Motus, LDA-1B, and MotuBrain~\cite{huang2025ladi,gao2025adaworld,bi2025motus,lyu2026lda,team2026motubrain} explore latent diffusion, structured visual forecasting, latent-action conditioning, and unified video--action modeling, while LingBot-VA~\cite{li2026causal} and related unified denoising architectures~\cite{guo2024prediction,feng2025vidar,zhu2025unified,cen2025worldvla,ye2026world,won2025dual,cen2025rynnvla,zhang2026dreamvla,liao2025genie} jointly model future prediction and action generation. More recent methods such as  Fast-WAM~\cite{yuan2026fast} further improve inference efficiency by avoiding explicit video decoding or future imagination at test time. Overall, these advances indicate that explicit future-state modeling can improve sample efficiency, robustness, and generalization.

\subsection{Latent Reasoning}
A complementary line of work routes chain-of-thought (CoT) reasoning through compact \emph{latent} representations rather than emitting full textual reasoning traces, motivated by inference efficiency and the ability to explore semantic-level reasoning trajectories beyond discrete token sequences~\cite{hao2024coconut,goyal2024think,kang2025ladir,kang2026beyond,lu2026onevl,zhong2026dualcot}. LaDiR~\cite{kang2025ladir} encodes reasoning steps into blocks of continuous ``thought tokens'' via a VAE and refines them with a latent diffusion model; its reinforcement-learning extension~\cite{kang2026beyond} further optimizes latent trajectories to preserve solution diversity. Recent VLA methods further extend this idea to physically grounded latent reasoning: LaST$_0$~\cite{liu2026last} introduces a latent spatio-temporal CoT that captures future visual dynamics, 3D structure, and proprioceptive states for robotic manipulation; LaST-VLA~\cite{luo2026last} distills geometric constraints and world-model foresight into latent spatio-temporal representations for autonomous driving; and LaRA-VLA~\cite{bai2026latent} progressively transfers textual and visual CoT supervision into latent reasoning dynamics for efficient action generation. In contrast, WALL-WM uses a staircase parallel decoder to inject and propagate latent CoT tokens through staggered depths of the VLM backbone. This design preserves as much of the pretrained VLM's hierarchical visual-linguistic priors and causal computation structure as possible, while amortizing the layer cost of latent reasoning.

\section{Architecture Design: Event-Centric World Action Modeling}
\label{sec:method}

\subsection{Overview}
\label{sec:method-overview}

\begin{figure*}[t]
    \centering
    \includegraphics[width=\linewidth]{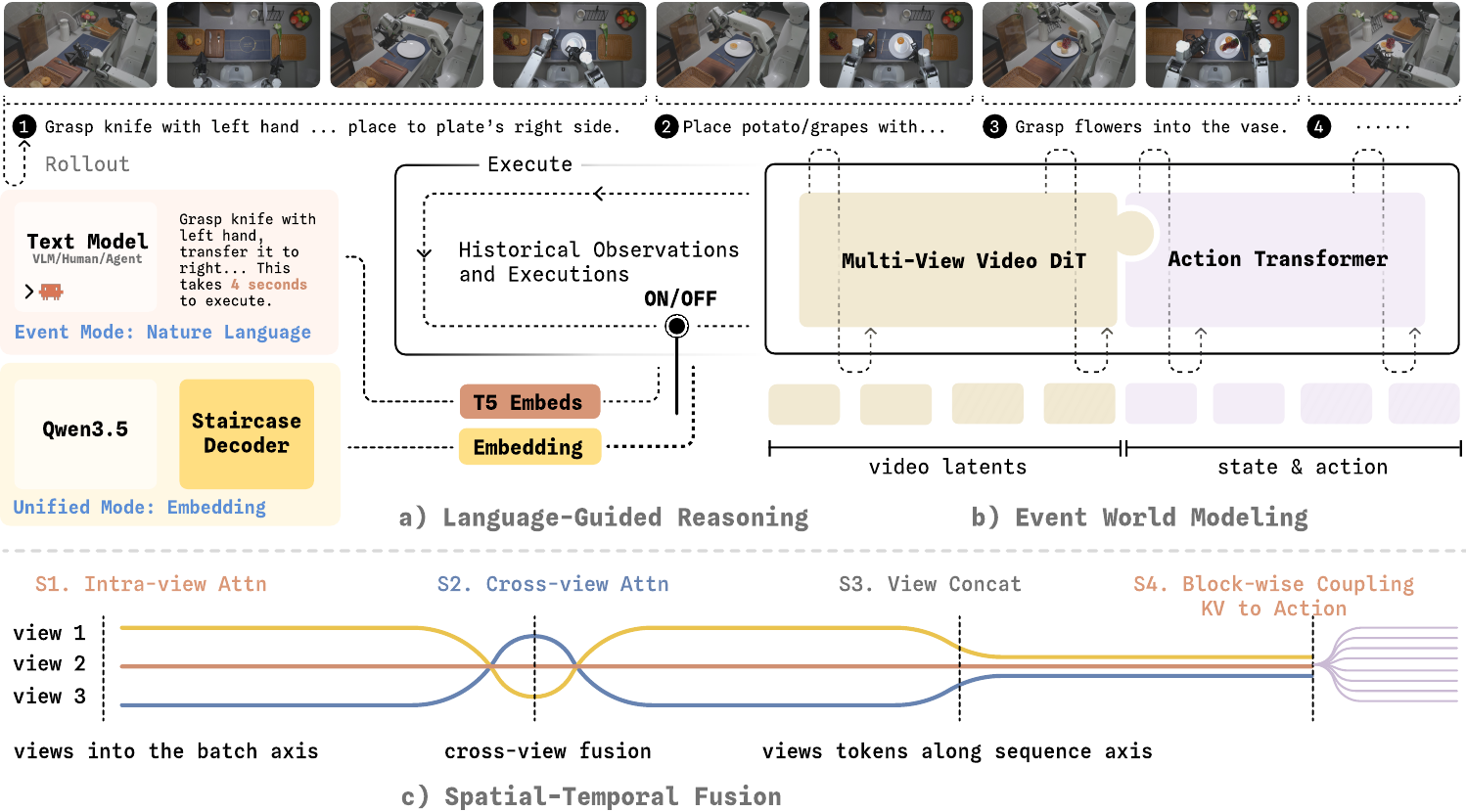} 
    \caption{\textbf{Overall framework of \ours.}
    \ours\ implements \emph{event-centric} world action modeling as a layer-coupled video-action denoiser: given the current multi-view observation and a next-event instruction, it jointly denoises future video latents and the corresponding end-effector trajectory.
    The figure is organized as three subfigures.
    (a) supplies instruction: event mode routes next-event language into the shared text conditioner (T5~\cite{raffel2020exploring} embeds in the figure), whereas unified mode routes through a Staircase decoder to CoT latents.
    (b) is the event-centric world model---a Multi-View Video DiT denoises video latents and an Action Transformer denoises actions; \emph{Execute} and \emph{Rollout} close the control loop above.
    (c) summarizes the spatial-temporal fusion that threads $N_v{=}3$ camera streams through the stack: view folding that preserves the inherited within-view self-attention with its 3D RoPE positional encoding (S1), cross-view mixing in each DiT block (S2), token-axis \texttt{ViewConcat} (S3), and one-way block-wise coupling of fused video keys/values into the action tower at every depth (S4).
    }
    \label{fig:framework}
\end{figure*}

As summarized in Fig.~\ref{fig:framework}, \ours\ instantiates \emph{event-centric} world action modeling through a prior-aligned, multimodal pretraining stack: a video tower inherited from a Wan Series text-to-video model~\cite{wan2025wan} and a randomly initialized action DiT are layer-coupled, with pretrained encoders held fixed and cross-modal alignment learned through layer-wise video-action couplings.
Pretraining is organized at the \emph{event} level: each sample is an atomic $(\mathbf V_e,\mathbf a_e)$ event carved out of a long-horizon episode rather than a fixed-length chunk, and the model is trained to denoise event-aligned futures from the current observation.
At each block, the action stream cross-attends to the matched video features without modifying the video stream.

Formally, \ours\ models
$p_\theta(\mathbf V_e,\mathbf a_e\mid \mathbf V_0,\mathbf s,c_e)$,
where $\mathbf V_0$ is the current multi-view observation (one keyframe per camera), $\mathbf s$ the current proprioceptive state, $(\mathbf V_e,\mathbf a_e)$ the event-aligned future multi-view video and end-effector trajectory (both with event-dependent length), and $c_e$ the per-event caption that describes the same action-grounded semantic event.
The remainder of \Cref{sec:method} follows the two paradigms of Fig.~\ref{fig:vla_alignment}: \emph{next-event prediction} with the \emph{event-centric} layout in subfigure~(A) for pretraining and event-mode inference, and \emph{next-chunk prediction} with the \emph{observation-centered} layout in subfigure~(B) for unified-mode deployment.
First, \Cref{sec:method-video,sec:method-action} build the shared video-action denoiser from event-centric pretraining through layer-wise coupling and temporal alignment, extending to the history-augmented observation-centered window.
Second, \Cref{sec:method-deployment} attaches inference-time language pathways to the same event-pretrained backbone: a vision-aware VLM bridge supports next-event conditioning in event mode and instruction grounding in unified mode, while Staircase decoding supplies event structure for fixed-length chunk prediction without token-by-token autoregression.

\subsection{Multi-View Visual World Events Modeling}
\label{sec:method-video}

The video tower inherits the Wan single-view DiT and extends it to \emph{multi-view, multi-embodiment} video generation.
The inherited within-view computation stays in place; we graft three additions onto the same backbone:
(i) \emph{multi-view adaptation from single-view priors}, which runs rearranged cross-view self-attention over per-frame multi-view tokens and merges the result back through a zero-initialized output projector;
(ii) \emph{Camera RoPE}, which gives each camera a learnable, calibration-free identity in that cross-view branch so the same DiT can operate across heterogeneous multi-embodiment camera setups;
and (iii) \emph{cross-view geometric masking}, a complementary training-time pair that strengthens cross-view geometric consistency.
The tower is then trained on event latents under Wan-style $v$-prediction flow matching~\cite{ho2020ddpm,salimans2022progressive,lipman2022flow}.

\paragraph{Multi-View Adaptation from Single-View Priors.}
When $N_v{>}1$, each DiT block runs an additional cross-view branch after the usual within-view Wan self-attention.
Each camera stream still enters as its own batch item; inside a view-attention block we regroup hidden states frame by frame, concatenate all spatial tokens from the $N_v$ cameras at each latent frame into one sequence, and run self-attention on that joint layout---with rotary codes that also encode the camera axis (Camera RoPE, next paragraph).
The attention weights are initialized from the block's within-view self-attention; its output is passed through a zero-initialized projector and added back to the per-view stream under an AdaLN gate.
Let $\mathbf h_i^V$ denote the hidden states at DiT block~$i$ (in the per-view Wan layout), $\mathrm{CrossViewAttn}_i(\cdot)$ the rearranged cross-view self-attention above, $W_\text{view}$ its zero-initialized output projector, and $g_i$ the AdaLN gate on that branch; then
\begin{equation}
  \mathbf h_i^V \;\leftarrow\; \mathbf h_i^V + g_i\,W_\text{view}\,\mathrm{CrossViewAttn}_i\!\bigl(\mathbf h_i^V\bigr),
  \qquad
  W_\text{view} \;\text{initialized to } 0.
\end{equation}
Because $W_\text{view}$ starts at zero, this branch contributes nothing at initialization and cross-view exchange turns on only as the projector learns during training.
The within-view Wan stack is otherwise unchanged, so pretrained appearance and language-alignment behavior are preserved while cross-view exchange is learned on top.

\paragraph{Camera RoPE.}
To support large-scale multi-embodiment training, Camera RoPE gives each view a learnable rotary identity without feeding calibration to the model at runtime.
We extend RoPE~\cite{su2024roformer} with a view axis, partitioning each head's frequency bank over $(f,h,w,\text{view})$; the view rotation is produced from a learnable per-view embedding shared across all view-attention layers.
Adding or removing a camera therefore only changes the embedding table.
The rotary code tells the network which camera each token came from; during training, the sight-cone mask below further specifies which other tokens it may plausibly correlate with.

\paragraph{Cross-View Geometric Masking.}
View attention alone can mix tokens across cameras even when their patches have no shared field of view, and when co-visible regions \emph{do} exist the network often recovers them through the shorter temporal path inside a single view, so cross-view attention is either abused as a generic feature mixer or left under-trained on the pairs that matter.
At training time, we address both failure modes with a complementary pair of geometry-aware masks built from the same per-robot calibration; both are dropped at inference so rollout stays calibration-free.

\emph{Sight-cone attention masking.}
For two video tokens $u{=}(v_u, h_u, w_u)$ and $u'{=}(v_{u'}, h_{u'}, w_{u'})$ on the same latent frame, we say the pair is \emph{co-visible} if the viewing frustums of their related patches in the original videos intersect (there could be observations of the same region from different viewing angles in pixel space).
For computational convenience, we model each frustum as a cone $C(u)=(\mathbf p_0(u), \hat {\mathbf v}(u), \gamma(u))$ with apex at the camera center, axis $\hat {\mathbf v}$ toward the patch center, and half-apex angle $\gamma$ initialized to tightly enclose the patch (optionally scaled by $l{\ge}1$).
Cone parameters follow per-robot extrinsics, intrinsics, and distortion; intersections are tested in parallel within a depth-of-field band $[d_{\text{min}},d_{\text{max}}]$:
\begin{align}
  \mathbf p(u, t) &= \mathbf p_0(u)+t\hat{\mathbf v}(u),\\
  (t_1, t_2) &= \mathop{\arg\min}_{(t_1,t_2)}\left\Vert\mathbf p(u,t_1)-\mathbf p(u',t_2)\right\Vert_2,\\
  \hat t_1&=\text{clamp}\left(t_1,d_{\text{min}},d_{\text{max}}\right),~~\hat t_2=\text{clamp}\left(\mathop{\arg\min}_{t_2}\left\Vert\mathbf p(u,\hat t_1)-\mathbf p(u',t_2)\right\Vert_2, d_{\text{min}}, d_{\text{max}}\right),\\
  &C(u)\text{ intersects } C(u')
  \;\;\Longleftrightarrow\;\;\left\Vert \mathbf p(u, \hat t_1) -\mathbf p(u',\hat t_2)\right\Vert_2\le \hat t_1\gamma(u) + \hat t_2\gamma(u')\qquad(\text{approximately}).\label{eqn:conesection}
\end{align}
This yields a binary mask
\begin{equation}
  \mathcal{M}_\text{sc}[u, u']\;=\;1
  \;\;\Longleftrightarrow\;\;
  C(u)\text{ intersects } C(u').
\end{equation}
We add $(1-\mathcal{M}_\text{sc})\cdot(-\infty)$ as an attention bias in every view-attention block, forbidding cross-view routing across geometrically incompatible patches; the mask is computed once per sample and reused across depth.
At inference the bias is dropped, recovering unmasked attention with Camera RoPE.
$\mathcal{M}_\text{sc}$ closes the loop with Camera RoPE above: rotary codes identify \emph{which} camera each token came from, and the sight-cone mask identifies \emph{which other tokens} it may correlate with.

\emph{Tube patch masking.}
The second mechanism creates an explicit \emph{demand} for cross-view attention.
With probability $p_\text{tube}$ we pick a view $v^*$ and a $k{\times}k$ spatial window with $k{\in}\{l_{\text{min}},\ldots,l_{\max}\}$, uniformly sampled inside $U{=}\{u\mid\exists u', v_{u'}{\neq}v^*, \mathcal{M}_\text{sc}[u,u']{=}1\}$ so recovery from other views is possible.
The resulting \emph{tube}---the same spatial window across all latent frames of $v^*$---is masked in the noised input $\mathbf z_t^V$ by replacing the tokens with pure noise; with nested probability $p_\text{tube}^\text{cond}$ the same tube is also masked on the conditioning channel $\mathbf y$.
The reconstruction target is unchanged, but the masked tube has no within-view temporal shortcut, so recovery must route through the other $N_v{-}1$ views.
Non-trivial $(p_\text{tube},p_\text{tube}^\text{cond})$ are used when hardening cross-view correspondence; how the masked tube enters the video loss is specified in the flow-matching paragraph below.

\begin{figure*}[!htbp]
    \centering
    \includegraphics[width=0.9\linewidth]{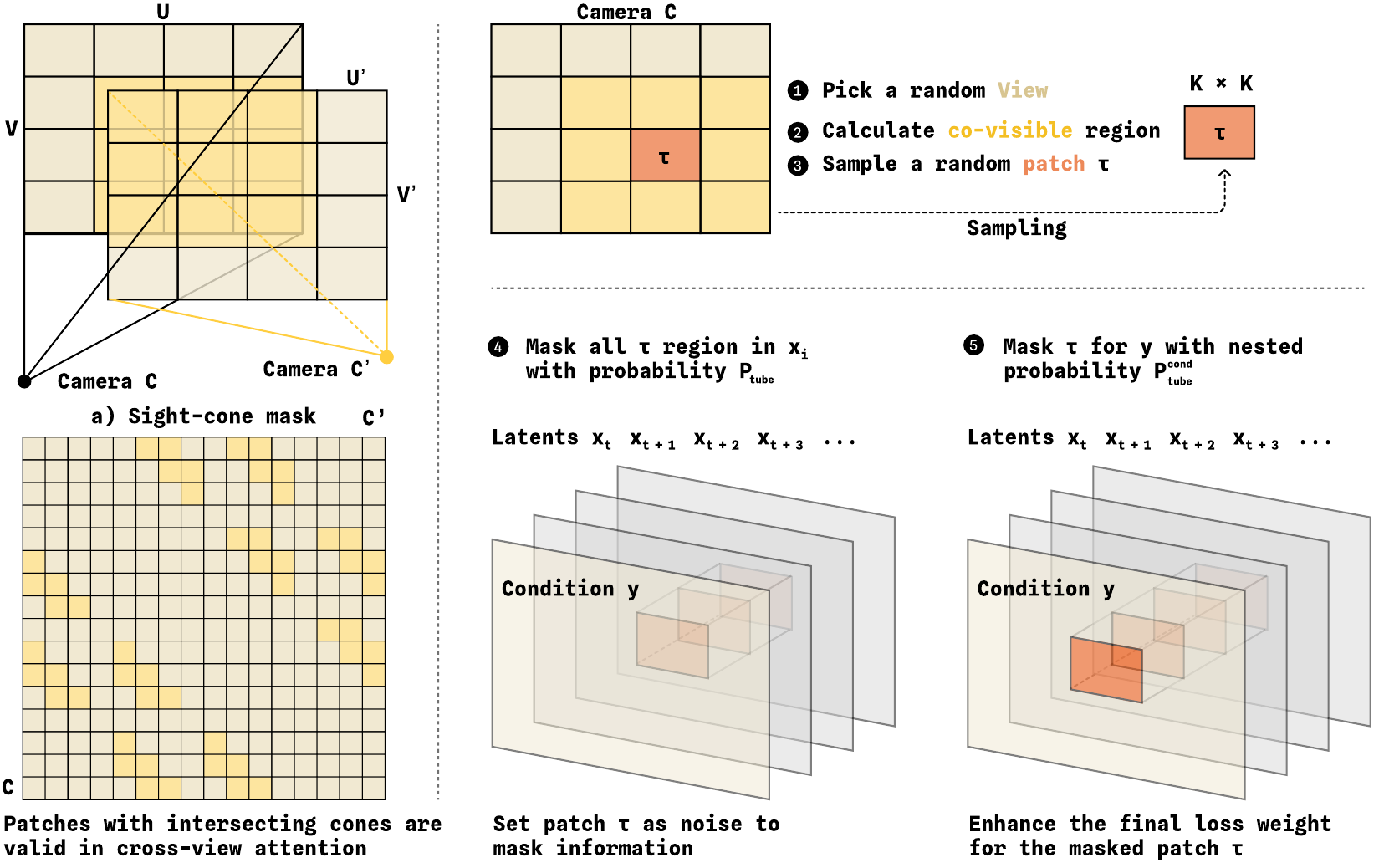}
    \caption{\textbf{Cross-view masking in \ours's view attention.}
    \emph{(a) Sight-cone mask.}
    A token pair $(u, u')$ is allowed to attend if and only if their back-projected sight cones share a 3D region in front of both cameras; the resulting binary mask $\mathcal{M}_\text{sc}$ is added as a $(1-\mathcal{M}_\text{sc})(-\infty)$ attention bias, with intra-view pairs always allowed (matrix diagonal).
    \emph{(b) Tube mask.}
    A spatial window on one view $v^{*}$ is masked across all latent frames---and, with nested probability, also on the InP conditioning channel $\mathbf{y}$---leaving cross-view attention to the other views as the only path that can recover the masked content; the in-tube $v$-prediction loss is up-weighted as in the flow-matching objective below.
    Both masks are training-only, so the runtime path stays calibration-free.}
    \label{fig:cross-view-masking}
\end{figure*}

\emph{Why the two are complementary.}
Sight-cone masking acts on attention \emph{topology}: it makes the cross-view graph reflect physical visibility, but does not push traffic along the remaining edges.
Tube patch masking acts on input \emph{content}: it removes information recoverable through temporal self-attention, but does not block geometrically nonsensical correlations.
Used together at training time, the model may attend across views only where geometry permits and is required to do so to minimize the video objective below (\Cref{fig:cross-view-masking}).

\paragraph{Video Flow-Matching Objective.}
The video tower inherits Wan-style flow matching and is trained on event latents before the action tower is attached (\Cref{sec:training-va-pretrain}).
Given clean latents $\mathbf z_0^V$, noise $\boldsymbol\varepsilon^V$, and a sampled video timestep $t^V\sim\Phi_V$, we form noisy latents $\mathbf z_{t^V}^V$ and regress the flow target $\mathbf C^{V\star}=\boldsymbol\varepsilon^V-\mathbf z_0^V$.
All video-side supervision is collected in a single flow-matching loss.
By default,
\begin{equation}
  \mathcal L_V = w_V(t^V)\,\bigl\lVert \hat{\mathbf C}^V - \mathbf C^{V\star}\bigr\rVert^2,
  \label{eq:video-flow-loss}
\end{equation}
with optional per-timestep weighting $w_V(t^V)$, evaluated only on \emph{valid} spatial tokens.
Out-of-frame and synthetic border regions are always excluded from this MSE---we refer to this as \emph{border masking} in the loss; it is independent of the sight-cone and tube mechanisms above and is applied in all video runs (\Cref{sec:training-va-pretrain}).
When tube-patch masking is active ($p_\text{tube}{>}0$), denote the masked spatio-temporal tube by $\mathcal{T}$.
Tokens inside $\mathcal{T}$ use the same $\mathbf C$ target but enter with an additional loss weight $\lambda_\text{mask}$:
\begin{equation}
  \mathcal L_V
  \;=\;
  w_V(t^V)\!\left(
  \sum_{u\notin\mathcal{T}} \bigl\lVert \hat C^V_u - C^{V\star}_u \bigr\rVert^2
  \;+\;
  \lambda_\text{mask}\!\sum_{u\in\mathcal{T}} \bigl\lVert \hat C^V_u - C^{V\star}_u \bigr\rVert^2
  \right),
  \label{eq:video-flow-loss-tube}
\end{equation}
where $C^{V\star}_u$ is the $u$-th component of $\mathbf C^{V\star}$.
Setting $\lambda_\text{mask}{=}1$ or leaving $\mathcal{T}{=}\varnothing$ recovers the uniform form of \Cref{eq:video-flow-loss}.
Sight-cone masking affects only cross-view attention and never modifies which tokens contribute to $\mathcal L_V$.
Our prior-preserving main recipe always keeps border masking and sight-cone attention supervision, disables tube sampling and tube up-weighting ($p_\text{tube}{=}0$ or $\lambda_\text{mask}{=}1$ with empty $\mathcal{T}$); non-trivial tube settings are reserved for runs that harden cross-view correspondence.
\subsection{Event-Centric Action Dynamics Modeling}
\label{sec:method-action}

The action tower is an action DiT with the same depth as the video tower.
At each layer, action tokens read multi-view video features from the paired video block; the stack denoises the end-effector trajectory with flow matching.
Below we cover layer-wise coupling, video-action temporal alignment, video-action denoising-step mapping, and the action training objective.

\paragraph{Action Transformer and Layer-Wise Coupling.}
A shared MLP encoder $\mathcal{E}_a:\mathbb{R}^{D_a}\to\mathbb{R}^d$ embeds both the proprioceptive state and the noisy actions into the transformer width; the resulting $T_s{+}T_a$ tokens ($T_s$ state tokens and $T_a$ action tokens) are processed by a stack of $N_\text{DiT}$ action blocks, each performing (a) self-attention over the action tokens, (b) a dedicated cross-attention to the state token alone, (c) the cross-attention to the matched video-DiT layer's features, and (d) a gated FFN; an MLP head decodes the action tokens back to $\mathbb{R}^{D_a}$.
The state token's separate cross-attention---in addition to participating in the self-attention---keeps absolute proprioception directly reachable from the action tokens at every depth, rather than diluted by the long video key/value sequence.
At step~(c), action block~$i$ cross-attends to video context from the matched video block:
\begin{equation}
  \tilde{\mathbf h}_i^V
  \;=\;\mathrm{ViewConcat}\!\bigl(\mathbf h_{\pi(i)}^V\bigr)
       +E_\tau(\boldsymbol\tau^V)
       +E_{\text{abs}}(\boldsymbol t_{\text{abs}}),
  \qquad
  \pi(i)\;\text{indexes the paired video block},
\end{equation}
Here $\mathbf h_{\pi(i)}^V$ denotes the hidden states leaving video DiT block~$\pi(i)$ as $N_v$ separate per-camera token sequences---views not yet merged along the sequence axis. The depth map $\pi$ pairs action block~$i$ with video block~$\pi(i)$, so block~$i$ cross-attends to features from that video layer; with equal-depth towers we use $\pi(i){=}i$.
$\mathrm{ViewConcat}(\cdot)$ gathers the $N_v$ per-view token sequences at that layer---each spanning all latent frames and spatial patches---and concatenates them along the sequence axis; adding $E_\tau$ and $E_{\text{abs}}$ yields $\tilde{\mathbf h}_i^V$, which serves as the cross-attention keys and values.
Coupling runs in one direction only: each action block cross-attends to per-layer video features, and no auxiliary branch is added to the video DiT.

\paragraph{Video-Action Temporal Alignment.}
Cross-attention requires video keys/values and action queries to share a temporal index.
Action self-attention orders the state-action sequence with 1D RoPE; at the video interface we instead add two learnable lookups, $E_\tau$ and $E_{\text{abs}}$, to action queries and to the assembled video keys/values.
$E_\tau$ indexes tokens \emph{inside} the current window; $E_{\text{abs}}$ indexes \emph{which} window is active when many fixed-length chunks roll out under one global instruction.
We first specify the two window layouts and their positional codes; we then show how the observation-centered layout is instantiated in raw frames, VAE latents, and action tokens without a separate history encoder.

\emph{Event-centric window.}
Used in event-centric pretraining (\Cref{sec:training-va-pretrain}), where a per-event caption already localizes each segment.
We disable $E_{\text{abs}}$ and assign every token an integer frame index $\tau$ within the window:
\begin{equation}
  \tau^V_{(f,h,w)} \;=\; f,
  \qquad
  \tau^A_0 \;=\; 0,
  \qquad
  \tau^A_{1+k} \;=\; \lfloor k/K_p\rfloor + 1
  \quad\text{for}\;k\in[0, T_a),
\end{equation}
where each latent frame pools $K_p$ action steps.
The state token and zeroth latent frame share $\tau{=}0$; action tokens partition into $T_a/K_p$ groups aligned with successive future latents; and all spatial tokens in latent frame $f$ share $\tau{=}f$.
The shared lookup $E_\tau(\tau)$, added on both sides of cross-attention, biases each action group toward the matching latent frame.

\begin{wrapfigure}{r}{0.45\columnwidth}
    \vspace{-2ex}
    \centering
    \includegraphics[width=1\linewidth]{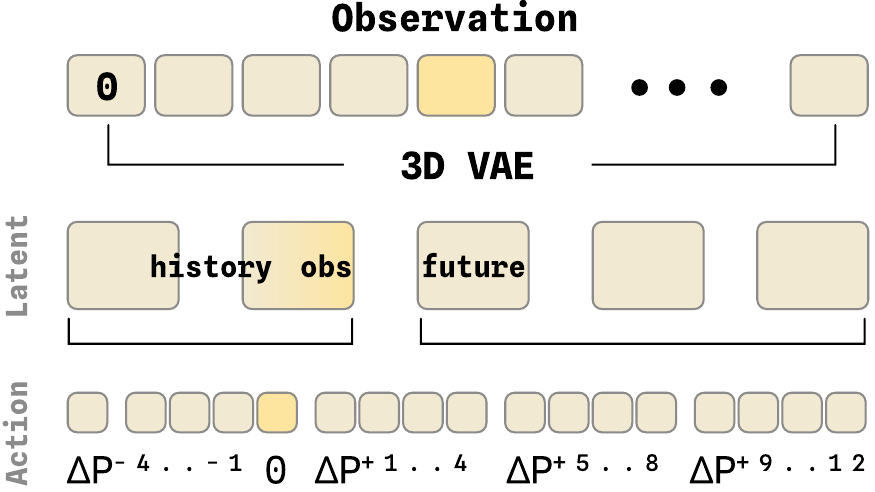}
    \label{fig:history-frame-design}
    \caption{\textbf{Observation-centered window ($M{=}1$).}
    One 3D VAE pass encodes a $1{+}4M{+}4N$ raw buffer into $1{+}M{+}N$ latents.
    Each latent carries $K_p$ relative-pose action tokens ($\mathbf{0}$ at the anchor)}
    \vspace{-2ex}
\end{wrapfigure}

\emph{Observation-centered window.}
Unified deployment slides fixed-length windows under a global instruction; $\tau$ alone cannot distinguish chunks.
We extend the window to $M$ history frames, one observation anchor, and $N$ future frames, and activate both embeddings on both sides of cross-attention:
\begin{equation}
  \mathbf{h} \;\mathrel{+}=\; E_\tau(\tau) + E_{\text{abs}}(t_{\text{abs}}).
\end{equation}
Here $t_{\text{abs}}$ indexes the sliding window; $E_{\text{abs}}$ marks which chunk is current, while $E_\tau$ spans history indices $-M,\ldots,-1$, the anchor at $0$, and future indices $+1,\ldots,+N$.
\Cref{fig:history-frame-design} summarizes this layout with $M{=}1$; below we spell out how one shared codec window materializes it on the video and action sides.

\textbf{\emph{VAE-aligned video stream.}}
The 3D VAE is a $1{+}4N$ temporal codec: a keyframe followed by $4N$ raw frames compresses to $1{+}N$ latents under a leading-one/trailing-four rule.
In event-centric pretraining the current frame occupies the keyframe slot and the chunk-DiT denoises the $N$ trailing latents.
For the observation-centered window we prefix $4M$ history frames and encode the full $1{+}4M{+}4N$ buffer in one pass; the observation keyframe separates prefixed history from the subsequent $4N$ raw frames.
Under the unchanged codec rule the enlarged window yields $1{+}M{+}N$ latents in temporal order---earliest history to the leading latent, the next four frames to the anchor latent, and each following group of four raw frames to a future latent---with no re-encoding seam between history and future; the chunk-DiT denoises only the $N$ trailing future latents.

\textbf{\emph{Relative-pose action stream.}}
The action side mirrors the $1{+}M{+}N$ latent stream: the tower assigns $K_p$ tokens to each video latent.
All non-anchor tokens encode end-effector displacement relative to the observation pose---history tokens back to the anchor, future tokens forward from it; anchor tokens are $\mathbf{0}$, fixing the action time origin.
Embodiment-absolute proprioception stays in the dedicated state token of the layer-wise coupling paragraph rather than this stream.

\paragraph{Video-Action Timestep Mapping.}
Under flow matching, video and action follow separate denoising schedules, yet each action block cross-attends to video features from some denoising step; training and inference must therefore specify how action step index $j$ maps to a video step index.
We write $N_d$ for the number of action denoising steps on the shared inference schedule (e.g.\ $N_d{=}50$) and a mapping $m(j)$ from each action step to the video step whose cross-attention key/value it reads.
Two regimes cover our recipes:
\begin{itemize}
  \item \textbf{Symmetric 1-to-1} (validation): action step $j$ is paired with video step $j$ ($m(j){=}j$; equivalently $t^A{=}t^V$ at each paired step).
  We use this when both towers are updated end-to-end on small data, denoising the two streams jointly for tight fitting.
  \item \textbf{Asymmetric 1-to-$N_d$} (default; large-scale training): a moderate-noise \emph{schedule anchor} $s^\star$ is fixed and broadcast to the full action schedule ($m(j){=}s^\star$ for all $j{\in}\{0,\dots,N_d{-}1\}$).
  At inference, each denoising pass runs video only through $s^\star$ while action completes all $N_d$ steps; every action step reads cross-attention key/value from that single anchored video forward ($m(j){=}s^\star$; \Cref{sec:training-inference}).
\end{itemize}

In our main recipe, the video stream is frozen while the action tower is trained at scale under asymmetric 1-to-$N_d$ mapping.
With the video stream frozen, each optimizer step uses only a single anchored video forward as cross-attention evidence while action supervision comes from ground truth, not from a jointly denoised video target.
A symmetric 1-to-1 pairing ($t^A{=}t^V$) is therefore a poor fit: high-noise video features need not match the ground-truth segment, whereas near-clean features retain too little structure to guide control.
We pin $s^\star$ at the step that best balances faithful visual structure with usable cross-attention evidence (see \Cref{sec:training-va-pretrain}) and train each action forward at an independently sampled noise level $t^A{\sim}\Phi_A$ while it cross-attends to that same anchor, so the action tower learns across the full action schedule without sampling a mismatched video step.
Purely as a training-throughput trick unrelated to inference, we optionally draw $K{>}1$ independent action noise levels per optimizer step while reusing the same anchored video forward; $K$ amortizes compute and plays no role at rollout.
Symbolically,
\begin{equation}
  m(j) \;=\; s^\star \;\;\text{for all}\;\; j{\in}\{0,\dots,N_d{-}1\},
  \qquad
  t^A_k \stackrel{\text{u.a.r.}}{\sim} \Phi_A,\;\;k=1,\dots,K,
\end{equation}
with $\Phi_A$ free in support and density; per optimizer step the cost is one video forward plus $K$ action forwards when the throughput trick is enabled ($K{=}1$ otherwise).

\paragraph{Action Objective.}
During action pretraining only the action tower is optimized under flow matching while the video tower remains fixed.
Our main runs use \textbf{$v$-prediction}, parallel to the video-side target in \Cref{sec:method-video}.
Under the asymmetric 1-to-$N_d$ mapping above, each draw $k$ uses an independently sampled action noise level $t_k^A{\sim}\Phi_A$ while reusing the same anchored video forward; the primary loss is the mean flow-matching MSE over these draws,
\begin{equation}
  \mathcal L_A
  \;=\;
  \frac{1}{K}\sum_{k=1}^{K} \mathcal L_A^k,
  \qquad
  \mathcal L_A^k
  \;=\;
  w(t_k^A)\,\bigl\lVert \hat{\mathbf y}^A_k - \mathbf y^{A\star}_k\bigr\rVert^2,
\end{equation}
where $w(t_k^A)$ optionally reweights by noise level as in \Cref{sec:method-video}.
By default, $\hat{\mathbf y}^A_k$ is the predicted velocity $\hat{\mathbf v}^A_k$ and the target is $\mathbf v^{A\star}=\boldsymbol\varepsilon^A-\mathbf a_0$.
With optional \textbf{$x$-prediction}, the network outputs clean actions instead ($\hat{\mathbf y}^A_k{=}\hat{\mathbf a}_0$, $\mathbf y^{A\star}_k{=}\mathbf a_0$); the weighted MSE above is unchanged.
On contact-heavy data (fine-grained dual-arm manipulation, contact-rich insertion, recovery/re-grasp episodes), contact events span only a few frames.
$v$-prediction can weight those frames too lightly, especially at high noise, so we expose $x$-prediction as an optional mode on such runs.
We can also add a Type-II \textbf{DCT} auxiliary on the recovered trajectory $\hat{\mathbf a}_0^{(k)}$ to emphasize overall motion shape over frame-to-frame jitter (exponential bin decay), down-weighted at high noise by $(1-t_k^A/T)^2$, where $T$ is the maximum action diffusion timestep (the end of the noise schedule):

\begin{equation}
  \mathcal L_A
  \;\mathrel{+}=\;
  \frac{w_\text{DCT}}{K}\sum_{k=1}^{K} \mathcal L_\text{DCT}^k,
  \qquad
  \mathcal L_\text{DCT}^k
  \;=\;
  (1-t_k^A/T)^2 \!\sum_{j=0}^{T_a-1} e^{-j/(T_a/4)}\,
  \bigl\lVert \mathrm{DCT}(\hat{\mathbf a}_0^{(k)})_j - \mathrm{DCT}(\mathbf a_0)_j \bigr\rVert^2,
\end{equation}
with $\hat{\mathbf a}_0^{(k)}$ recovered from draw~$k$.
Our main runs use $v$-prediction with $w_\text{DCT}{=}0$, leaving both $x$-prediction and the DCT auxiliary disabled unless otherwise specified.

\subsection{Language-Guided Reasoning}
\label{sec:method-deployment}
\label{sec:method-qwen}
\label{sec:method-text}
\label{sec:staircase}

Long-horizon embodied manipulation requires reasoning capabilities beyond static language conditioning, including scene grounding, event decomposition, temporal progress estimation, and rollout-time replanning. Our reasoning module is constructed on top of a Qwen3.5-9B backbone~\cite{qwen25,qwen35} and is accelerated via a compact sequence of continuous latent reasoning states (Staircase decoding).

Given the current multi-view observation $\{\mathbf V_0^{(v)}\}_v$ and language instruction $\ell$, the VLM produces hidden states $\mathbf H_q$. The text tokens are extracted using a boundary mask $\mathbf m_{\texttt{txt}}$ and projected into the DiT conditioning space:
\begin{equation}
\mathbf c_\ell
=
\mathcal M_T
\bigl(
\mathcal P_Q(
\mathbf H_q[\mathbf m_{\texttt{txt}}]
)
\bigr),
\end{equation}
where $\mathcal P_Q$ denotes a learnable projection and $\mathcal M_T$ denotes the native text MLP of the DiT backbone. The resulting conditioning sequence is injected into the WAM cross-attention layers together with image tokens.

The standard sequential rollout decoding process suffers from severe computational inefficiency for long-chain latent reasoning. Traditional autoregressive rollout generates latent reasoning states one by one in a serial manner, which demands exhaustive forward propagation of Transformer layers and repetitive calculation of overlapping visual-linguistic features at each reasoning step. This serial dependency bottleneck drastically increases inference latency, restricts the real-time interaction capability of embodied manipulation tasks, and hinders the scalability of long-horizon reasoning rollouts. To address the serial decoding overhead and redundant computation in conventional latent reasoning rollout, we propose a novel Staircase latent decoding paradigm to achieve efficient parallel generation of continuous latent reasoning sequences.

\vspace{0.5em}
\noindent
\textbf{Staircase latent CoT decoding.}
Instead of autoregressively generating reasoning tokens one-by-one, WALL-WM models reasoning as a compact sequence of $K_c$ continuous latent reasoning states:
\begin{equation}
\hat y_{1:K_c}
=
\{\hat y_1,\dots,\hat y_{K_c}\}.
\end{equation}

The reasoning branch is initialized from the fine-tuned Qwen3.5-9B checkpoint and implemented as a lightweight Mixture-of-Transformers (MoT) structure coupled to the frozen backbone. To enable parallel latent rollout, we partition the Transformer at a relay depth $N_r$, where lower layers encode shared visual-language grounding features and higher layers progressively specialize into different reasoning steps.

Concretely, only the first latent position traverses the lower Transformer layers, producing a shared relay representation reused across all reasoning positions. The remaining latent states are then generated in parallel through the upper Transformer blocks with independent causal cache updates. This staircase scheduling mechanism transforms the conventional sequential reasoning process into a depth-parallel latent rollout:
\begin{equation}
\hat y_{1:K_c}
=
\mathcal F_{\mathrm{stair}}
(
x; N_r
),
\end{equation}
where $N_r$ controls the relay depth separating shared grounding computation from parallel reasoning computation.

Compared with standard autoregressive decoding, the proposed staircase design avoids repeatedly recomputing low-level visual-language features for every reasoning step, substantially reducing inference latency for long-horizon embodied rollout. The resulting latent reasoning states are fully differentiable and directly injected into the WAM cross-attention pathway without discrete token sampling.

\vspace{0.5em}
\noindent
\textbf{Frozen latent-to-text reconstruction supervision.}
Rather than directly distilling autoregressive hidden states, WALL-WM supervises latent reasoning through a frozen latent-to-text reconstruction objective~\cite{kang2025ladir,zhong2026dualcot}. The generated latent reasoning states $\hat y_{1:K_c}$ are projected by a prefix projector $\mathcal P_{\mathrm{pref}}$ into a soft prefix $\mathbf z_{1:K_c}{=}\mathcal P_{\mathrm{pref}}(\hat y_{1:K_c})$ in the embedding space of a lightweight frozen language model (Qwen3.5-0.8B), which reconstructs the corresponding textual CoT trace autoregressively:
\begin{equation}
P_\phi(r_{1:M_r}\mid \mathbf z_{1:K_c}).
\end{equation}

The training objective is defined as the token-level reconstruction loss:
\begin{equation}
\mathcal L_{\mathrm{CoT}}
=
-
\sum_{m=1}^{M_r}
\log
P_\phi
\left(
r_m
\mid
\mathbf z_{1:K_c},
r_{<m}
\right).
\end{equation}

Only the staircase reasoning branch and the prefix projector $\mathcal P_{\mathrm{pref}}$ are optimized, while the reconstruction language model remains frozen throughout training. Consequently, the latent reasoning states are encouraged to encode compact high-level reasoning semantics rather than replicate exact token-level decoding trajectories. The resulting latent CoT serves as a differentiable, compact, and computationally efficient replacement for explicit textual reasoning during embodied rollout.

\begin{figure}[t]
\centering
\vspace{-3ex}
\includegraphics[width=1\linewidth]{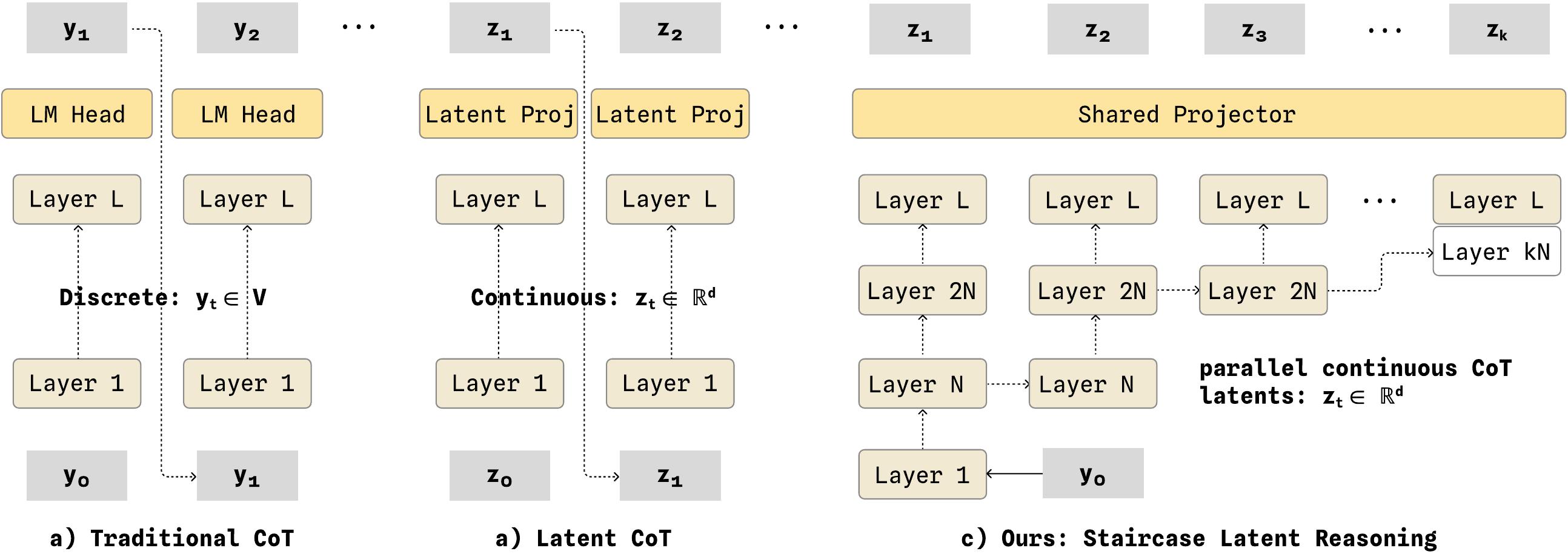}
\caption{
\textbf{Comparison of three CoT reasoning schedules.}
\textit{(a)} Traditional CoT autoregressively emits discrete vocabulary tokens.
\textit{(b)} Latent CoT replaces discrete tokens with continuous latent vectors while preserving the serial dependency.
\textit{(c)} Our staircase latent reasoning relays intermediate hidden states across staggered layer depths, producing parallel continuous CoT latents through a shared projector.
}
\label{fig:cot_mot_arch}
\end{figure}

\textbf{Inference modes.}
The same event-pretrained backbone exposes two complementary inference modes through this language-guided pathway. In \textbf{event mode}, the system rolls out in event space: a VLM, human, or agent proposes the next-event description for the current observation, and WALL-WM executes the corresponding variable-length video--action segment before observing the next state, so that rollout follows the natural duration of each task stage rather than a fixed control horizon. In \textbf{unified mode}, the staircase decoder emits $K_c$ continuous latent CoT states in a single parallel pass, which slot into the WAM cross-attention exactly where an atomic-instruction condition would path. Because both modes condition the WAM through the same reasoning interface, WALL-WM unifies open-ended, event-driven rollout, and standard fixed-horizon VLA execution within a single model.

\section{Training Data}
\label{sec:data}
The \ours data ecosystem contains large-scale internet videos, egocentric videos, robot-free UMI-style recordings, and heterogeneous teleoperation data, with cautious post-processing and rich annotation for general-purpose pretraining.
The data ecosystem is organized along five axes: \emph{(i) source coverage}, a data map spanning general internet video, egocentric human video, non-embodiment UMI data, and robot teleoperation/open robot data; \emph{(ii) temporal synchronization and data postprocessing}, a data post-processing framework for training and deployment alignment; \emph{(iii) hierarchical caption granularity}, with every episode annotated at four temporal scales (Task / Subtask / Action / Segment) plus an optional human-annotated layer; \emph{(iv) joint vision-language clustering paired with action clustering}, used to balance the dataloader and to surface long-tail non-nominal trajectories; and \emph{(v) targeted recovery augmentation}, covering contact-rich action regions and geometry / occlusion regimes that are difficult to capture through real-robot data collection.

\subsection{Data Composition}
\label{sec:data-composition}
We organize the \ours dataset as a \textbf{data-source map} rather than a strict hierarchy.
The map separates source families by viewpoint and action availability.
General internet video provides broad visual and temporal-dynamics priors at a scale that no embodied corpus can match; in our current build this includes a 1.2M-clip OpenVID~\cite{nan2024openvid} slice together with other web video sources.

\begin{wrapfigure}{r}{0.38\linewidth}
    \vspace{-11pt}
    \centering
    \includegraphics[width=\linewidth]{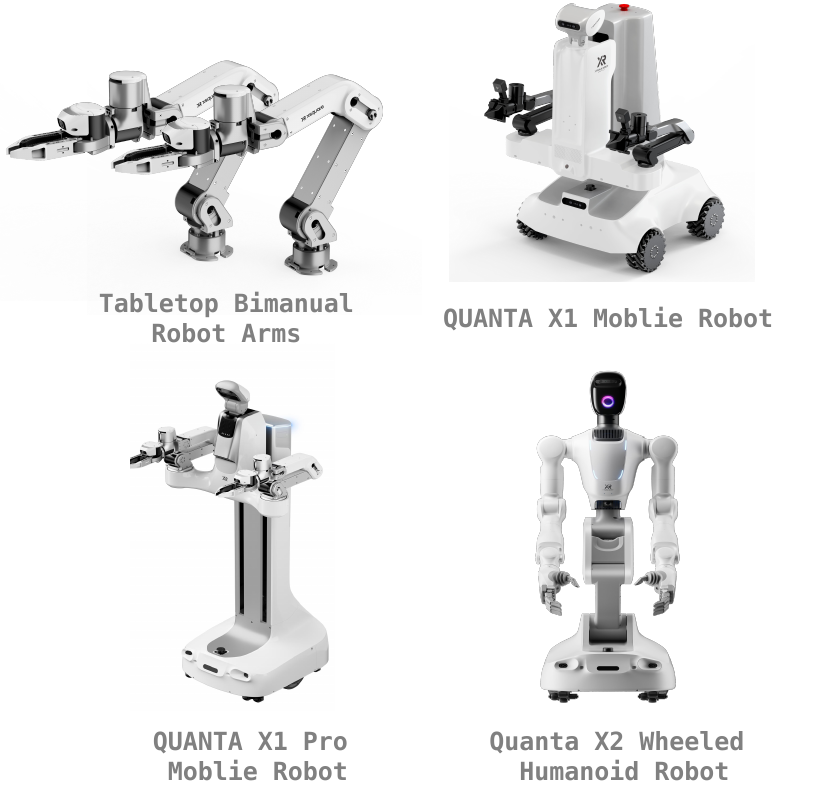}
    \vspace{-16pt}
    \caption{\textbf{Internal self-developed deployment platform.}}
    \label{fig:x2-robots}
    \vspace{-10pt}
\end{wrapfigure}

Egocentric and human-action videos narrow this prior toward first-person manipulation geometry without requiring robot actions.
Robot-free UMI-style recordings add deployment-like hand and camera geometry with retargetable controller trajectories, while heterogeneous robot datasets and self-collected teleoperation provide real \texttt{[video+action]} pairs across platforms.
Human-intervention and failure-recovery data are treated as a central source because they can be sampled from or attached to multiple quadrants whenever contact-rich correction behavior is needed.

The robot-data quadrant combines embodied public datasets and geometry-distribution-consistent self-collected data, where both observation geometry and action measurement are aligned with the deployment platform.
Our internal self-developed embodiment suite spans four deployment platforms: high-performance desktop bimanual robot arms; two mobile robot platforms, QUANTA X1 and QUANTA X1 Pro; and the wheeled humanoid robot QUANTA X2, which is equipped with high-degree-of-freedom dexterous hands.
Recovery and takeover episodes explicitly cover the action solution space near contact-rich events and provide failure-recovery signal that nominal demonstrations rarely supply by sheer volume.

\begin{figure}[t]
\centering
\includegraphics[width=1\linewidth]{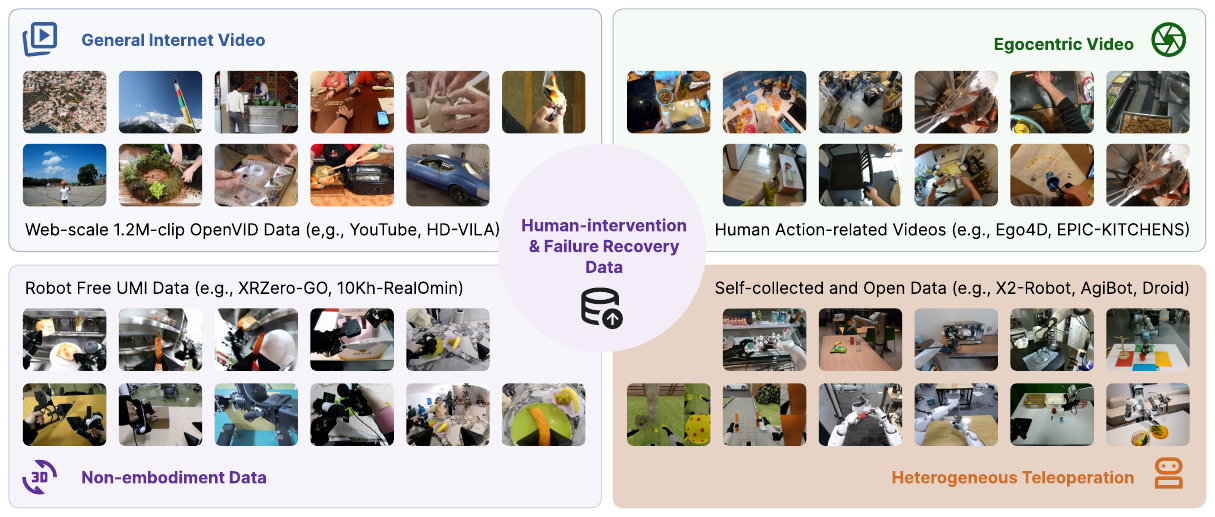}
\caption{\textbf{Data-source map of the \ours dataset.}
The four quadrants group general internet video, egocentric video, non-embodiment data, and heterogeneous teleoperation/open robot data; the center denotes human-intervention and failure-recovery data used to enrich contact-rich corrections.
Data sources include OpenVID~\cite{nan2024openvid}, HD-VILA~\cite{xue2022advancing}, Ego4D~\cite{grauman2022ego4d}, EPIC-KITCHENS~\cite{damen2022rescaling}, DROID~\cite{khazatsky2024droid}, AgiBot World~\cite{bu2025agibot}, and self-collected data~\cite{wang2026xrzero}.}
\label{fig:data-sources}
\end{figure}

\begin{figure}[t]
\centering
\includegraphics[width=1\linewidth]{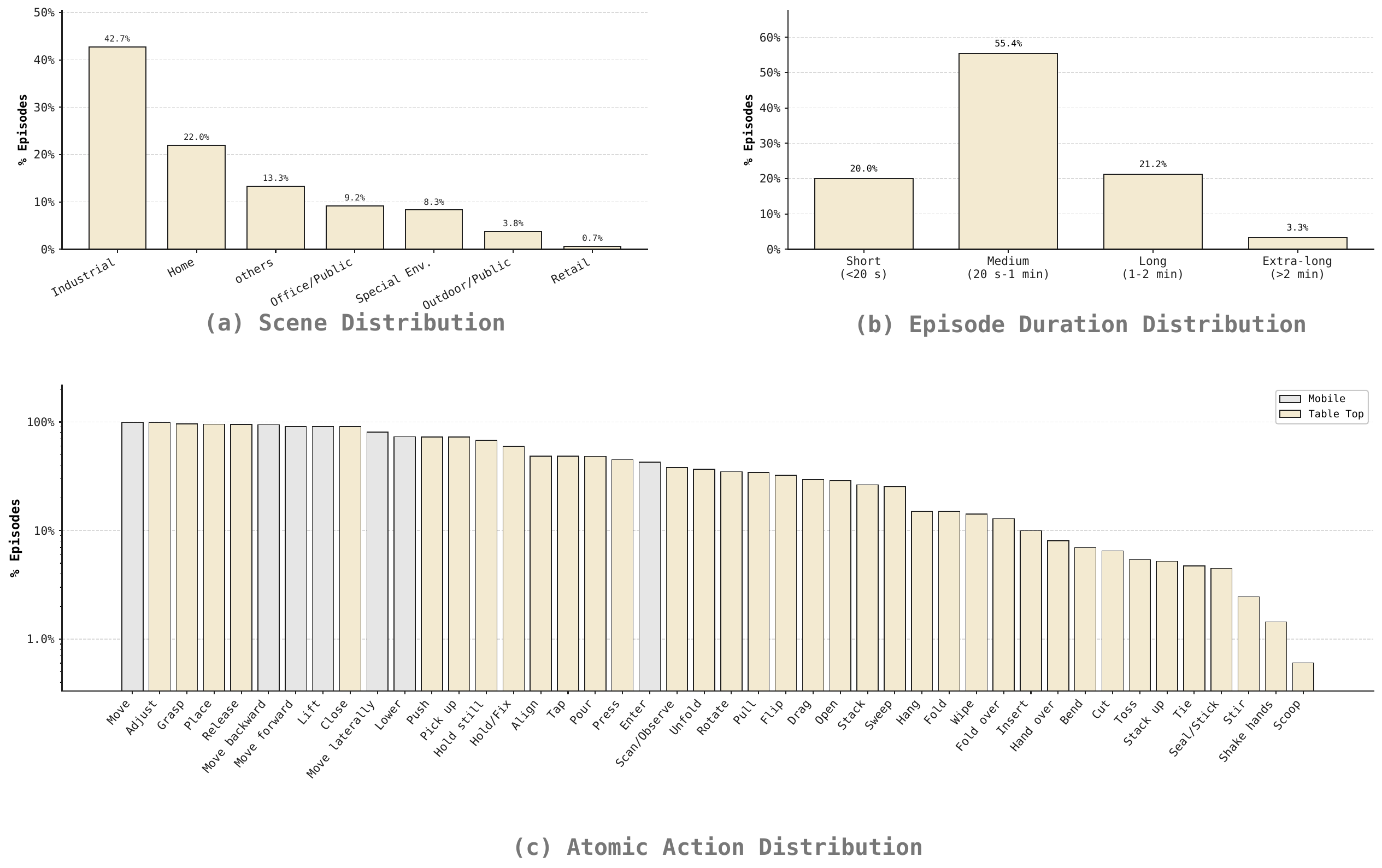}
\caption{\textbf{Distribution statistics of the \texttt{[video+action]} data.}
The plots show normalized coverage across scene categories, episode-duration ranges, and atomic-action labels. Scene and duration categories are reported as percentage distributions, while atomic actions indicate the percentage of episodes containing each action label.}
\label{fig:data-stats}
\end{figure}

\begin{figure}[!htbp]
\centering
\includegraphics[width=0.98\linewidth]{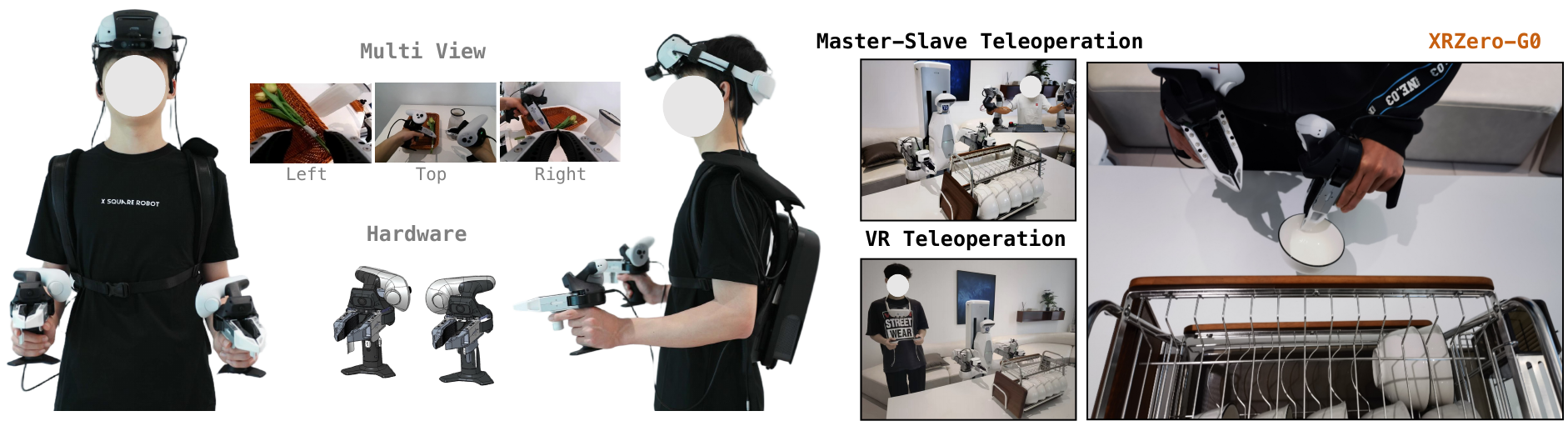}
\caption{\textbf{XRZero-G0 no-embodiment collection rig and its place alongside teleoperation~\cite{wang2026xrzero}.}
\emph{(Left)} The wearable rig from two views: a VR-tracked headset carries three egocentric cameras (one top + two wrist-side feeds shown), an operator-borne backpack handles synchronization and edge-side compute, and a pair of handheld grippers reproduces the deployment robot's end-effector geometry; the inset thumbnails show the three on-rig views (left wrist / top / right wrist) that the policy will eventually consume.
\emph{(Right)} XRZero-G0 sits alongside master-slave and VR teleoperation rather than replacing them: in the production recipe a small real-robot \emph{anchor} fraction (the two teleoperation modes) is paired with a much larger no-embodiment volume (the XRZero-G0 rig), realizing the few-shot physical-anchoring regime characterized in~\cite{wang2026xrzero}.}
\label{fig:xrzero-g0}
\end{figure}

\paragraph{Non-Embodiment Collection with a Wearable UMI-Style Rig.}
The non-embodiment quadrant of \Cref{fig:data-sources} is collected with a wearable, robot-free rig rather than a teleop'd robot.
An operator wears a VR-tracked headset carrying multi-view egocentric cameras and a pair of handheld grippers whose physical geometry is calibrated to match the deployment robot's end-effectors (\Cref{fig:xrzero-g0}, left); the resulting 6-DoF controller trajectories, after IK retargeting against the deployment robot's URDF, yield an admissible action stream, while the ego$+$dual-wrist video already sits in the geometry the real robot will see.
We use our own \textbf{XRZero-G0} system~\cite{wang2026xrzero} for this quadrant (\Cref{fig:xrzero-g0}, right): it sits alongside master-slave and VR teleoperation rather than replacing them, and contributes the VR-tracked acquisition stack with millimeter-level pose accuracy.
Because the operator is no longer attached to a robot, collection throughput is no longer bound by robot time; in the production recipe a small real-robot \emph{anchor} fraction is paired with a much larger volume of no-embodiment clips, following the few-shot physical-anchoring regime characterized in~\cite{wang2026xrzero}.

\paragraph{Collection Protocols: Structured and Unstructured Streams.}
The self-collected robot data in \Cref{fig:data-sources} is drawn from two collection protocols whose distinction sits entirely on the collection side, not on the training side.
\emph{Structured} collection follows a predefined task scope and reset protocol: each episode is one teleop'd demonstration of a named task, with explicit start / goal conditions and clean episode boundaries.
\emph{Unstructured} collection, by contrast, lets operators move the robot freely in the deployment scene without committing to a task scope, a reset protocol, or an episode boundary in advance, yielding a long, multi-event in-distribution motion stream that a demonstration-only dataloader would have to discard.
Both protocols feed the \emph{same} downstream pipeline: the hierarchical caption schema of \Cref{sec:data-caption} segments the raw stream into Task / Subtask / Action / Segment spans \emph{after} collection, and the clustering pass of \Cref{sec:data-clustering} treats the segmented clips identically regardless of which protocol produced them.
Admitting unstructured collection is what lets this source scale beyond the natural ceiling of teleop'd demonstrations: it removes the per-episode protocol overhead that otherwise binds collection throughput, while the caption-then-cluster pipeline absorbs the resulting heterogeneity into the same balanced sampler used everywhere else.
The overall distribution statistics of \texttt{[video+action]} data can be found in ~\Cref{fig:data-stats}.

\subsection{Deployment-Aligned Temporal Synchronization and Data Postprocessing}
\label{sec:data-postprocess}
For the non-embodiment data and robot data collected from tele-operation, the \texttt{[video+action]} layers are only useful if the visual observation and the action stream refer to the same physical instant.
However, this alignment is not guaranteed in practice: camera encoding, controller logging, teleoperation middleware, and disk writing can introduce a nearly constant video--action offset within a recording source.
If left uncorrected, the model is trained to associate an image at time $t$ with an action from $t+\Delta$ or $t-\Delta$.
This is especially damaging near contact, where a few frames can change the semantic state from ``approaching'' to ``touching'', ``grasping'', or ``recovering''.
It can also lead to undesirable shaking behavior when the trained policy is deployed on robot hardware.

\begin{figure}[t]
\centering
\includegraphics[width=1\linewidth]{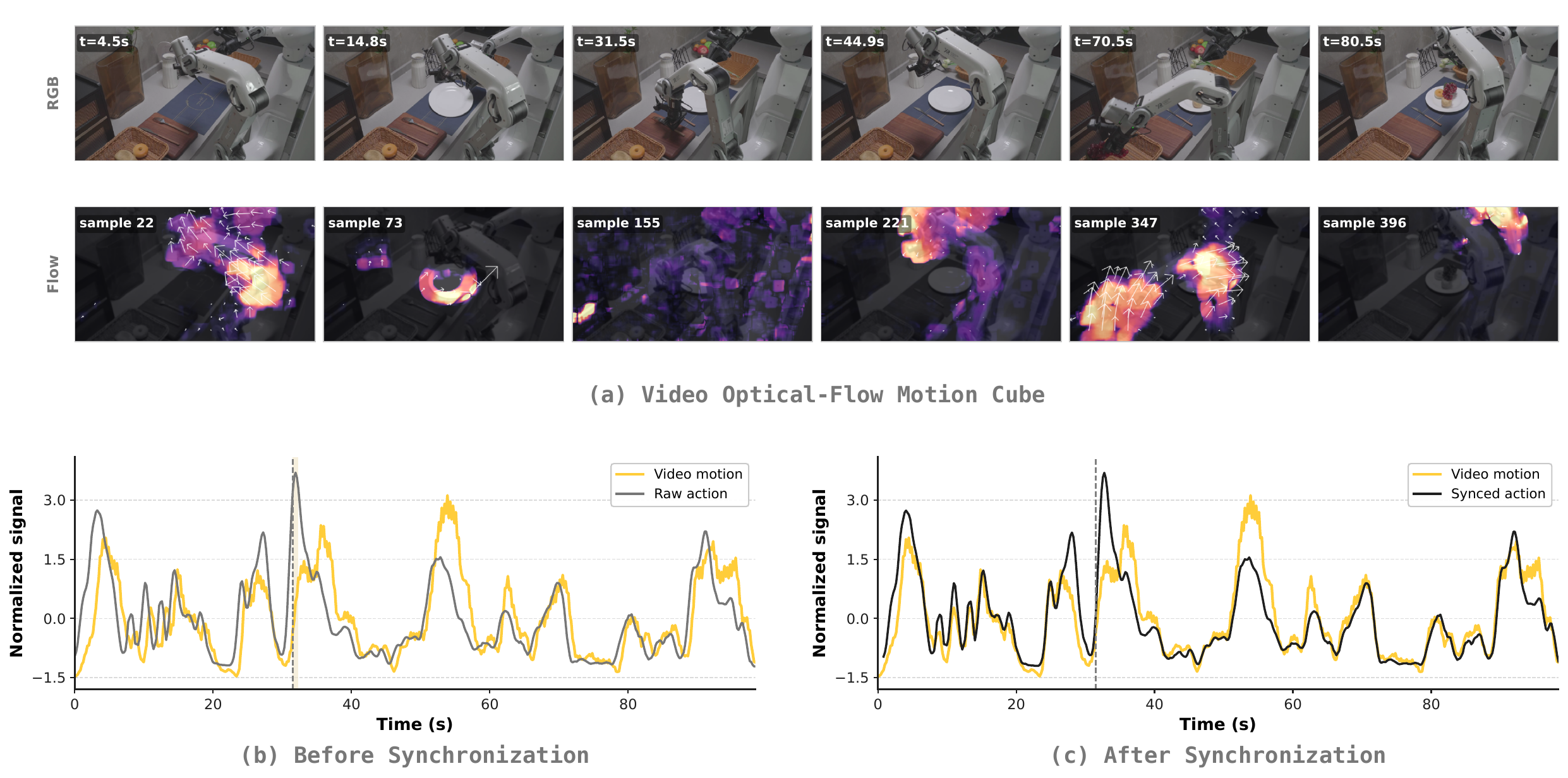}
\caption{\textbf{Temporal Synchronization of \texttt{[video+action]} layers.}
We illustrate the process of temporal synchronization for a specific timestep. 
(a) Calculating the optical-flow motion cube from video data: we grayscale and normalize the camera RGB images, then calculate the \textit{optical flow} from the video data.
(b) Before synchronization: there is a constant lagging between video signal and action signal.
(c) We performed temporal synchronization calculating the correlation between the video signal and the action signal, and synchronized the action signal based on the results.
}
\label{fig:data-sync}
\end{figure}

Therefore, we conduct \textbf{temporal-synchronization} of the \texttt{[video+action]} layers before captioning, clustering, and sampling.
Specifically, for each episode, we construct a visual motion signal by measuring frame-to-frame image change across available ego and wrist cameras, and an action-motion signal by taking the finite difference of the left and right end-effector position streams.
Both signals are smoothed and normalized, and we sweep a small integer lag window to find the offset that maximizes their correlation.
The estimate is aggregated over cameras and action channels so that a transient occlusion or a stationary gripper does not dominate the decision.
The illustration of this whole process can be found in \Cref{fig:data-sync}.

Rather than assuming perfect video-action timing a priori, this post-processing stage explicitly estimates the residual phase difference and calibrates it before downstream use.
For example, at 20 FPS, if an episode is estimated to have a two-frame offset, corresponding to roughly 100 ms, the synchronizer applies the corresponding temporal shift to align the logged action stream with the encoded video timeline.
The correction is applied deterministically: the action sequence is re-indexed to the visual frame timeline, boundary frames introduced by the shift are trimmed, and continuous action fields are resampled only after the offset has been removed.
Episodes whose best-lag score is weak, whose source-level vote is inconsistent, or whose required correction falls outside the expected window are not silently used; they are either quarantined for manual inspection or down-weighted until a source-level correction can be verified.

After synchronization, we run a second cleaning pass for \textbf{deployment alignment}.
This pass removes episodes with missing camera streams, non-monotonic or duplicated action records, abnormal FPS, large frame/action length mismatch, invalid robot states, kinematic discontinuities, gripper-state corruption, or motion segments that violate the deployment robot's reachable workspace after retargeting.
For structured demonstrations, failed resets and truncated starts/goals are removed.
For unstructured streams, long recordings are split only after synchronization, so that discovered Task/Subtask/Action/Segment spans inherit a consistent video-action clock.
The cleaned output of this stage is a canonical episode format: synchronized multi-view video, aligned action records, calibrated camera metadata, robot-state metadata, source identity, confidence flags, and cleaning provenance.
These flags are consumed downstream by the hierarchical captioner and by the cluster-balanced dataloader.

\subsection{Event-Centric Hierarchical Captioning}
\label{sec:data-caption}
Every episode in the action-paired sources of \Cref{fig:data-sources} is annotated with a temporally nested caption hierarchy whose spans are \emph{temporally grounded} on atomic manipulation actions~\cite{shou2021generic}---short, executable primitives such as reach, grasp, close, lift, and place---rather than on fixed frame windows.
Concretely, we segment the synchronized video--action stream at action boundaries first, then assign captions so that each linguistic interval aligns with the same physical interval in vision and control.
Rather than assigning a single free-form description to an entire trajectory, we decompose each episode into four core levels of temporal abstraction, together with an optional human-annotated layer. 
This hierarchy is designed to preserve both the global intent of the demonstration and the local event structure that determines how the robot actually completes the task.

The four core levels are \emph{Task}, \emph{Subtask}, \emph{Action}, and \emph{Segment}.
The \emph{Task} level is an episode-global string that summarizes the overall objective, such as opening a drawer, placing an object into a container, or wiping a surface.
The \emph{Subtask} level partitions the episode into a small number of contiguous, semantically meaningful stages. These stages typically correspond to intermediate goals, for example approaching the target object, establishing a grasp, transporting the object, and placing or releasing it. 
The \emph{Action} level further refines each subtask into shorter manipulation primitives, such as reaching, aligning the gripper, closing the fingers, lifting, translating, inserting, or retracting. Finally, the \emph{Segment} level provides the finest temporal decomposition, capturing short and localized events that may last only a small fraction of the full trajectory.
A fifth \emph{Human} level stores manually annotated segments. This layer is populated only for the subset of episodes that received human annotation, and serves two complementary purposes. First, it provides high-quality supervision for validating the automatically generated hierarchy. 
Second, it gives us a trusted reference for evaluating caption accuracy, temporal boundary quality, and the consistency of semantic labels across tasks. In practice, the human layer is not required for every episode; the core four-level hierarchy allows the dataset to scale, while the human subset anchors the annotation process and supports quality control.

\begin{figure}[!htbp]
\centering
\includegraphics[width=1.0\linewidth]{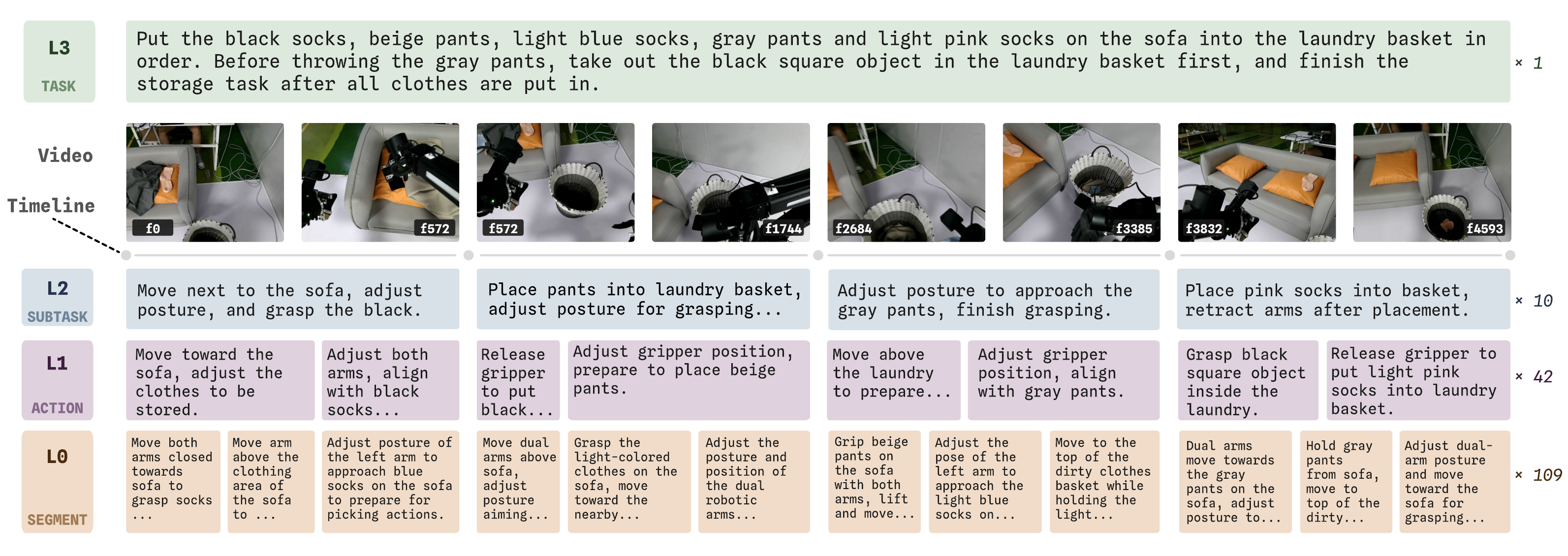}
\caption{\textbf{Four-level caption tracks for a single episode.}
The visualization stacks the four core lanes shown in the figure: Task (L3), Subtask (L2), Action (L1), and Segment (L0), aligned to the video timeline.}
\label{fig:caption-tracks}
\end{figure}

This multi-level schema is especially important for robotic recovery behavior. Many useful demonstrations are not perfectly linear executions of a task.
They contain regrasps, failed contact attempts, small pose corrections, retries after slipping, and other short recovery motions.
These events are often essential for robust policy learning, because they expose the model to states that arise when execution deviates from the nominal path.
However, if an entire episode is represented only by a single caption, these corrective behaviors are averaged into the global task description and become difficult for the model to identify. The trajectory may still be labeled as ``pick up the object,'' even though a crucial portion of the episode involved missing the grasp, repositioning the wrist, and trying again.
The hierarchical caption structure makes these events explicit. At the task level, the episode remains associated with its intended goal.
At coarser subtask and action levels, the model observes the main phases of execution. At the finest segment level, short corrective behaviors can be localized in time and assigned their own descriptions.
This enables the dataloader to sample or re-weight specific temporal regions, rather than treating all frames within a successful episode as equally informative. As a result, recovery-relevant intervals can be emphasized during training without discarding the broader task context.

\subsection{Cluster-Balanced Sampling over Vision-Language and Action}
\label{sec:data-clustering}

The production recipe trains on only a small subset sampled from our large-scale self-collected embodied corpus, selected by the cluster-balanced sampler described below. 
This subsampling is not merely a budget concession: it follows from the observation that the raw embodied corpus is highly long-tailed, while the multi-level caption schema of \Cref{sec:data-caption} exposes fine-grained statistics over vision, language, and action.
In particular, the caption windows are segmented according to action boundaries before clustering, rather than by uniform temporal slicing. This action-aligned decomposition turns each long demonstration into semantically localized units, making the resulting language and vision--language distributions substantially more balanced than the original task-level mixture.

On top of this multi-level schema, WALL-WM runs two offline clustering passes whose purpose is to give the dataloader a controllable sampling distribution.
The first is \emph{vision--language} (VL) clustering: a frozen multimodal encoder maps each visual observation and caption pair into a joint embedding, and a clustering procedure partitions this space into topic clusters that summarize the corpus's instruction-scene coverage.
The second is \emph{action} clustering: action chunks are clustered separately in trajectory space, where the long tail concentrates non-nominal motion such as recoveries, re-grasps, retries, and contact-driven corrections.
These are precisely the behaviors that the action-aligned multi-level schema is designed to surface. Importantly, after action-based decomposition from task-level captions to finer action and segment captions, the language-only and vision--language clustering views become more evenly distributed, with fewer samples absorbed by a small number of dominant topics. This provides a practical basis for balanced sampling: rare-but-important instruction-scene and trajectory modes become explicit sampling units instead of being overwhelmed by frequent verbs and nominal motions.

During training, the dataloader samples batches that balance both VL clusters and action clusters. 
VL balancing improves coverage over instruction-scene combinations, while action balancing repeatedly exposes the action head to trajectories that share common verbs but require different motions.
This is the regime in which a vanilla SFT objective on the raw long-tailed mixture would tend to collapse toward dominant targets. 
Both clusterings are precomputed offline; training consumes only cluster assignments, so cluster-balanced sampling adds neither encoder forward passes nor optimizer state to the existing recipe.

\begin{figure}[!htbp]
\centering
\includegraphics[width=0.95\linewidth]{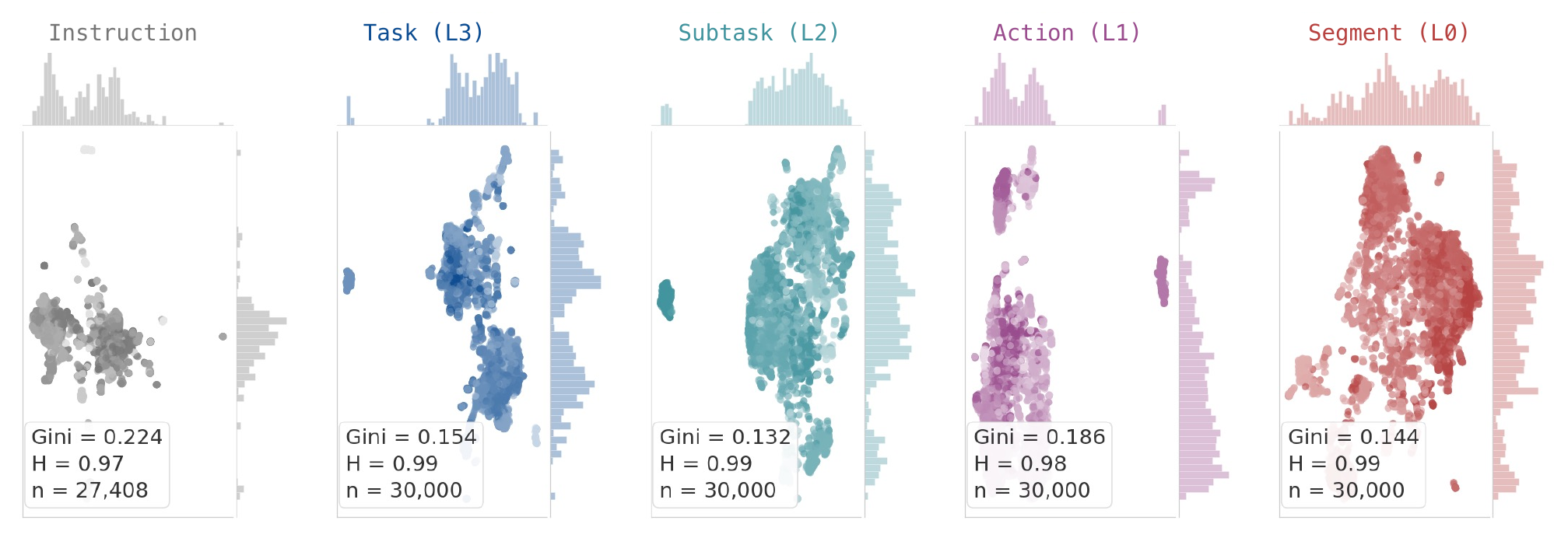}
\caption{\textbf{Caption-granularity distributions for balanced sampling.}
The five panels compare 30K samples at Instruction, Task (L3), Subtask (L2), Action (L1), and Segment (L0) granularities after action-boundary segmentation.
Finer caption levels expose more localized long-tail modes that are consumed by the cluster-balanced dataloader.}
\label{fig:vl-clustering-grid}
\end{figure}

\subsection{Recovery Data via Contact-Rich Random Initialization}
\label{sec:data-recovery}

Pretraining is augmented with \emph{recovery / takeover} episodes that explicitly cover the action solution space around contact-rich events.
For each such event $e$, we define a local contact-pose distribution $p(\mathbf{q}\mid e)$ over end-effector configurations $\mathbf{q}\in\mathbb{R}^{D_q}$.
Its support lies in a small geodesic ball $\mathcal{B}_\epsilon(\mathbf{q}^\star_e)$ centered at the nominal contact pose $\mathbf{q}^\star_e$, so the sampled states remain local to the event while preserving its contact geometry.

The resulting pretraining distribution is
\[
\tilde{p}_\text{train}(\mathbf{q},e)
= (1-\alpha)\, p_\text{nominal}(\mathbf{q},e)
+ \alpha\, \mathbb{E}_e\!\left[p(\mathbf{q}\mid e)\right],
\]
where $\alpha$ is a small mixing weight.
In practice, we sample the second term by perturbing the robot initialization around $\mathbf{q}^\star_e$ and then either replaying the original demonstration from the perturbed state or collecting a fresh recovery rollout.
This creates controlled local coverage around each contact event, rather than observing only the single nominal trajectory that passes through it.

This recovery mixture occupies the central recovery source in \Cref{fig:data-sources} and complements the cluster-balanced sampling protocol of \Cref{sec:data-clustering}. 
Clustering surfaces non-nominal trajectories that already exist in the corpus, while recovery initialization actively creates such trajectories in contact-space regions where the nominal data is too sparse for clustering alone to provide adequate coverage.

\section{Training and Inference Recipe}
\label{sec:training}

This section details the full training and inference recipe for \ours.
\Cref{tab:training-stages} summarizes which components are trained at each stage.
We first pretrain the layer-coupled video-action denoiser under the \emph{next-event prediction} scheme of Fig.~\ref{fig:vla_alignment}(A) (\Cref{sec:training-va-pretrain}), training the video tower first and then the action tower with the video tower frozen.
We then adapt the VLM text conditioner (\Cref{sec:training-text-encoder}) and train the Staircase decoder for unified-mode parallel chain-of-thought (\Cref{sec:training-latent-reasoning}).
An optional fine-tuning stage adapts both towers to the observation-centered history window under the \emph{next-chunk prediction} scheme of Fig.~\ref{fig:vla_alignment}(B) (\Cref{sec:training-deployment-refinement}); event-mode rollout requires only the preceding stages.
\Cref{sec:training-inference} concludes with the event-mode and unified-mode inference protocols supported by the assembled stack.

\begin{table}[!htbp]
\centering
\small
\caption{\textbf{Trainable / frozen matrix.}
\ding{51} indicates a component trained in that stage; \ding{55} indicates a component kept frozen; ``--'' indicates a component not yet attached.
Columns follow the training pipeline in order: event-centric video and action pretraining (\Cref{sec:training-va-pretrain}, first two columns), VLM text-conditioning (\Cref{sec:training-text-encoder}), Staircase distillation (\Cref{sec:training-latent-reasoning}), and optional next-chunk adaptation (\Cref{sec:training-deployment-refinement}).
Video and action pretraining use frozen T5 features; next-chunk adaptation keeps T5 frozen for global instructions while updating the V-A denoiser.
The VLM stack attaches from stage~3 onward; only MoT branches train in the Staircase stage.}
\label{tab:training-stages}
\begin{tabular}{l c c c c c}
\toprule
\multirow{2}{*}{\textbf{Component}} & \multicolumn{5}{c}{\textbf{Training stages}} \\
\cmidrule(lr){2-6}
& \textbf{Video PT} & \textbf{Action PT} & \textbf{VLM text} & \textbf{Staircase} & \textbf{Next-chunk} \\
\midrule
3D causal VAE $\mathcal{E}_V$        & \ding{55} & \ding{55} & \ding{55} & \ding{55} & \ding{55} \\
T5 text encoder                       & \ding{55} & \ding{55} & \ding{55} & --         & \ding{55} \\
VLM backbone (Qwen3.5-9B)             & --         & --         & \ding{55} & \ding{55} & \ding{55} \\
VLM project-out / auxiliary heads     & --         & --         & \ding{51} & \ding{55} & \ding{55} \\
Video DiT (incl.\ view attention)     & \ding{51} & \ding{55} & \ding{55} & \ding{55} & \ding{51} \\
Action DiT (incl.\ layer coupling)    & --         & \ding{51} & \ding{55} & \ding{55} & \ding{51} \\
Staircase MoT branches                & --         & --         & --         & \ding{51} & \ding{55} \\
\bottomrule
\end{tabular}
\end{table}

\subsection{Event-Centric Video-Action Pretraining}
\label{sec:training-va-pretrain}

Event-centric video-action pretraining implements \emph{next-event prediction} on individual semantic events, each providing paired ground-truth video and action for one manipulation segment.
Training proceeds in two sequential stages: video pretraining, followed by action pretraining.
In the first stage, we train only the video DiT on event latents under the Wan-style $v$-prediction flow-matching objective $\mathcal{L}_V$ of \Cref{eq:video-flow-loss}, adapting the inherited Wan prior into an embodied, multi-view event-future predictor; the action tower is not yet attached.
In the second stage, we freeze the video DiT and train only the action tower under action flow matching to predict the aligned end-effector trajectory.

\paragraph{Video Pretraining.}
Video pretraining includes the cross-view branch of \Cref{sec:method-video}, whose output projector is initialized to zero.
Starting from the Wan prior, the DiT is conditioned on the current multi-view observation (one keyframe per camera) and on event-aligned captions from \Cref{sec:data-caption}, encoded by a frozen T5; video timesteps are sampled uniformly.
Because event spans vary in duration, we truncate each clip to at most $65$ latent frames (corresponding to $129$ raw frames under stride-$2$ subsampling).
We further apply a length-aware caption-drop schedule: let $L_e$ denote the raw event span length in frames and let $\rho(L_e)\in[\rho_{\min},\rho_{\max}]$ be the probability of omitting the event caption on a training step,
\begin{equation}
\rho(L_e)=
\begin{cases}
\rho_{\min}, & L_e\le L_{\min},\\[4pt]
\rho_{\min}+(\rho_{\max}-\rho_{\min})\,\dfrac{1-\cos\!\bigl(\pi\tfrac{L_e-L_{\min}}{L_{\max}-L_{\min}}\bigr)}{2}, & L_{\min}<L_e<L_{\max},\\[6pt]
\rho_{\max}, & L_e\ge L_{\max},
\end{cases}
\label{eq:cfg-ratio}
\end{equation}
We set $\rho_{\min}{=}0.1$, $\rho_{\max}{=}0.9$, $L_{\min}{=}129$, and $L_{\max}{=}220$, with $L_{\min}$ matching the truncation cap above.
On caption-free steps, training reduces to \emph{observation-anchored} future synthesis: conditioned only on the current multi-view observation, the DiT must infer a physically plausible continuation of contact and end-effector dynamics rather than lexically specified sub-goals.
We prune quasi-static frames during event construction so that flow-matching supervision concentrates on segments with salient end-effector motion.
We additionally use three training practices: (i) border masking that excludes out-of-frame regions and synthetic black borders from $\mathcal{L}_V$ (\Cref{eq:video-flow-loss}); (ii) an EMA of the DiT weights (decay $\beta_{\mathrm{ema}}{=}0.9999$) updated each step; and (iii) resolution/length bucketing that groups events of similar latent shapes to amortize padding across variable-length clips.
Training draws from the cluster-balanced sampler of \Cref{sec:data-clustering}.

\paragraph{Action Pretraining.}
After video pretraining, we load the resulting video DiT and keep it fixed. We then optimize only the action transformer under action flow matching on the same event-centric video-action pairs of \Cref{sec:data-caption}, disabling the video flow-matching loss so that action gradients do not update the video tower.
Following the asymmetric 1-to-$N_d$ mapping of \Cref{sec:method-action}, we pin the frozen video forward to the schedule anchor $s^\star{=}45$ on a $50$-step denoising schedule (selected empirically, with a small jitter window during training). This single anchored video forward supplies cross-attention keys/values to every action denoising step ($m(j){=}s^\star$ for all $j{\in}\{0,\dots,N_d{-}1\}$, with $N_d{=}50$).
We additionally set $K{=}6$ parallel action-noise draws per optimizer step to reuse the same anchored video forward; this throughput trick is training-only and is not used at inference.
The action tower is randomly initialized; training follows the event-centric layout of \Cref{sec:method-action} and uses the cluster-balanced sampler of \Cref{sec:data-clustering}.

\subsection{VLM Text-Conditioner Pretraining}
\label{sec:training-text-encoder}
\label{sec:training-stage1-text}
In this stage we adapt only its project-out head and two lightweight auxiliary heads, without modifying the downstream DiT.
A single VLM forward pass over the current event is supervised to emit two signals: (i) a \emph{next-event} description---a task-decomposition caption of the upcoming sub-task---and (ii) an estimate of the \emph{remaining time} until the current event terminates.
The hidden states from this same forward pass provide the conditioning features for the DiT, so the latency profile matches a standard T5 conditioning path.

Crucially, the VLM hidden states are aligned to the downstream DiT's \textbf{T5} feature space, making the upgraded VLM a \emph{drop-in replacement} for T5 from the DiT's perspective.
This alignment preserves the DiT-side text-conditioning prior (the DiT continues to consume text in the geometry it was trained on) while importing the broader reasoning bandwidth of a multimodal VLM, including scene-grounded disambiguation, task decomposition, and a temporal anchor via remaining-time prediction.
Throughout this stage, the VLM backbone remains fully frozen; only the project-out head that produces $\mathbf{c}_\ell$, the next-event head, and the remaining-time regressor are trained.
Concretely, we minimize a three-term objective that aligns the VLM conditioning $\mathbf{c}_\ell^\text{VLM}$ (the VLM-produced version of $\mathbf{c}_\ell$ from \Cref{sec:method-text}) to the corresponding $\mathbf{c}_\ell^\text{T5}$ produced by the original T5 encoder on the same instruction, while supervising the two auxiliary heads,
\begin{equation}
  \mathcal{L}_{\text{text}}
  \;=\;
  \lambda_\text{align}\,
       \bigl\lVert\mathbf{c}_\ell^\text{VLM} - \mathbf{c}_\ell^\text{T5}\bigr\rVert^2
  \;+\;
  \lambda_\text{next}\,
       \mathrm{CE}\!\bigl(\hat c_{\text{next}},\,c_{\text{next}}\bigr)
  \;+\;
  \lambda_\text{time}\,
       \mathrm{Huber}\!\bigl(\widehat{\Delta t},\,\Delta t\bigr),
\end{equation}
where $c_{\text{next}}$ is the ground-truth next-event caption (a discrete token sequence, written with the same $c$ form as the per-event caption $c_e$ of \Cref{sec:method-action}) and $\hat c_{\text{next}}$ is its prediction; the first term aligns features to the original T5 encoding, the second is a token-level language-modeling loss on the next-event caption, and the third is a Huber regression on the remaining-time scalar $\Delta t$.
The three loss weights $\lambda_\text{align},\lambda_\text{next},\lambda_\text{time}$ are kept fixed throughout this stage.

\subsection{Staircase Distillation}
\label{sec:training-latent-reasoning}

The staircase latent CoT module in \Cref{sec:staircase} is implemented as a lightweight reasoning branch coupled to the frozen Qwen3.5-9B backbone through a Mixture-of-Transformers (MoT) architecture. Initialized from pretrained Qwen3.5 weights, the reasoning branch inherits the semantic reasoning capability of the original autoregressive model while enabling efficient parallel latent rollout. Given the visual-language input, the staircase module generates a compact sequence of continuous latent reasoning states $\hat y_{1:K_c}$, which serve as implicit reasoning variables instead of explicit textual tokens. These latent states are directly injected into the WAM cross-attention layers during embodied rollout.

For supervision, the latent sequence is projected by a prefix projector $\mathcal P_{\mathrm{pref}}$ into a soft prefix representation $\mathbf z_{1:K_c}$ in the embedding space of a frozen lightweight Qwen3.5-0.8B language model, which autoregressively reconstructs the corresponding textual CoT trace $r_{1:M_r}$:
\begin{equation}
P_\phi(r_{1:M_r}\mid \mathbf z_{1:K_c}).
\end{equation}

Training optimizes the latent-to-text reconstruction objective $\mathcal L_{\mathrm{CoT}}$, with gradients applied only to the staircase reasoning branch and the prefix projector $\mathcal P_{\mathrm{pref}}$. All other components remain frozen during this stage.

\subsection{Next-Chunk Adaptation}
\label{sec:training-deployment-refinement}

Following event-centric pretraining (\Cref{sec:training-va-pretrain}), this optional stage fine-tunes the event-pretrained backbone under \emph{next-chunk prediction} on the observation-centered layout of \Cref{sec:method-action}.
Under one global instruction, history frames localize each $H_a$-step chunk---$H_a$ denoting the fixed number of action steps per chunk, the fixed-length counterpart of the variable event horizon used in event-centric pretraining---substituting for the per-event captions used during event-centric pretraining.

\paragraph{Observation-centered fine-tuning.}
Global instructions come from the Task level of \Cref{sec:data-caption} and are encoded by the same frozen T5 as in pretraining; window geometry, relative-pose action targets, and the extended $E_{\text{abs}}$/$E_\tau$ indices follow \Cref{sec:method-action}.
Both DiT towers are updated on these fixed-shape windows under the same asymmetric 1-to-$N_d$ anchor protocol ($s^\star{=}45$, $m(j){=}s^\star$, with optional training-time $K{=}6$ action-noise reuse during optimization).

\paragraph{Cluster-balanced sampling.}
Sliding unified windows shift the instruction-trajectory co-occurrence relative to event-centric pretraining, so we rerun the offline VL and action clustering passes of \Cref{sec:data-clustering} on history-conditioned windows from the same corpus before optimization.
The dataloader balances the resulting clusters to retain coverage over rare global instructions and non-nominal trajectories surfaced by \Cref{sec:data-caption,sec:data-recovery}, without adding any encoder forwards or optimizer state at train time.

\subsection{Inference Recipe}
\label{sec:training-inference}

The assembled stack supports two inference modes that mirror the two schemes of Fig.~\ref{fig:vla_alignment}.
Event-mode inference runs \emph{next-event prediction} on the event-centric scheme~(A) layout and needs neither history nor a fixed chunk contract.
Unified-mode inference runs \emph{next-chunk prediction} on the observation-centered scheme~(B) layout after \Cref{sec:training-deployment-refinement}, and is used mainly for conventional fixed-horizon VLA evaluation.
Both modes use the asymmetric 1-to-$N_d$ anchor protocol of \Cref{sec:method-action} ($s^\star{=}45$, $m(j){=}s^\star$ on the default 50-step schedule).

\subsubsection{Event-Mode Inference}
\label{sec:training-inference-event}

At each rollout step the video and action DiTs condition on the current multi-view observation, proprioceptive state, and a \emph{next-event description} for the upcoming semantic event.
The description comes from a human or from the fine-tuned Qwen3.5-9B heads of \Cref{sec:method-text}, which also emit a remaining-time estimate for the present event.
The model denoises the full event-aligned video and action segment in the scheme~(A) variable-length window; when execution completes, the observation advances and a new next-event description conditions the next segment.

\subsubsection{Unified-Mode Inference}
\label{sec:training-inference-unified}

Unified-mode inference rolls out fixed $H_a$-step video-action chunks on the scheme~(B) layout, conditioning at each step on the current observation, a configurable history window, and a global instruction under the same anchor protocol.
During unified rollout, three interchangeable sources supply the per-chunk text-side context $\mathbf{c}_\ell$, either by reusing one encoded global instruction or by providing a fresh caption each chunk, and may be mixed without retraining.
\begin{itemize}
  \item \textbf{Gradient-continuous source.} The text encoder of \Cref{sec:method-text} maps the global instruction once into a continuous $\mathbf{c}_\ell$ that is fed verbatim into cross-attention and reused across all rolling chunks, with no discrete CoT bottleneck between language and the denoiser.
  \item \textbf{Atomic-instruction source.} An upstream planner or human operator provides one short instruction for each fixed $H_a$-step chunk boundary rather than for a variable-length semantic event, and the same text encoder maps each instruction to the $\mathbf{c}_\ell$ that drives the corresponding chunk.
  \item \textbf{VLM-CoT source.} The Staircase decoder of \Cref{sec:staircase}, together with the cloned MoT branches trained in \Cref{sec:training-latent-reasoning}, emits CoT latents in a single parallel forward, and their representations serve as the per-chunk text-side context for end-to-end VLA rollouts.
\end{itemize}
Because every source only replaces the cross-attention text input, sources can be switched mid-rollout without altering denoiser state.

\section{Infrastructure}

\subsection{General Frameworks}
Training and serving our model at scale exposes a set of systems challenges that differ in kind from those of pure language models. The co-existence of a heterogeneous backbone , pretrained latent prediction expert and randomly initialized action expert, and a mixed objective of video generation and action prediction, and tight latency budgets at deployment together create bottlenecks that off-the-shelf training and inference stacks do not address. We therefore develop a series of infrastructure-level optimizations tailored to our model.

\paragraph{Muon Optimizer.}
We adopt Muon as the optimizer for the majority of modules in our model, motivated by its consistent gains in convergence speed and training stability.
However, the Newton-Schulz (NS) iteration introduces significant overhead when applied directly to sharded training, in the vanilla implementation, the optimizer step alone can approach the 2x cost of the forward and backward passes combined.
To make Muon practical at scale, we implement \textbf{DMuon}, a distributed realization of Muon designed for hybrid parallelism and decoupled from the underlying training framework.
DMuon resolves the granularity mismatch between matrix-level NS and sharded parameters through co-designed distributed execution, reducing the optimizer step to a secondary cost rather than the dominant bottleneck.
\begin{itemize}
    \item \textbf{Pipeline Scheduling.}
    We adopt a dedicated-ownership scheme in which a constrained Longest-Processing-Time assignment maps each matrix parameter to a unique owner rank.
    The native all-reduce / reduce-scatter pattern is replaced by reduce / broadcast, with the post-step broadcast issued asynchronously on a dedicated CUDA stream and overlapped with the subsequent forward pass, while an adaptive runtime monitor falls back to synchronous broadcast under network contention. With the fine-grained scheduling optimization, both the computation and communication redundancy are eliminated.
    \item \textbf{Kernel Optimization.}
    The Gramian formulation surfaces a symmetric structure that general-purpose GEMM cannot exploit, leaving nearly half of the tile-level computation redundant by construction.
    We reclaim it through a CuteDSL kernel aligned to the symmetric factor, with shape-aware autotuning sustaining the speedup across the matrix geometries encountered in training.
\end{itemize}

\paragraph{\textbf{Kernel Library.}}
We develop a customized kernel library spanning both training and inference, motivated by the observation that stock kernels operate at single-operator granularity and consequently incur unnecessary spills to global memory on tensors that are produced and immediately consumed within a small neighborhood of the dataflow graph. We identify a set of recurring fusable patterns whose arithmetic intensity is bounded by memory bandwidth rather than tensor-core throughput in their default implementations. We consolidate these patterns into composite kernels that keep intermediate activations in registers or shared memory across the fused region, shifting the effective roofline operating point toward the compute-bound regime and sustaining high GPU utilization across the dominant cost centers of the model.

Our kernel library is built on top of TVM FFI, a language-agnostic foreign function interface that exposes compiled kernels through a stable C ABI with zero-copy tensor handles, bypassing the per-call cost of the PyTorch dispatcher and the GIL while remaining interoperable with the PyTorch training and inference frontend. On top of this dispatch layer, we provide a complete wrapper that mirrors the native PyTorch op interface, so that kernels from our library can be substituted for their PyTorch counterparts without any change to the surrounding model code. This design preserves the host-side launch savings while making it straightforward to maintain numerical consistency between training and inference, since both paths invoke the same compiled kernel through the same call surface.

\paragraph{\textbf{Fine-Grained Overlapping.}}
We introduce a view attention alongside the original spatial-temporal attention, resulting in two sequential attentions per layer.
Under Ulysses sequence parallelism, each attention requires a pair of all-to-alls to redistribute the sequence between the sequence and head dimensions, so a naive implementation serializes four all-to-alls per layer on the critical path and substantially inflates the communication cost of adding view attention.
To mitigate this, we apply fine-grained scheduling to hide the back-to-back communication within attention computation, eliminating most of the added communication overhead.

\paragraph{\textbf{Multi-Event Sequence Packing.}}
A common training recipe first \emph{offline}-encodes and caches video latents at \emph{entire-episode} granularity, then samples from those caches during optimization.
Whenever a minibatch is formed, the pipeline must therefore load and re-materialize a full episode's worth of latents (or repeat the VAE encode on the fly); because episode lengths vary widely, episodes cannot be packed into uniform tensors without aggressive padding or truncation, and the effective batch size often collapses in practice.
We instead pack \emph{multiple events} into a single long sequence at the dataloader level and train on the packed sequence in parallel, with an attention mask that blocks cross-event leakage.
Events are concatenated up to a fixed total length, so each training step always runs at the full configured effective batch size and the GPU stays close to compute-bound; the per-step cost is amortized across all packed events instead of being paid once per episode.

\subsection{Model Compression}
World models are notoriously expensive to deploy, the underlying diffusion-style backbone requires tens of denoising steps per generated frame, and each step itself involves a full forward pass through a multi-billion-parameter network, making naive deployment fundamentally incompatible with the latency budgets of real-time robotic control. To close this gap, we apply two complementary compression techniques along orthogonal axes—distillation to reduce the number of denoising steps, and FP8 quantization to lower the per-step compute and memory cost.
\paragraph{\textbf{Distillation.}}
To reduce the number of denoising steps, we adopt \emph{distribution-matching distillation} (DMD)~\cite{yin2024improved, yin2024onestep}, which trains a few-step student generator to align its output distribution with that of the multi-step teacher rather than regressing onto teacher trajectories pointwise.
Concretely, the distribution-matching term supervises the student via the gap between two score estimates—one from the frozen teacher and one from an auxiliary \emph{fake score network} trained online to track the student's evolving distribution—pushing the student toward regions where its own samples are under-represented relative to the teacher.
We further adopt a \emph{joint distillation} objective that retains the original action-prediction loss alongside this distributional term, so that the few-step student is simultaneously aligned with the teacher's video distribution and anchored to the action supervision it was pre-trained on—preventing the action head from drifting as the denoising trajectory is compressed.
Ablating the latter degrades action MAE by 53\%, confirming that compressing the denoising trajectory under distributional supervision alone causes the action head to drift away from its pre-trained calibration.
We initialize the student from teacher weights and distill it down to a small number of sampling steps; at deployment only the student is retained, yielding an order-of-magnitude reduction in denoising steps with acceptable generation quality.

\paragraph{\textbf{FP8 Quantization.}}
The second axis of compression targets per-step compute and memory cost rather than step count, we adopt numerical quantization to the dominant matrix multiplications of the distilled student into 8-bit floating-point precision.
FP8 is uniquely well-suited to this setting—its exponent-mantissa split preserves the dynamic range required by activations with heavy-tailed distributions, where integer formats such as INT8 typically incur substantial accuracy loss even elaborate calibration or outlier handling.
Modern accelerators amplify this benefit at the hardware level, by reusing the same Tensor Core datapath at reduced precision, FP8 matrix multiplications achieve roughly 2x throughput over BF16 on current-generation GPUs, while simultaneously halving the bandwidth pressure of weight loading cost that often dominates inference latency at small batch sizes typical of robotic control loops.
Concretely, we apply \emph{post-training quantization} (PTQ) in FP8 with per-block scaling, weights and activations are partitioned into fixed-size blocks along the reduction dimension, each assigned an independent scaling factor that absorbs local magnitude variation while keeping the per-tensor metadata overhead negligible.
This granularity strikes a favorable trade-off between quantization fidelity and runtime efficiency—coarser tensor-wise scaling fails to track the channel-wise outliers characteristic of transformer activations, while finer per-element schemes introduce dequantization overhead that erodes the throughput advantage FP8 is intended to deliver. 

A naive implementation of per-block FP8 quantization can erase most of these throughput gains, since the cost of standalone computation of per-block scales and casting tensors at runtime is non-trivial relative to the GEMM itself.
We therefore implement the quantization operators carefully along two fronts,weights are quantized \emph{offline} into pre-packed FP8 tensors with their scaling factors baked in, so the runtime path bears no weight-side quantization cost, and on-the-fly \emph{activation} quantization is fused into the epilogue of the preceding operator, computing the per-block scale and casting to FP8 within the same kernel that produced the activations—avoiding a separate read/write pass over the tensor and reducing the quantization overhead to a negligible fraction of overall GEMM time.

Combined with general-purpose deployment optimizations such as CUDA Graph capture to eliminate host-side launch overhead in the per-step critical path, the full stack of optimizations described above, brings end-to-end inference to \textbf{10Hz}, meeting the latency budget required for closed-loop robotic control.

\section{Experiments}
\label{sec:exp}

\subsection{Embodied Video Generation Evaluation}
\noindent\textbf{Evaluation Protocol.}
To evaluate video-generation capability in embodied settings, we assess WALL-WM as an \emph{embodied video generator}. Given an initial observation and a language instruction, the model must predict the subsequent event while preserving object identity, robot motion, contact dynamics, and cross-view geometric consistency, so that the generated sequence remains useful for downstream control.
Unless otherwise specified, WALL-WM uses a video tower initialized from Wan2.2-5B, while the Wan-series baselines are evaluated without WALL-WM's event-centric embodied training.
Unlike conventional video-generation benchmarks that focus primarily on perceptual fidelity, evaluating embodied world models must reflect practical value for real embodied applications. We therefore follow the WorldArena evaluation protocol~\cite{shang2026worldarena} to test whether the model has evolved from a generic video prior into an embodied world model prior.

\noindent\textbf{Benchmark construction.}
We construct an Embodied Video Generation benchmark from our generalized embodied data mixture.
The held-out split contains 200 representative in-distribution tasks and 50 out-of-distribution tasks, covering diverse verbs, object categories, scene layouts, camera configurations, and robot embodiments.
The OOD split stresses text and compositional generalization through novel object--verb pairings, paraphrased instructions, unseen scene arrangements, and task compositions that share high-level language with the training set yet require distinct visual trajectories.
Overall, this protocol tests whether the model learns an event-centric mapping from language and observation to future physical states under the next-event prediction paradigm.

\begin{table*}[ht]
\centering
\small
\setlength{\tabcolsep}{3pt}
\renewcommand{\arraystretch}{1.1}

\resizebox{\textwidth}{!}{
\begin{tabular}{l|cc|ccc|ccc|cccc}
\toprule
\multirow{2}{*}{Models} 
& \multicolumn{2}{c|}{Visual Quality}
& \multicolumn{3}{c|}{Motion Quality} 
& \multicolumn{3}{c|}{Semantic Consistency} 
& \multicolumn{4}{c}{Physical Plausibility} \\
\cmidrule(lr){2-3}
\cmidrule(lr){4-6}
\cmidrule(lr){7-9}
\cmidrule(lr){10-13}

& \begin{tabular}[c]{@{}c@{}}Image\\Quality\end{tabular}
& \begin{tabular}[c]{@{}c@{}}Aesthetic\\Quality\end{tabular}
& \begin{tabular}[c]{@{}c@{}}Dynamic\\Degree\end{tabular}
& \begin{tabular}[c]{@{}c@{}}Flow\\Score\end{tabular}
& \begin{tabular}[c]{@{}c@{}}Motion\\Smoothness\end{tabular}
& \begin{tabular}[c]{@{}c@{}}Subject\\Consist.\end{tabular}
& \begin{tabular}[c]{@{}c@{}}Background\\Consist.\end{tabular}
& \begin{tabular}[c]{@{}c@{}}Semantic\\Alignment\end{tabular}
& \begin{tabular}[c]{@{}c@{}}Interaction\\Quality\end{tabular}
& Perspective
& \begin{tabular}[c]{@{}c@{}}Instruction\\Following\end{tabular}
& \begin{tabular}[c]{@{}c@{}}Trajectory\\Acc.\end{tabular} \\
\midrule

Wan2.1-1.3B~\cite{wan2025wan}
& \textbf{0.577} & 0.389 
& 0.199 & 0.061 & 0.619
& 0.476 & 0.522 & 0.857
& 0.219 & 0.819 & 0.308 & 0.214 \\

Wan2.2-5B~\cite{wan2025wan}
& 0.527 & \textbf{0.409} 
& 0.418 & 0.109 & 0.683
& 0.769 & 0.817 & 0.805
& 0.226 & 0.807 & 0.298 & 0.223 \\

\textbf{WALL-WM}
& 0.503 & 0.393
& \textbf{0.484} & \textbf{0.148} & \textbf{0.771}
& \textbf{0.795} & \textbf{0.838} & \textbf{0.886}
& \textbf{0.434} & \textbf{0.821} & \textbf{0.391} & \textbf{0.234} \\

\bottomrule
\end{tabular}
}

\caption{Quantitative comparisons with foundational video generation models.
Our method improves the embodied-relevant metrics of Motion Quality, Semantic Consistency, and Physical Plausibility while preserving competitive visual quality.
The best performing metrics are highlighted in bold.}
\label{tab:video_eval}
\end{table*}

\noindent\textbf{Main results.}
As shown in Table~\ref{tab:video_eval}, WALL-WM consistently outperforms the Wan-series baselines (Wan2.1 and Wan2.2) on embodied-relevant dimensions, including Motion Quality, Semantic Consistency, and Physical Plausibility.
These gains suggest that large-scale embodied training, together with our training recipe, turns the inherited video prior into a stronger physical prior that preserves coherent motion and contact dynamics during generation.
Beyond competitive perceptual quality, WALL-WM leads on physically grounded and interaction-oriented criteria, indicating tighter alignment with realistic manipulation dynamics.
Qualitatively, generic video models frequently exhibit semantic drift, whereas our embodied world model produces more coherent and goal-consistent rollouts.
\begin{table}[h]
    \centering
    \small
    \setlength{\tabcolsep}{4pt}
    \renewcommand{\arraystretch}{1.08}
    \resizebox{0.6\columnwidth}{!}{
    \begin{tabular}{l cccc}
        \toprule
        Probed Feature & Point Err($\downarrow$) & Depth Err($\downarrow$) & AUC@5 ($\uparrow$) & AUC@30 ($\uparrow$) \\
        \midrule
        DINOv2~\cite{oquab2023dinov2}          & 0.559 & 0.209 & 0.051 & 0.508 \\
        V-JEPA~\cite{bardes2024vjepa}           & 0.439 & 0.214 & 0.076 & 0.619 \\
        CogVideoX~\cite{yang2025cogvideox}    & 0.485 & 0.231 & 0.051 & 0.569 \\
        Aether~\cite{zhu2025aether}          & 0.501 & 0.249 & 0.054 & 0.571 \\
        Open-Sora2.0~\cite{zheng2025open} & 0.391 & 0.196 & 0.096 & 0.643 \\
        WAN2.1-14B~\cite{wan2025wan}       & \underline{0.284} & 0.151 & \underline{0.200} & \textbf{0.736} \\
        \textbf{WALL-WM}                       & \textbf{0.271} & \textbf{0.132} & \textbf{0.210} & \underline{0.727} \\
        \bottomrule
    \end{tabular}
    }
    \caption{\textbf{3D awareness benchmark on CO3Dv2~\cite{reizenstein2021common}.} We evaluate different visual representations using point error, depth error, and AUC metrics. WALL-WM achieves competitive 3D awareness, complementing its strong multi-view consistency observed in the cross-view evaluation.}
    \label{tab:co3dv2}
\end{table}

\begin{figure}[!t]
    \centering
    \includegraphics[width=\linewidth]{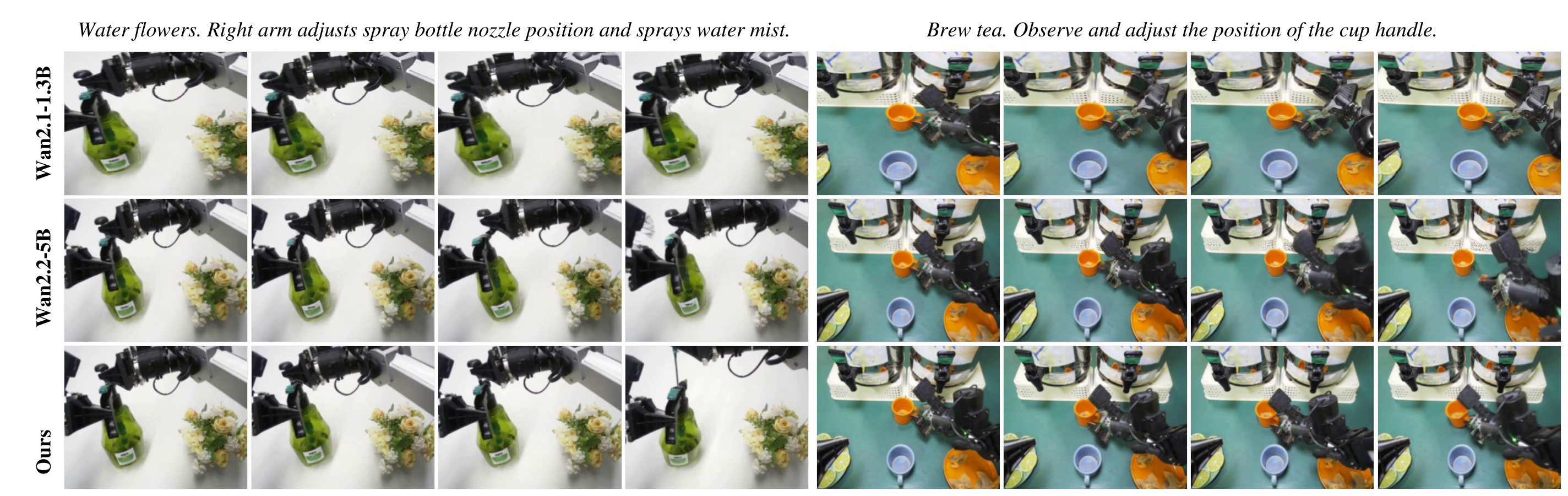}
    \caption{\textbf{Qualitative results.} Our method exhibits better alignment with real-world physical laws, showing more consistent motion and physically plausible interactions, while Wan2.1 and Wan2.2 frequently generate hallucinated objects or unrealistic dynamics (e.g., contact failures, object deformation).}
    
    \label{fig:video_vga}
\end{figure}

\noindent\textbf{Analysis.}
Compared with Wan2.1 and Wan2.2, WALL-WM performs consistently better on our generalized embodied benchmark.
As shown in Fig.~\ref{fig:video_vga}, improvements in Physical Plausibility, Interaction Quality, and Trajectory Accuracy indicate that large-scale embodied training turns the inherited Wan video prior into a physical prior over robot--object interaction and contact evolution.
This effect is enabled by our diverse embodied data mixture, which spans varied tasks, robot configurations, camera views, and language instructions and thus exposes the model to a broader range of embodied dynamics.

WALL-WM also strengthens multi-view collaboration through cross-view attention modules, which perform cross-view information exchange inside the pretrained WAN backbone. Compared with direct multi-view concatenation, this prior-preserving design retains Wan's strengths in texture synthesis and language alignment while injecting synchronized multi-view interaction dynamics. As a result, generated rollouts are more geometrically consistent across views.

\begin{wrapfigure}{r}{0.42\linewidth}
    \vspace{-12pt}
    \centering
    \includegraphics[width=\linewidth]{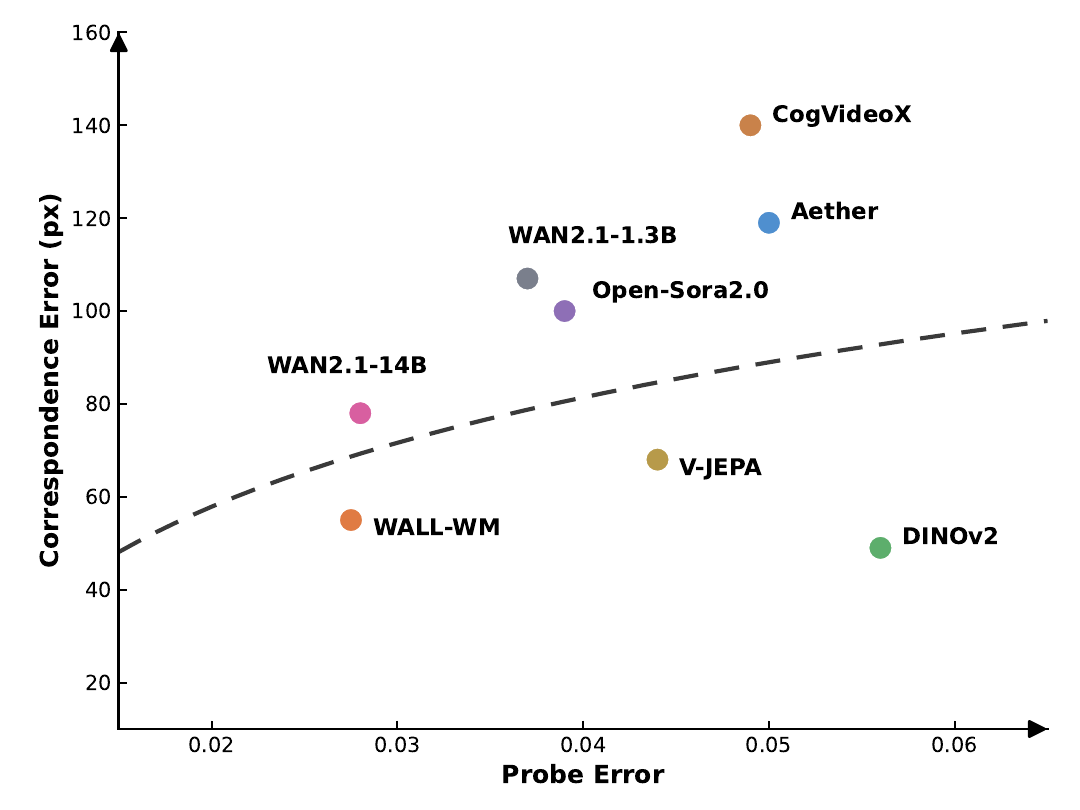}
    \vspace{-18pt}
    \caption{\textbf{Multi-view consistency and 3D awareness.} 3D Probe Error vs.\ Cross-view Correspondence Error (lower is better). WALL-WM achieves jointly low probe and correspondence errors, evidencing 3D-aware and view-consistent representations.}
    \label{fig:multi-view}
    \vspace{-12pt}
\end{wrapfigure}
To quantitatively examine this property, we follow the VidFM-3D\cite{vidfm-3d} evaluation protocol and measure multi-view consistency using cross-view correspondence error, together with 3D awareness metrics based on point and depth probing. As shown in Fig.~\ref{fig:multi-view} and Table~\ref{tab:co3dv2}, WALL-WM achieves strong performance on both multi-view correspondence and 3D-aware probing. In particular, the lower correspondence error indicates better cross-view geometric consistency, while the competitive 3D probe results suggest that the learned representations preserve spatial structure beyond appearance-level generation. These results demonstrate that WALL-WM not only improves visual fidelity and action generation, but also learns view-consistent and 3D-aware dynamics.

Our model also exhibits stronger text generalization.
Wan models often capture broad nouns and verbs, but become less reliable when prompts describe new compositions of familiar manipulation concepts.
WALL-WM is trained with multi-level event captions and prompt augmentation, tying language to local physical phases such as reach, contact, grasp, transfer, and release.
This allows generated sequences to better follow paraphrased or compositional instructions, rather than simply matching familiar canonical descriptions.

\subsection{Real-Robot Evaluation}
\label{sec:exp-real-robot}

\noindent\textbf{Experimental setup.}
The real-robot evaluation is conducted on the high-performance tabletop bimanual robot arms from the internally developed deployment platform suite (Fig.~\ref{fig:x2-robots}).
The platform uses synchronized multi-view observations and language instructions, with evaluation scenes aligned to the tabletop deployment geometry described in Sec.~\ref{sec:data}.
The benchmark is organized into four complementary suites covering direct manipulation, instruction-level reasoning, fine contact control, and generalization in cluttered scenes.
All models are compared using \textit{Task Progress}, a dense 0--100 score that credits partial completion of the specified task according to task-specific rubrics listed in Table~\ref{tab:scoring-rubrics}.
We use Task Progress as the primary metric rather than binary task success because a pure success-rate criterion is often too coarse for real-robot evaluation.
Because task completion is causal and thresholded, binary success can obscure meaningful differences in grounding, contact handling, and intermediate physical reasoning, and thus deviate from the actual competence reflected by the pretrained model.
Task Progress provides a more continuous measure of how far the robot advances toward the specified goal, allowing near-completions, partial state changes, and meaningful intermediate progress to be reflected in the score; the rubric scores are normalized to the 0--100 scale used in the main results.
We therefore report Task Progress uniformly across all real-robot suites to better characterize the practical performance of the pretrained models.
The detailed task-level Task Progress scores for these experiments are recorded in Table~\ref{tab:real-robot-detailed-scores}.
The baselines include $\pi_{0.5}$~\cite{black2025pi_}, DreamZero~\cite{ye2026world}, and LingBot-VA~\cite{li2026causal} when corresponding scores are available.
To ensure a fair comparison, all deployed policies are evaluated under the same physical task definitions, language instructions, multi-view observation streams, scene randomization protocol, and Task Progress rubrics.
Baseline policies are adapted only through their standard action interfaces for the target embodiment, and no method receives privileged state information or task-specific scoring feedback during rollout.
When a baseline could not be deployed on a particular suite under this protocol, we mark the entry as unavailable rather than mixing results from a different evaluation setting.
The main \ours policy is evaluated in \textit{event mode}, where the policy starts from the pretrained event-centric world-action backbone and uses the proposed text reasoning module to provide event-conditioned control.
For ablation, we additionally report \textsc{WALL-WM}-U-Scratch, a from-scratch unified baseline trained directly in a fixed-length instruction-to-action setting on the same real-robot task supervision.
This baseline should not be confused with the unified inference mode introduced in Sec.~\ref{sec:training-inference-unified}, which operates on the event-pretrained backbone; instead, \textsc{WALL-WM}-U-Scratch removes the event-centric pretraining and event-conditioned reasoning pathway to test how much transferable real-robot control prior they provide.

\subsubsection{Diverse Manipulation}
Diverse Manipulation is designed to assess a broad range of direct physical manipulation skills under relatively explicit task goals.
The task set consists of \texttt{Arrange Cup Inverted Triangle}, \texttt{Put Spoon to Bowl}, \texttt{Put Glasses on Woodshelf}, and \texttt{Put Ring onto Rod}.
It further includes \texttt{Put Blocks to Color}, \texttt{Pour Water from Bottle}, and \texttt{Pick Items into Basket}.
Together, these tasks test spatial arrangement, grasp--place coordination, object relocation, tool-object interaction, color-conditioned placement, pouring, and visually guided transfer.
They are not primarily designed to test abstract reasoning; instead, they stress whether the policy can localize small tabletop objects, maintain stable contact transitions, and complete the intended physical state change from multi-view observations.

\begin{figure}[!t]
    \centering
    \includegraphics[width=\linewidth]{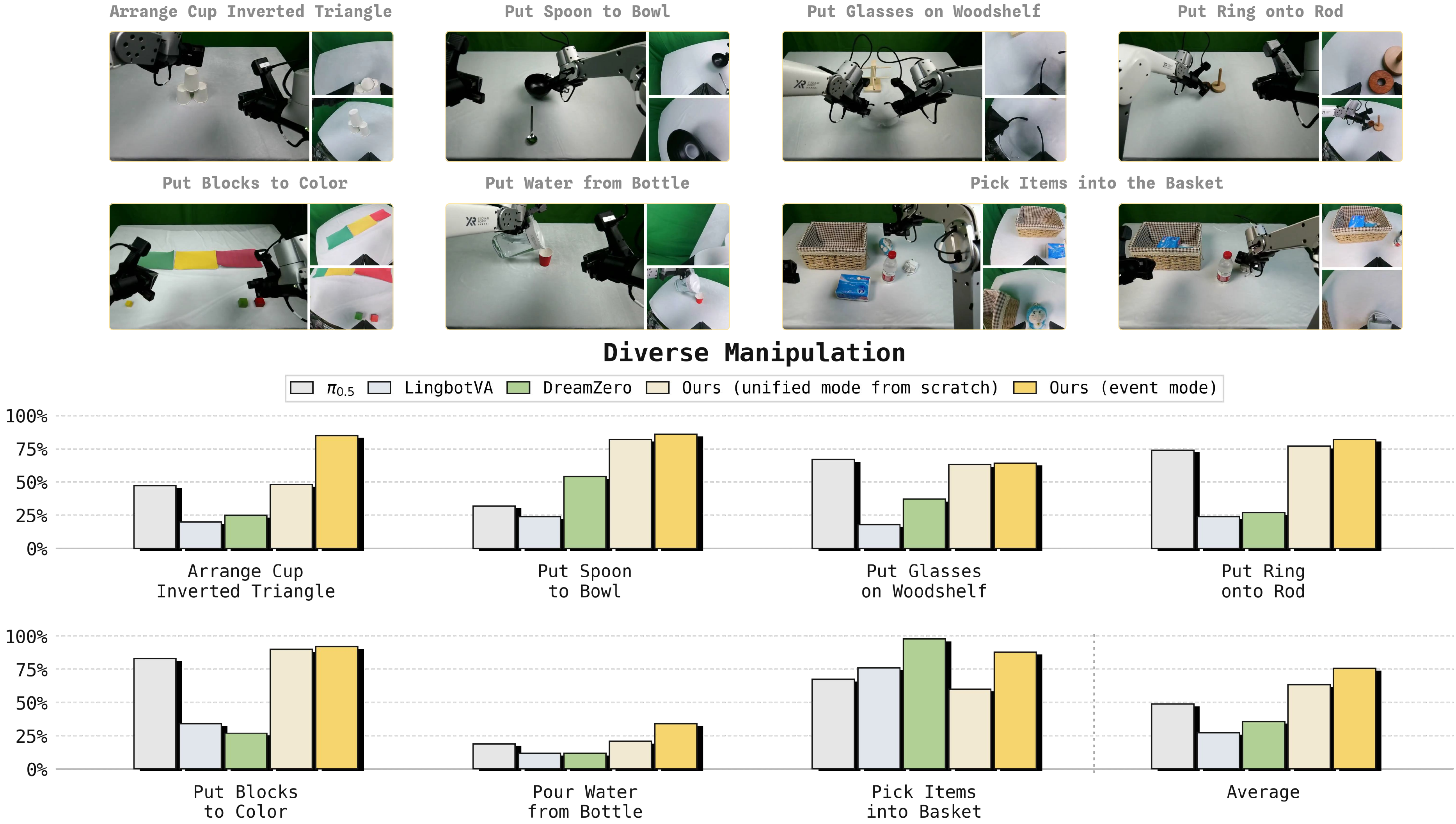}
    \caption{\textbf{Real-robot Diverse Manipulation benchmark.}
    Task Progress scores on direct tabletop manipulation tasks, including grasping, placing, pouring, spatial relocation, and color-conditioned placement.}
    \label{fig:real-bench-basic}
\end{figure}

\noindent\textbf{Results.}
Fig.~\ref{fig:real-bench-basic} reports that event-mode \ours achieves the strongest average Task Progress on the Diverse Manipulation suite, reaching 75.86 compared with 63.00 for \textsc{WALL-WM}-U-Scratch, 55.64 for $\pi_{0.5}$, 39.97 for DreamZero, and 29.71 for LingBot-VA.
The gains are most visible on tasks that require complete object-state transitions rather than a single short reach, such as arranging cups into an inverted triangle, placing a spoon into a bowl, putting a ring onto a rod, moving blocks to target colors, and pouring water from a bottle.
\textsc{WALL-WM}-U-Scratch is also competitive on several direct manipulation tasks, which suggests that the architecture can learn common motor patterns from task supervision alone.
However, event mode improves the average score by 12.86 points over \textsc{WALL-WM}-U-Scratch and remains more consistent across the suite, suggesting that event-centric pretraining may provide a reusable physical prior for grasp, transfer, alignment, and release phases.
The comparison also reveals meaningful task-level variation: $\pi_{0.5}$ slightly leads on shelf placement, and DreamZero obtains the highest score on basket picking.
Thus, the main advantage of event-mode \ours lies not in winning every individual task, but in delivering the strongest overall Task Progress across a diverse set of tabletop manipulation skills.

\subsubsection{Reasoning Manipulation}
Reasoning Manipulation examines instruction grounding beyond single object motion.
The task set consists of \texttt{Sort Headphone}, \texttt{Classify Items as Shape}, and \texttt{Press Button in Order}.
It further includes \texttt{Pair Up Items} and \texttt{Pick Fruits into Basket}.
These tasks require category grounding, relational matching, ordered execution, and instruction-conditioned object selection.
Unlike Diverse Manipulation, the same scene can admit multiple plausible actions, so the policy must infer which object, category, relation, or sequence is intended by the language instruction before executing the corresponding manipulation.

\begin{figure}[!t]
    \centering
    \includegraphics[width=\linewidth]{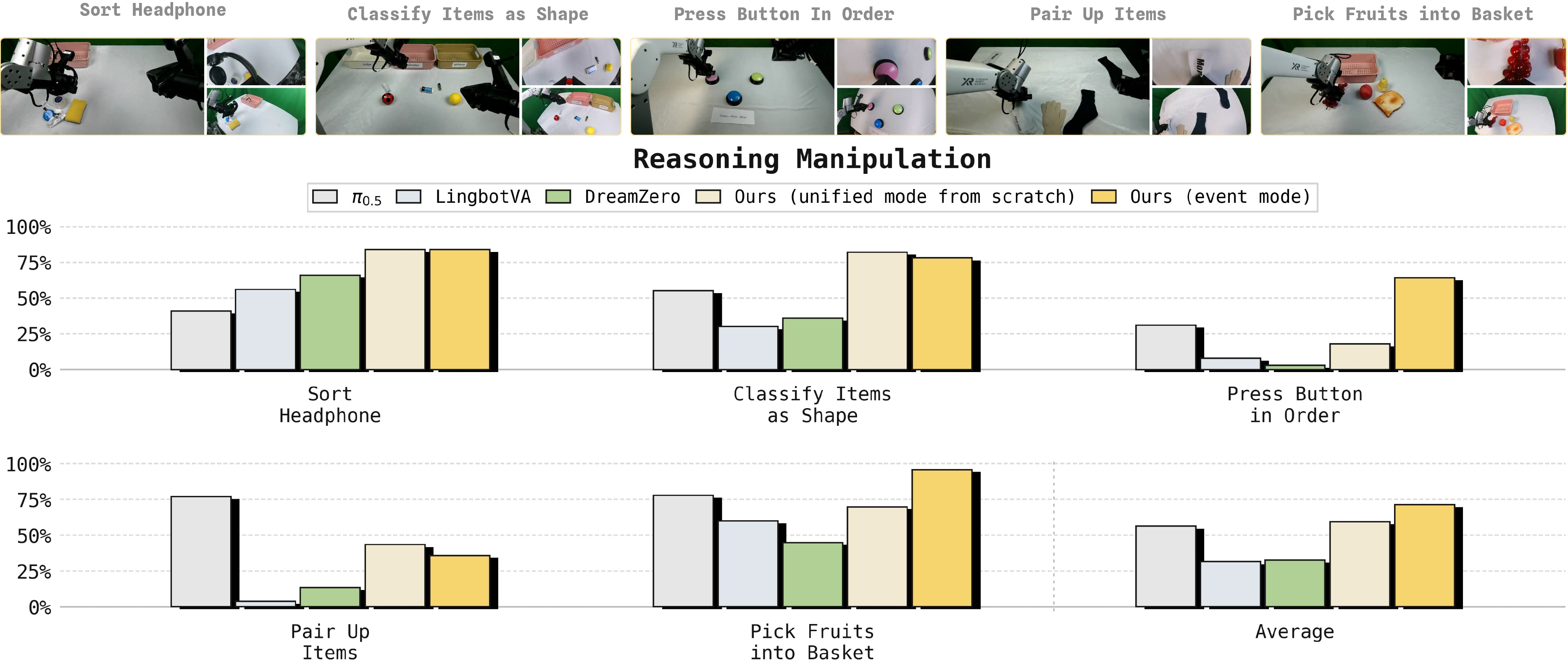}
    \caption{\textbf{Real-robot Reasoning Manipulation benchmark.}
    Task Progress scores on tasks requiring semantic classification, ordering, matching, and instruction-conditioned selection.}
    \label{fig:real-bench-reasoning}
\end{figure}

\noindent\textbf{Results.}
Fig.~\ref{fig:real-bench-reasoning} indicates that event-mode \ours obtains the strongest average result on reasoning-heavy tasks, reaching 71.60 compared with 59.50 for \textsc{WALL-WM}-U-Scratch, 56.40 for $\pi_{0.5}$, 32.70 for DreamZero, and 31.60 for LingBot-VA.
The stable gain over \textsc{WALL-WM}-U-Scratch suggests that the event-mode improvement is not only due to the architecture, but is also associated with the pretrained manipulation backbone and the introduced language-guided reasoning mechanism.
In our real-robot experiments, this mechanism is implemented with a fine-tuned Qwen3.5-VL-9B model that maps broad task requirements and current multi-view observations into concrete next-event descriptions, consistent with the event-mode inference recipe in Sec.~\ref{sec:training-inference-event}.
These next-event descriptions provide explicit event-centric control targets for the pretrained world-action backbone, which may help when the instruction changes the required event structure, such as ordered button pressing and fruit selection.
\textsc{WALL-WM}-U-Scratch remains competitive on recognition-oriented cases such as sorting and classification, and $\pi_{0.5}$ performs best on relational pairing.
Thus, the main advantage of event-mode \ours is its stronger overall balance across category grounding, ordered execution, and instruction-conditioned selection, supported by language-guided conversion from high-level instructions to executable event descriptions.

\subsubsection{Dexterous Manipulation}
Dexterous Manipulation emphasizes contact-sensitive behaviors that are difficult to capture with coarse semantic instructions alone.
The figure visualizes two fine-grained tasks: \texttt{Put Stationery in Case}, and \texttt{Insert Wireline}.
The averaged Task Progress score is reported on the scored insertion and constrained-placement tasks.
These tasks require narrow-tolerance insertion, precise end-effector alignment, careful contact timing, and, in the fan-unboxing example, longer object-handling sequences with packaging constraints.
Small pose errors can prevent progress even when the high-level object selection is correct, making the suite a stress test for local action precision rather than semantic understanding alone.

\begin{figure}[!t]
    \centering
    \includegraphics[width=\linewidth]{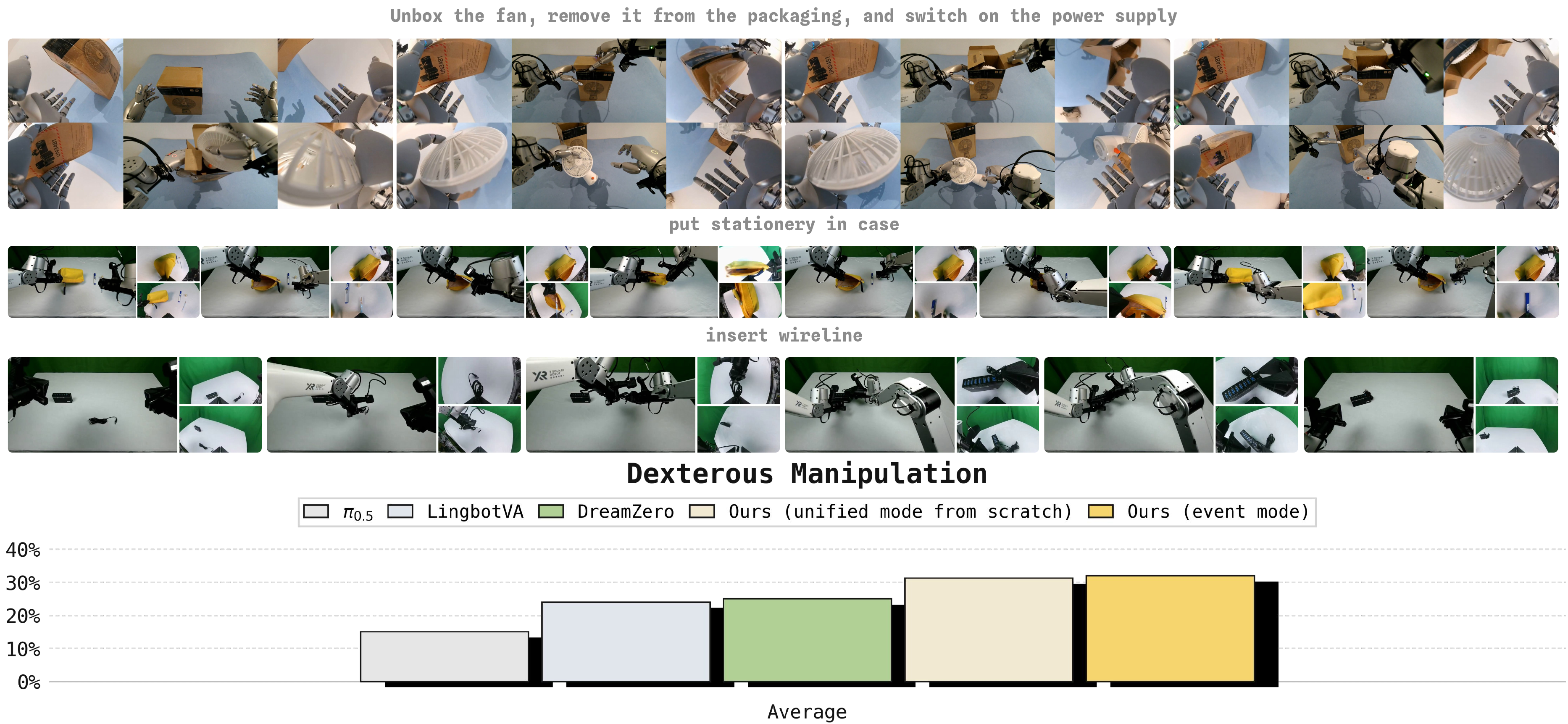}
    \caption{\textbf{Real-robot Dexterous Manipulation benchmark.}
    Average Task Progress on contact-sensitive tasks that require precise insertion and fine object handling.}
    \label{fig:real-bench-dexterous}
\end{figure}

\noindent\textbf{Results.}
Fig.~\ref{fig:real-bench-dexterous} shows that event-mode \ours obtains the highest averaged Task Progress on the Dexterous Manipulation suite, reaching 32.00, closely followed by \textsc{WALL-WM}-U-Scratch at 31.25.
Both modes improve over the averaged scores of $\pi_{0.5}$, LingBot-VA, and DreamZero, suggesting that the pretrained video-action backbone may provide a useful local dynamics prior for contact-rich motion.
However, the gap between event mode and \textsc{WALL-WM}-U-Scratch is small, indicating that these tasks are less limited by high-level event decomposition and more constrained by low-level pose accuracy, contact timing, and narrow-tolerance alignment.
At the task level, both \ours modes achieve the best score on \texttt{Insert Wireline}, while DreamZero performs better on \texttt{Put Stationery in Case}.
Overall, the relatively low absolute scores show that dexterous real-robot manipulation remains highly challenging, especially for precise insertion and constrained placement where small contact errors can prevent task progress.

\subsubsection{Generalization}
Generalization is designed to assess instruction-conditioned transfer in a complex tabletop scene rather than in isolated task-specific resets.
During evaluation, multiple objects remain in the scene and different instructions are issued in randomized order, requiring the policy to ground the current command against the shared scene state.
The tasks include \texttt{Place Plates into Storage Slots}, \texttt{Cover Pot with Lid}, \texttt{Push Cleaning Cloth to Table Edge}, and \texttt{Insert Screwdriver into Cup}.
The setting targets scene-level grounding, instruction switching, robustness to distractors, and compositional transfer.
The robot must avoid overfitting to a fixed task order and instead select the action implied by the current instruction under visual clutter and shared object context.

\begin{figure}[!t]
    \centering
    \includegraphics[width=\linewidth]{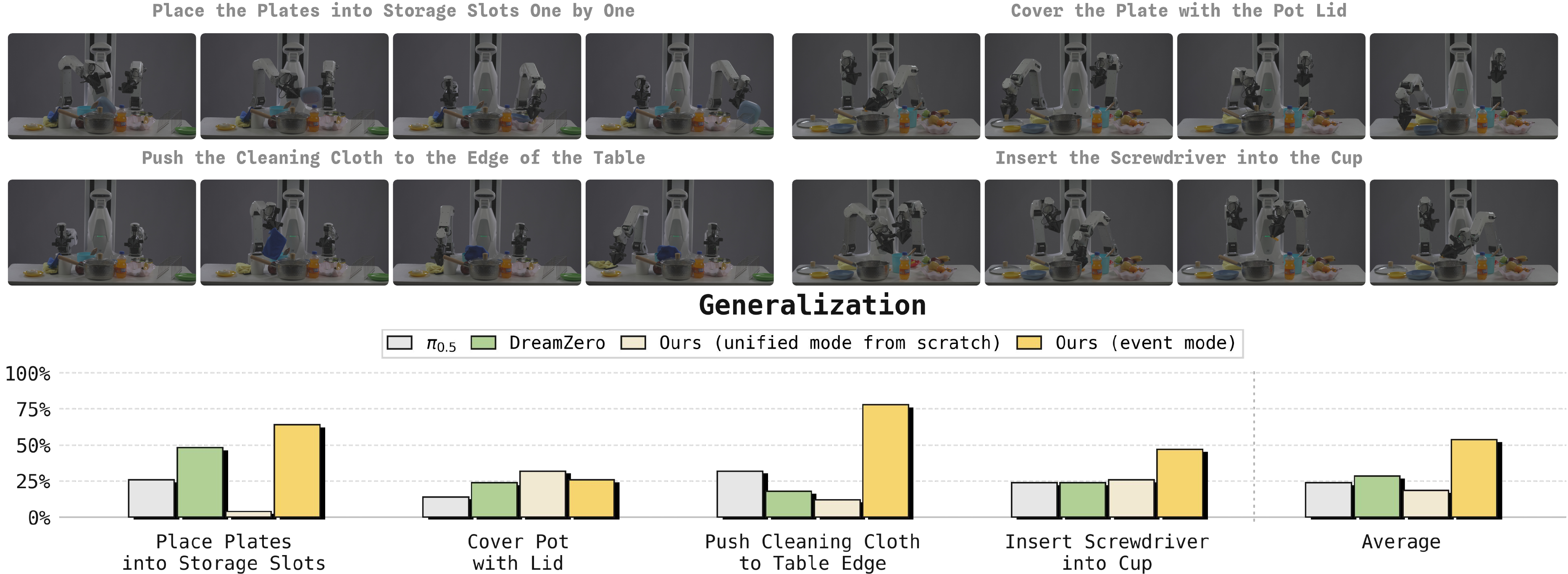}
    \caption{\textbf{Real-robot Generalization benchmark.}
    Task Progress scores in a complex tabletop scene where different instructions are issued in randomized order, testing scene grounding and compositional transfer.}
    \label{fig:real-bench-generalization}
\end{figure}

\noindent\textbf{Results.}
Fig.~\ref{fig:real-bench-generalization} reports that event-mode \ours obtains the highest average Task Progress on the Generalization suite, reaching 53.75 compared with 28.50 for DreamZero, 24.00 for $\pi_{0.5}$, and 18.50 for \textsc{WALL-WM}-U-Scratch.
The improvement is most visible on \texttt{Place Plates into Storage Slots}, \texttt{Push Cleaning Cloth to Table Edge}, and \texttt{Insert Screwdriver into Cup}, where the policy must ground the current instruction among multiple objects and select the corresponding manipulation event.
This trend is consistent with event-mode inference: the language-guided reasoning module converts the current command into a next-event description, while the pretrained world-action backbone executes reusable transitions such as approach, align, insert, push, cover, and release.
At the same time, the results are not uniform across all tasks; \textsc{WALL-WM}-U-Scratch performs better on \texttt{Cover Pot with Lid}, suggesting that some visually direct motions can still be learned from task-level supervision without explicit event decomposition.
Overall, the suite indicates that event-conditioned execution can improve robustness under instruction switching and shared scene context, while generalization in cluttered real-robot settings remains challenging.

\subsubsection{Ablation on Event and View Modeling}
To examine the contribution of event-conditioned execution and cross-view modeling, an additional pretrained baseline is evaluated on the reasoning and generalization suites.
Different from the \textsc{WALL-WM}-U-Scratch baseline reported above, this baseline starts from the pretrained backbone but removes View-Interaction Self-Attention (VI-SA) and uses instruction-conditioned fixed-length unified decoding.
Thus, the comparison controls for the presence of pretraining while removing the two design choices that most directly support multi-view scene grounding and variable-length event execution.
Because the baseline differs from event mode in both VI-SA and the execution format, this ablation should be interpreted as measuring their combined effect rather than assigning the full gain to either component alone.

\begin{table}[!t]
    \centering
    \footnotesize
    \setlength{\tabcolsep}{1pt}
    \renewcommand{\arraystretch}{1.08}
    \resizebox{0.75\linewidth}{!}{
    \begin{tabular}{lcc@{\hspace{2.2em}}lcc}
        \toprule
        \multicolumn{3}{l}{Reasoning Manipulation} &
        \multicolumn{3}{l}{Generalization} \\
        Task & Base & Event & Task & Base & Event \\
        \midrule
        \texttt{Sort Headphone} & 55 & 84 &
        \texttt{Place Plates into Storage Slots} & 30 & 64 \\
        \texttt{Classify Items as Shape} & 40 & 78 &
        \texttt{Cover Pot with Lid} & 18 & 26 \\
        \texttt{Press Button in Order} & 0 & 64 &
        \texttt{Push Cleaning Cloth to Table Edge} & 26 & 78 \\
        \texttt{Pair Up Items} & 8 & 36 &
        \texttt{Insert Screwdriver into Cup} & 14 & 47 \\
        \texttt{Pick Fruits into Basket} & 60 & 96 &
        -- & -- & -- \\
        \midrule
        \textbf{Average} & \textbf{32.6} & \textbf{71.6} &
        \textbf{Average} & \textbf{22.0} & \textbf{53.75} \\
        \bottomrule
    \end{tabular}
    }
    \caption{\textbf{Ablation on event-conditioned execution and VI-SA.}
    Base denotes a pretrained fixed-length unified baseline without View-Interaction Self-Attention, while Event denotes pretrained event-mode \ours. All entries report Task Progress.}
    \label{tab:real-ablation-event-view}
\end{table}

\noindent\textbf{Results.}
Table~\ref{tab:real-ablation-event-view} indicates that pretrained event-mode \ours improves over the pretrained unified baseline without VI-SA on both evaluated suites.
On Reasoning Manipulation, the average Task Progress increases from 32.6 to 71.6.
The baseline retains some recognition capability from pretraining, as reflected by nontrivial scores on sorting and fruit selection, but it is less reliable when the instruction requires explicit temporal ordering or relational execution.
For example, \texttt{Press Button in Order} improves from 0 to 64, while \texttt{Pair Up Items} improves from 8 to 36, although the latter remains challenging in absolute terms.
On Generalization, the average score increases from 22.0 to 53.75.
The gains are most visible on tasks such as \texttt{Push Cleaning Cloth to Table Edge} and \texttt{Insert Screwdriver into Cup}, where the robot must identify the instruction-relevant object and execute the corresponding semantic transition in a shared scene.
Overall, these results suggest that pretrained representations alone are not sufficient for these settings, and are consistent with the combined benefit of cross-view interaction and event-conditioned execution for robust reasoning and generalization.

\noindent\textbf{Summary.}
Across the four real-robot suites, event-mode \ours achieves the strongest overall Task Progress.
The Diverse Manipulation results suggest that event-centric pretraining can supply a useful physical control prior for common tabletop skills, while the Reasoning Manipulation results indicate that coupling this prior with a text reasoning module may improve instruction grounding for category, order, and relation-based tasks.
The Dexterous Manipulation suite highlights the remaining difficulty of fine contact control, while also suggesting that video-action pretraining can benefit local contact-sensitive motion.
Finally, the Generalization suite indicates that \ours can reuse pretrained event knowledge under cluttered scenes and randomized instructions, where \textsc{WALL-WM}-U-Scratch is less reliable.
The event-and-view ablation is consistent with the view that pretraining is most effective when paired with cross-view interaction and variable-length event execution, rather than used only as a fixed-length instruction-conditioned policy.
The consistent gap between event mode and \textsc{WALL-WM}-U-Scratch supports the hypothesis that pretrained event-centric world-action modeling provides transferable physical and semantic priors that are difficult to recover from task-level real-robot supervision alone.

\FloatBarrier

\section{Discussion}

\paragraph{Scaling, latency, and the objective of generalization.}
Our current WALL-WM family spans model sizes from below 10B parameters to the tens-of-billions regime. Across this range, we observe a consistent trend: larger models improve both action precision and out-of-distribution generalization, especially on tasks that require fine contact timing, long-horizon state tracking, and compositional instruction grounding. This trend raises a broader question for embodied foundation models. The field has often treated real-time latency as a first-order constraint, because robotic deployment ultimately requires closed-loop execution. However, if latency is enforced too early, it may cap the reachable performance ceiling before the model has learned a sufficiently general world-action prior. We therefore view scale-driven generalization as the primary frontier for general-purpose WAMs, while treating latency reduction as a comparatively more determinate engineering problem. Distillation, quantization, speculative or streaming execution, and systems-level overlap can progressively reduce latency once a strong model exists; recovering a lost generalization ceiling from an under-scaled model is much less straightforward.

\paragraph{Beyond success rate as the sole evaluation target.}
Real-robot success rate remains the most direct measure of embodied competence, but it is also a coarse and noisy metric. A single failure can be caused by perception, language grounding, contact dynamics, hardware variation, reset quality, or stochastic environment conditions, and these causes are not uniformly distributed across tasks or evaluation runs. As a result, success rate alone has limited resolution for large-scale pretraining: it is expensive, slow, and often weak at identifying which part of the model improved. We are therefore exploring more efficient evaluation protocols for pretraining-scale iteration, including finer-grained per-event scoring, dense progress measures, human correction and relabeling during evaluation, and general-purpose evaluation models that predict robot performance from generated video, action traces, and intermediate states. Such an evaluator does not need to perfectly replace real hardware. It only needs to maintain a reliable positive correlation with downstream robot performance, so that model and data decisions can be screened cheaply before the most promising candidates are validated on real robots.

\paragraph{Event-centric instruction generalization as an interface to agents.}
We view per-event textual generalization as a bridge between embodied control and broader agentic intelligence. Current embodied foundation models often spend trainable parameters on understanding and reasoning over high-level instructions, but the rapid progress of frontier multimodal models and agents~\cite{openai_gpt5,comanici2025gemini,claude37} forces a boundary question: should a high-frequency manipulation model compete for parameters on over-high-level task reasoning that a general agent can already perform? In many embodied benchmark designs, this boundary remains underspecified. Our preference is to factor information from high-level tasks into per-event textual descriptions, and then from per-event language into vision and action. Under this decomposition, full-domain action-text instruction generalization over semantic events becomes the highest-priority objective for the embodied model, while excessive concern over the reasoning cost or latency of the upstream agent may itself become a bottleneck to the emergence of embodied intelligence.

\paragraph{When fixed-length inference can be the faster practical choice.}
Event-centric pretraining provides a natural and scalable training objective for WAMs regardless of the final inference interface, because it aligns language, video, and action at the same semantic unit and produces a reusable world-action prior. This does not imply that event-mode inference is always the best short-term deployment choice. In settings with only a small number of tasks, fixed instructions for each task, limited need for out-of-distribution generalization, and scarce training data, a fixed-length inference policy can sometimes reach its local optimum faster than direct event-mode rollout. In such regimes, the advantage of explicit event decomposition may be muted because the deployment distribution is narrow and the instruction-action mapping is nearly deterministic. The intended role of event-centric modeling is therefore not to replace all fixed-horizon execution, but to provide a stronger foundation from which both event-mode and fixed-length unified policies can be derived.

\paragraph{Evaluation advantages from platform alignment and tuning resources.}
Our real-robot evaluation is conducted on an internally developed embodiment suite, and the \ours models receive large-scale pretraining on data collected from or aligned with this platform. This gives our system an evaluation-environment advantage that cannot be fully removed by sharing task definitions, observations, instructions, and scoring rubrics across baselines. We also emphasize that the empirical upper bound of a robot foundation model is determined not only by the high-level method and dataset scale, but also by detailed hyperparameter choices, data filtering, loss balancing, deployment adaptation, and many low-level training and systems decisions. Under finite compute and engineering budgets, comparisons against resource-intensive systems such as LingBot-VA and DreamZero inevitably involve unequal amounts of method-specific tuning.

\paragraph{Future Work}
The current data construction recipe still relies on large-scale temporal grounding and fine-grained captions to expose event structure before training. A central next direction is self-supervised pretraining over vision, language, and action, where event boundaries are not provided as dense annotations but are captured by the training objective itself. In parallel, distillation can partially reduce the difficulty of learning latent-level future prediction from scratch, while redefining the autoregressive unit from fixed tokens or chunks to event-level latent primitives may support a next-generation WAM pretraining paradigm. Combined with deeper infrastructure optimization, the goal is to make pretraining more efficient and to move part of the annotation burden into scalable training computation, reducing dependence on hand-crafted temporal labels while preserving physical grounding.

% \newpage
\section*{Contributors}
\label{sec:contributors}

WALL-WM is a collaborative effort of the X Square Robot team. The full contributor list is given below; $^{\ast}$ denotes core contributors, $^{\dagger}$ denotes the project lead, and $^{\ddagger}$ denotes the corresponding author.

\vspace{0.5em}
\noindent
Shalfun Li$^{\ast\dagger}$, Victor Yao$^{\ast}$, Charles Yang$^{\ast}$, Truth Qu$^{\ast}$, Regis Cheng$^{\ast}$, Ryan Yu$^{\ast}$, Howard Lu$^{\ast}$, Newton Von$^{\ast}$, Vincent Chen$^{\ast}$, Yohann Tang, Maeve Zhang, Ellie Ma, Gody Li, Starrick Liu, Sage Yang, Lorien Shu, J.W. Gao, Ethan Chen, Colin Ye, Yu Sun, Elise Mon, PS Zhang, Neo Li, Lily Li, James Wang, Ping Yang, Chris Pan, Lucy Liang, Hang Su, Roy Gan, Hao Wang$^{\ddagger}$, Qian Wang.

\clearpage
\section{Appendix}
\label{sec:analysis}

\subsection{Design Philosophy: Pixel-Space, Native T2V, and a Capacity-Balanced Dual-Tower}
\label{sec:analysis-design-axes}
\label{sec:method-design-axes}

Three design principles, taken together, place \ours\ in the design space.
The recent VLA literature tends to collapse them into a single ``pixel vs.\ latent'' question, but they are better read as three distinct trade-offs with separate logic, and \ours\ takes the more demanding side on each axis.
We discuss the three in turn.

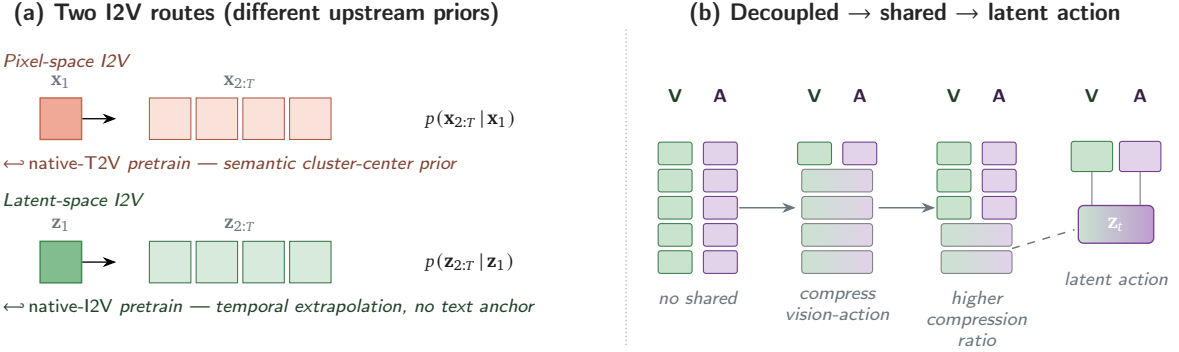
\begin{figure*}[!htbp]
    \centering
    \definecolor{vlavidcol}{HTML}{E87A5D}
    \definecolor{vlatextcol}{HTML}{4DA060}
    \definecolor{vlaactcol}{HTML}{8A4BAA}
    \definecolor{peCol}{HTML}{6C757D}
    \resizebox{0.95\textwidth}{!}{
    \begin{tikzpicture}[
        font=\sffamily\footnotesize,
        rawfst/.style={draw=vlavidcol!85!black, fill=vlavidcol!65, line width=0.55pt,
                       minimum width=0.55cm, minimum height=0.55cm, inner sep=0pt},
        rawftr/.style={draw=vlavidcol!75!black, fill=vlavidcol!22, line width=0.4pt,
                       minimum width=0.55cm, minimum height=0.55cm, inner sep=0pt},
        latfst/.style={draw=vlatextcol!85!black, fill=vlatextcol!70, line width=0.55pt,
                       minimum width=0.55cm, minimum height=0.55cm, inner sep=0pt},
        latftr/.style={draw=vlatextcol!75!black, fill=vlatextcol!22, line width=0.4pt,
                       minimum width=0.55cm, minimum height=0.55cm, inner sep=0pt},
        Vblk/.style={draw=vlatextcol!80!black, fill=vlatextcol!30, line width=0.4pt,
                     rounded corners=1pt,
                     minimum width=0.45cm, minimum height=0.30cm, inner sep=0pt},
        Ablk/.style={draw=vlaactcol!80!black, fill=vlaactcol!25, line width=0.4pt,
                     rounded corners=1pt,
                     minimum width=0.45cm, minimum height=0.30cm, inner sep=0pt},
        shared/.style={draw=black!60, dashed, line width=0.45pt, rounded corners=2pt,
                       fill=peCol!8, inner sep=0pt},
        zcode/.style={draw=vlaactcol!85!black,
                      shade, left color=vlatextcol!30, right color=vlaactcol!55,
                      line width=0.5pt,
                      rounded corners=2pt,
                      minimum width=1.0cm, minimum height=0.50cm, inner sep=0pt,
                      font=\sffamily\footnotesize\bfseries, text=white},
        capL/.style={font=\sffamily\small\bfseries, anchor=south, text=black!85},
        sub/.style={font=\sffamily\scriptsize\itshape, text=peCol},
        ar/.style={-{Stealth[length=1.8mm]}, line width=0.5pt},
    ]
        \begin{scope}
            \node[capL] at (3.1, 3.55)
                {(a) Two I2V routes (different upstream priors)};

            \node[sub, anchor=west, text=vlavidcol!55!black]
                at (-0.40, 3.20) {Pixel-space I2V};

            \node[rawfst] (x1) at (0.50, 2.45) {};
            \node[sub, anchor=south] at (x1.north) {$\mathbf{x}_1$};
            \draw[ar] (x1.east) -- ++(0.45,0);
            \foreach \j in {0,1,2,3}{
                \node[rawftr] (xT\j) at ({1.95+0.62*\j}, 2.45) {};
            }
            \node[sub, anchor=south]
                at ($(xT0.north)!0.5!(xT3.north)$) {$\mathbf{x}_{2:T}$};

            \node[anchor=west, font=\sffamily\scriptsize, text=black!90]
                at (5.20, 2.45) {$p(\mathbf{x}_{2:T}\!\mid\!\mathbf{x}_1)$};

            \node[sub, anchor=west, text=vlavidcol!55!black]
                at (-0.40, 1.85)
                {$\hookleftarrow$\,\emph{native-T2V} pretrain --- semantic cluster-center prior};

            \node[sub, anchor=west, text=vlatextcol!45!black]
                at (-0.40, 1.35) {Latent-space I2V};

            \node[latfst] (z1) at (0.50, 0.55) {};
            \node[sub, anchor=south] at (z1.north) {$\mathbf{z}_1$};
            \draw[ar] (z1.east) -- ++(0.45,0);
            \foreach \j in {0,1,2,3}{
                \node[latftr] (zT\j) at ({1.95+0.62*\j}, 0.55) {};
            }
            \node[sub, anchor=south]
                at ($(zT0.north)!0.5!(zT3.north)$) {$\mathbf{z}_{2:T}$};

            \node[anchor=west, font=\sffamily\scriptsize, text=black!90]
                at (5.20, 0.55) {$p(\mathbf{z}_{2:T}\!\mid\!\mathbf{z}_1)$};

            \node[sub, anchor=west, text=vlatextcol!40!black]
                at (-0.40, -0.05)
                {$\hookleftarrow$\,\emph{native-I2V} pretrain --- temporal extrapolation, no text anchor};
        \end{scope}

        \draw[densely dotted, peCol!55, line width=0.5pt]
            (8.0, -0.55) -- (8.0, 3.65);

        \begin{scope}[xshift=8.4cm]
            \node[capL] at (3.3, 3.55)
                {(b) Decoupled $\to$ shared $\to$ latent action};

            \def\sOne{0.55}
            \foreach \i in {0,1,2,3,4}{
                \node[Vblk] at (\sOne-0.30, 0.55+0.36*\i) {};
                \node[Ablk] at (\sOne+0.30, 0.55+0.36*\i) {};
            }
            \node[sub, anchor=north] at (\sOne, 0.30) {no shared};
            \node[font=\sffamily\scriptsize\bfseries,
                  text=vlatextcol!40!black, anchor=south]
                at (\sOne-0.30, 2.55) {V};
            \node[font=\sffamily\scriptsize\bfseries,
                  text=vlaactcol!40!black, anchor=south]
                at (\sOne+0.30, 2.55) {A};

            \draw[ar, line width=0.55pt, peCol]
                (\sOne+0.55, 1.27) -- (2.40-0.55, 1.27);

            \def\sTwo{2.40}
            \node[Vblk] at (\sTwo-0.30, 0.55+0.36*4) {};
            \node[Ablk] at (\sTwo+0.30, 0.55+0.36*4) {};
            \foreach \i in {0,1,2,3}{
                \pgfmathsetmacro{\yy}{0.55+0.36*\i}
                \path[
                    shade,
                    left color=vlatextcol!30,
                    right color=vlaactcol!25,
                    draw=black!50,
                    rounded corners=1.2pt,
                    line width=0.4pt
                ]
                    (\sTwo-0.47,\yy-0.15) rectangle (\sTwo+0.47,\yy+0.15);
            }
            \node[sub, anchor=north, text width=1.7cm, align=center]
                at (\sTwo, 0.30) {compress\\vision-action};
            \node[font=\sffamily\scriptsize\bfseries,
                  text=vlatextcol!40!black, anchor=south]
                at (\sTwo-0.30, 2.55) {V};
            \node[font=\sffamily\scriptsize\bfseries,
                  text=vlaactcol!40!black, anchor=south]
                at (\sTwo+0.30, 2.55) {A};

            \draw[ar, line width=0.55pt, peCol]
                (\sTwo+0.55, 1.27) -- (4.25-0.55, 1.27);

            \def\sThree{4.25}
            \foreach \i in {2,3,4}{
                \node[Vblk] at (\sThree-0.30, 0.55+0.36*\i) {};
                \node[Ablk] at (\sThree+0.30, 0.55+0.36*\i) {};
            }
            \foreach \i in {0,1}{
                \pgfmathsetmacro{\yy}{0.55+0.36*\i}
                \path[
                    shade,
                    left color=vlatextcol!30,
                    right color=vlaactcol!25,
                    draw=black!50,
                    rounded corners=1.2pt,
                    line width=0.4pt
                ]
                    (\sThree-0.47,\yy-0.15) rectangle (\sThree+0.47,\yy+0.15);
            }
            \node[sub, anchor=north, text width=1.7cm, align=center]
                at (\sThree, 0.30) {higher\\compression ratio};
            \node[font=\sffamily\scriptsize\bfseries,
                  text=vlatextcol!40!black, anchor=south]
                at (\sThree-0.30, 2.55) {V};
            \node[font=\sffamily\scriptsize\bfseries,
                  text=vlaactcol!40!black, anchor=south]
                at (\sThree+0.30, 2.55) {A};

            \draw[dashed, line width=0.55pt, black!55]
                (\sThree+0.47, 0.73) -- (6.10-0.55, 0.98);

            \def\sFour{6.10}
            \node[Vblk, minimum width=0.55cm, minimum height=0.40cm]
                (V4) at (\sFour-0.32, 1.95) {};
            \node[Ablk, minimum width=0.55cm, minimum height=0.40cm]
                (A4) at (\sFour+0.32, 1.95) {};
            \node[font=\sffamily\scriptsize\bfseries,
                  text=vlatextcol!40!black, anchor=south]
                at (\sFour-0.32, 2.55) {V};
            \node[font=\sffamily\scriptsize\bfseries,
                  text=vlaactcol!40!black, anchor=south]
                at (\sFour+0.32, 2.55) {A};

            \node[zcode] (zStar) at (\sFour, 1.05) {$\mathbf{z}_t$};
            \draw[line width=0.4pt, black!55] (V4.south) -- (zStar.north -| V4.south);
            \draw[line width=0.4pt, black!55] (A4.south) -- (zStar.north -| A4.south);
            \node[sub, anchor=north] at (\sFour, 0.55) {latent action};
        \end{scope}
    \end{tikzpicture}}
    \caption{\textbf{Two modeling-choice axes that place \ours\ in the design space.}
    \emph{(a) Pretraining-route axis.}
    Pixel-space (top, orange) vs.\ latent-space (bottom, green) backbones both perform the same downstream I2V job, but inherit qualitatively different upstream priors---a \emph{native-T2V} cluster-center prior on the pixel side, a denser \emph{native-I2V} temporal-extrapolation prior on the latent side.
    \ours\ keeps the pixel\,$+$\,native-T2V combination (\Cref{sec:method-design-axes}).
    \emph{(b) Dual-tower spectrum.}
    This is an abstract illustration rather than a concrete architectural sweep: the latent video-action representation is implicitly distributed across the DiT parameters, and under certain capacity-sharing regimes, intermediate features of the coupled video-action DiT become operationally equivalent to the narrow notion of a latent action.}
    \label{fig:modeling-choices}
\end{figure*}

\emph{(i) Pretrained pixel-space prior vs.\ pretrained latent-space prior.}
The first axis is largely empirical: pixel-space video foundation models have, by now, been pretrained on far broader and far thicker corpora than any latent-space counterpart available off the shelf, and the resulting weights---redundant as they often are---encode a strong prior over the visual world.
Setting aside this historical asymmetry, the latent side still carries an intrinsic advantage worth naming: by inducing a smoother and more structured representation manifold, latent training reduces reliance on stochastic score-based generation and leaves a flatter, more tractable optimization landscape behind.
For a video$\,+\,$action world model whose downstream burden is dynamics and control, however, the sheer breadth of the pixel-space pretraining prior dominates: initialising from a pixel-space backbone is therefore a pragmatic choice rather than a stance against latent representations.

\emph{(ii) Native T2V vs.\ native I2V, even when both are trained from scratch.}
The second axis is easy to miss but consequential once noticed: most of today's pixel-space video foundation models---including those formally trained under an I2V objective---are themselves built on top of an upstream \emph{native-T2V} pretraining map, while a separate line of work (e.g.\ V-JEPA-2~\cite{assran2025v}, LeWorldModel~\cite{maes2026leworldmodel}) starts directly from \emph{native-I2V} mappings.
Neither route admits a clean winner; the relevant trade-off is the following.
A native-I2V objective tracks the temporal evolution of the visual signal more tightly than text-to-vision ever can---given the previous frame, the visual continuation is densely supervised at every pixel---which is a genuine strength.
A native-T2V objective, in turn, brings in a prior of a different flavour.
Text behaves like a low-dimensional \emph{cluster-center label} over the high-dimensional manifold of plausible visual futures: a T2V objective therefore implicitly asks the network to first commit to the semantic class of the trajectory it is about to render and only then paint it, which is a form of semantically anchored visual self-supervision delivered for free at internet scale.
I2V removes that anchor.
With nothing to commit to up front, the easiest minimum of the I2V loss is high-bandwidth pixel-extrapolation conditioned on the previous frame---closer in spirit to a learned optical flow than to a world model---and any gradient that would teach high-level physical regularities flows through that route only very thinly.
Higher-level structure (intent, object identity, sub-task boundaries) is correspondingly only weakly captured even when scaling is generous.
WALL-WM remains on the T2V side, so that this cluster-center prior is paid for once and inherited by everything downstream (\Cref{fig:modeling-choices}a).

\emph{(iii) Latent-action vs.\ dual-tower video$\,+\,$action.}
The two paradigms appear opposed at first sight: one compresses dynamics into an explicit, often discrete, action code, while the other places two DiT-scale towers side by side.
On closer inspection, however, they converge.
Latent-action methods explicitly bottleneck the next observation into a vision-aligned, implicit dynamic-action code that the downstream policy must decode; a dual-tower video$\,+\,$action design, once the ratio between cross-tower shared capacity (the layer-wise coupling \textsf{CA}, AdaLN modulation, the shared frame-index PE) and tower-private capacity (the per-tower FFN width, the action-only state cross-attention) settles at the right operating point, performs essentially the same compression.
The shared subspace \emph{is} a latent action, except that it is emergent and end-to-end learned rather than imposed through an architectural bottleneck whose width and codebook have to be guessed in advance.
The dual-tower formulation is therefore better read as a more permissive form of latent action than as its competitor: tightening the shared sub-block recovers any explicit-bottleneck regime, while loosening it leaves room for extra width when the true dynamics outgrow any latent dimensionality picked by hand.
Formally, let $\mathbf{h}^{\text{shared}}_t\!\in\!\mathbb{R}^d$ denote the cross-tower bottleneck activation through which V information reaches the A tower, and let $\mathbf{z}_t$ be the explicit latent-action code of an LAPA-style encoder/codebook/decoder; when $d$ matches the codebook width, the two carry the same information bottleneck and are interchangeable up to the discreteness of $\mathbf{z}_t$ (\Cref{fig:modeling-choices}b).

Together these three choices place \ours\ on a pixel-space initialization, a native-T2V objective, and a dual-tower with deliberately sized shared and private capacity.

\subsection{Temporal-Robust Reconstruction in the Wan VAE}
\label{sec:analysis-vae}

A practical question is whether the Wan VAE's temporal downsampling relies on strong short-range temporal correlations.
We stress-test this assumption with deliberately adversarial inputs: frames sampled far apart in time, and even heterogeneous still images stacked along the temporal axis as if they formed a video.
The VAE still reconstructs these sequences successfully, suggesting that its compression is not heavily biased toward exploiting local temporal redundancy.
This property benefits WALL-WM: even frame-skipped embodied videos need not resemble smooth web-video snippets for the VAE to preserve per-frame visual content.

\subsection{Visual-Semantic and Spatial Strength of the Inherited Wan Features}
\label{sec:analysis-wan-substrate}

This subsection explains why the inherited Wan features sit at the bottom of the conditioning stack as an \emph{immovable substrate}, rather than as just another swappable encoder.
Although Wan is trained with a video-generation objective (instead of a discriminative one), its hidden-state geometry encodes both (i) object/instance-level visual-semantic structure and (ii) 3D, camera-consistent spatial structure. On the spatial axis, it is in practice stronger than canonical contrastive image-pretraining features.

\begin{figure}[!htbp]
    \centering
    \includegraphics[width=0.9\linewidth]{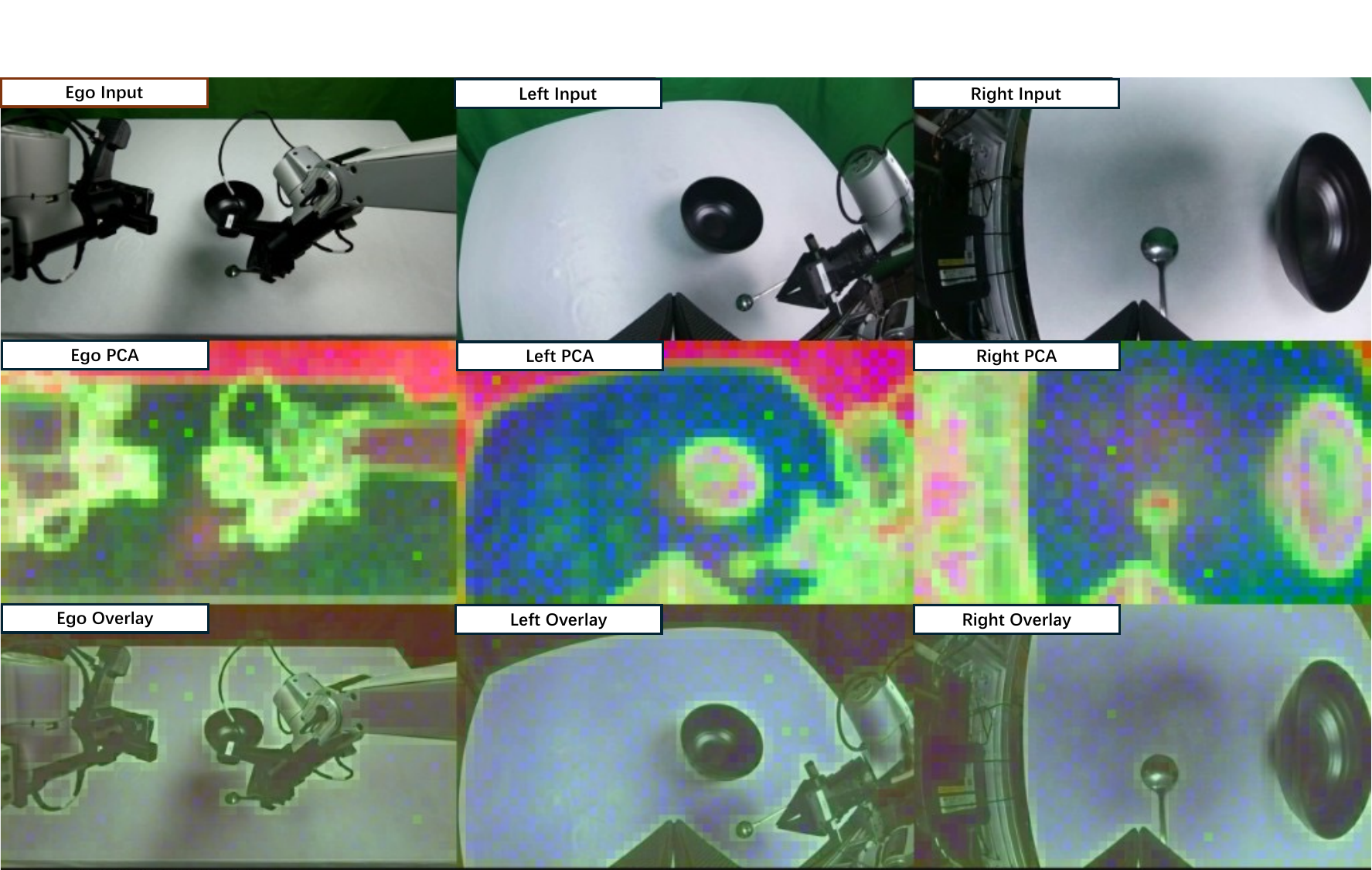}
    \caption{\textbf{Tri-view PCA consistency of Wan generative features.}
    For a synchronized \emph{ego / left-wrist / right-wrist} triplet (top: raw inputs), we extract intermediate Wan2.1 DiT features and project them with a PCA basis fit jointly across the three views (middle); the bottom row overlays the resulting PCA coloring onto the input frames.
    Semantically equivalent surfaces (the manipulated object, the operator's hand, the empty bowl) receive consistent PCA signatures across cameras despite minimal raw-pixel overlap, and per-token cross-view cosine consistency stays at $\sim\!0.97$.
    Read alongside~\cite{wu2026generation,huang2025much}, this provides a concrete visual rationale for treating the inherited Wan features as a visual-semantic \emph{and} spatial substrate, rather than as a generic image prior.}
    \label{fig:wan-pca-triview}
\end{figure}

\paragraph{Cross-view PCA: a direct visual diagnostic.}
We probe this on a synchronized \emph{ego / left-wrist / right-wrist} triplet from the deployment camera rig.
We extract intermediate Wan DiT features at the same denoising step from all three views, project them with a \emph{shared} PCA basis fit jointly across the triplet (\Cref{fig:wan-pca-triview}, middle row), and overlay the resulting PCA coloring onto the original frames (bottom row).
Three patterns emerge.
First, semantically equivalent surfaces---the manipulated object, the operator's hand, and the empty bowl---collapse to a small set of PCA components and receive \emph{consistent} colors across views, despite the three cameras sharing little raw-pixel overlap.
Second, the coloring tracks underlying physical surfaces rather than view-specific image gradients: the same object largely preserves its PCA signature under the camera-pose change between left- and right-wrist.
Third, per-token cross-view cosine consistency stays at $\sim\!0.97$ for every pair of views (see the meta-data strip in \Cref{fig:wan-pca-triview}). Under the same protocol, ViT-SigLIP and self-supervised image features extracted from the same triplet yield reproducibly lower consistency; the gap concentrates on geometry-sensitive tokens (object edges, contact regions) rather than background.

\subsection{Can KV-Cache Streaming Solve the Limits of V-A Temporal Alignment?}
\label{sec:analysis-va-alignment}
\label{sec:analysis-ar-vs-bidir}
\label{sec:analysis-episode-latent-misalign}

A recurring difficulty in vision--action world models is \emph{temporal alignment}. Language typically specifies a semantic intent whose temporal support is underspecified, while video and action unfold over many frames (\Cref{fig:vla_alignment}, \Cref{sec:method-overview}).
Recent work---including DreamZero~\cite{ye2026world} and LingBot-VA~\cite{li2026causal}---adopts KV-cache--compatible rollouts to stream visual/action context across chunks. This helps retain historical state and can partially reduce V--A alignment drift during long-horizon prediction.

However, the fix remains \emph{partial}. These systems still train and infer with \emph{fixed-horizon} future prediction: the model generates a predetermined number of frames after the current anchor, rather than stopping at the instruction's semantic endpoint.

\subsection{Zero-Shot Evaluation on RoboTwin}
\label{app:robotwin_zero_shot}

\paragraph{Evaluation protocol.}
To complement the real-robot experiments, we evaluate the zero-shot
manipulation capability of pretrained WALL-WM on RoboTwin.
The pretrained model is deployed without task-specific fine-tuning
on the benchmark.
The evaluation comprises 50 manipulation tasks, with 10 episodes
per task and 500 episodes in total.
We measure performance using binary task success, which assesses
complete task execution and complements the dense \emph{Task Progress}
metric used in our real-robot evaluation.

For task $i$, let $s_i$ denote the number of successful episodes
out of $n_i = 10$ trials. We report the task-level success rate
$\mathrm{SR}_i = s_i / n_i$ and the macro-averaged success rate
\begin{equation}
    \mathrm{SR}_{\mathrm{avg}}
    =
    \frac{1}{T}
    \sum_{i=1}^{T}
    \frac{s_i}{n_i},
    \qquad T = 50.
    \label{eq:robotwin_success_rate}
\end{equation}
Because every task is evaluated with the same number of episodes,
the macro average is identical to the pooled success rate over
all 500 episodes. All success rates are reported as percentages.

\paragraph{Quantitative results.}
Table~\ref{tab:robotwin_zero_shot} presents the complete task-level
results. WALL-WM achieves an average success rate of
\textbf{15.2\%}, corresponding to \textbf{76 successful episodes
out of 500}, and records at least one successful episode on
\textbf{26 of the 50 tasks}.
Performance varies substantially across tasks.
The highest success rates are obtained on
\texttt{click\_bell} (90\%),
\texttt{press\_stapler} (80\%), and
\texttt{click\_alarmclock} (60\%).
Successful execution also extends to object placement and
manipulation:
\texttt{place\_container\_plate},
\texttt{place\_empty\_cup}, and
\texttt{shake\_bottle} each achieve 50\% success,
while \texttt{grab\_roller},
\texttt{move\_playingcard\_away}, and
\texttt{shake\_bottle\_horizontally} each achieve 40\%.
These observations provide evidence that the pretrained model
retains executable manipulation capabilities when deployed
in simulation without task-specific fine-tuning.
\par
Overall, these results provide initial evidence that pretrained
WALL-WM can execute a range of manipulation tasks in simulation
without task-specific fine-tuning, although performance varies
substantially across tasks.

\begin{table}[tbp]
    \centering
    \small
    \setlength{\tabcolsep}{5pt}
    \renewcommand{\arraystretch}{1.08}
    \begin{tabular*}{\textwidth}{
        @{\extracolsep{\fill}}lrlr@{}
    }
        \toprule
        Task & SR (\%) & Task & SR (\%) \\
        \midrule
        \texttt{adjust\_bottle}
            & 10
            & \texttt{place\_can\_basket}
            & 0 \\
        \texttt{beat\_block\_hammer}
            & 20
            & \texttt{place\_cans\_plasticbox}
            & 0 \\
        \texttt{blocks\_ranking\_rgb}
            & 0
            & \texttt{place\_container\_plate}
            & 50 \\
        \texttt{blocks\_ranking\_size}
            & 0
            & \texttt{place\_dual\_shoes}
            & 0 \\
        \texttt{click\_alarmclock}
            & 60
            & \texttt{place\_empty\_cup}
            & 50 \\
        \texttt{click\_bell}
            & 90
            & \texttt{place\_fan}
            & 0 \\
        \texttt{dump\_bin\_bigbin}
            & 20
            & \texttt{place\_mouse\_pad}
            & 10 \\
        \texttt{grab\_roller}
            & 40
            & \texttt{place\_object\_basket}
            & 0 \\
        \texttt{handover\_block}
            & 0
            & \texttt{place\_object\_scale}
            & 10 \\
        \texttt{handover\_mic}
            & 0
            & \texttt{place\_object\_stand}
            & 20 \\
        \texttt{hanging\_mug}
            & 0
            & \texttt{place\_phone\_stand}
            & 10 \\
        \texttt{lift\_pot}
            & 10
            & \texttt{place\_shoe}
            & 30 \\
        \texttt{move\_can\_pot}
            & 0
            & \texttt{press\_stapler}
            & 80 \\
        \texttt{move\_pillbottle\_pad}
            & 0
            & \texttt{put\_bottles\_dustbin}
            & 0 \\
        \texttt{move\_playingcard\_away}
            & 40
            & \texttt{put\_object\_cabinet}
            & 0 \\
        \texttt{move\_stapler\_pad}
            & 0
            & \texttt{rotate\_qrcode}
            & 0 \\
        \texttt{open\_laptop}
            & 30
            & \texttt{scan\_object}
            & 0 \\
        \texttt{open\_microwave}
            & 0
            & \texttt{shake\_bottle}
            & 50 \\
        \texttt{pick\_diverse\_bottles}
            & 10
            & \texttt{shake\_bottle\_horizontally}
            & 40 \\
        \texttt{pick\_dual\_bottles}
            & 0
            & \texttt{stack\_blocks\_three}
            & 0 \\
        \texttt{place\_a2b\_left}
            & 10
            & \texttt{stack\_blocks\_two}
            & 0 \\
        \texttt{place\_a2b\_right}
            & 20
            & \texttt{stack\_bowls\_three}
            & 0 \\
        \texttt{place\_bread\_basket}
            & 0
            & \texttt{stack\_bowls\_two}
            & 10 \\
        \texttt{place\_bread\_skillet}
            & 0
            & \texttt{stamp\_seal}
            & 10 \\
        \texttt{place\_burger\_fries}
            & 10
            & \texttt{turn\_switch}
            & 20 \\
        \midrule
        \multicolumn{3}{l}{
            \textbf{Average across all 50 tasks}
        } & \textbf{15.2} \\
        \bottomrule
    \end{tabular*}
    \caption{
        \textbf{Zero-shot evaluation of pretrained WALL-WM
        on RoboTwin.}
        Task success rates (SR) are reported in percentages,
        with 10 evaluation episodes per task and no task-specific
        fine-tuning.
        The average includes all 50 tasks and corresponds to
        76 successful episodes out of 500.
        A zero entry denotes no successful episodes among the
        10 evaluated trials.
    }
    \label{tab:robotwin_zero_shot}
\end{table}

\subsection{Qualitative Overview: Pre-training Data and Deployed Execution}
\label{sec:supp-qualitative}

\begin{figure*}[t]
    \centering
    \includegraphics[width=\linewidth]{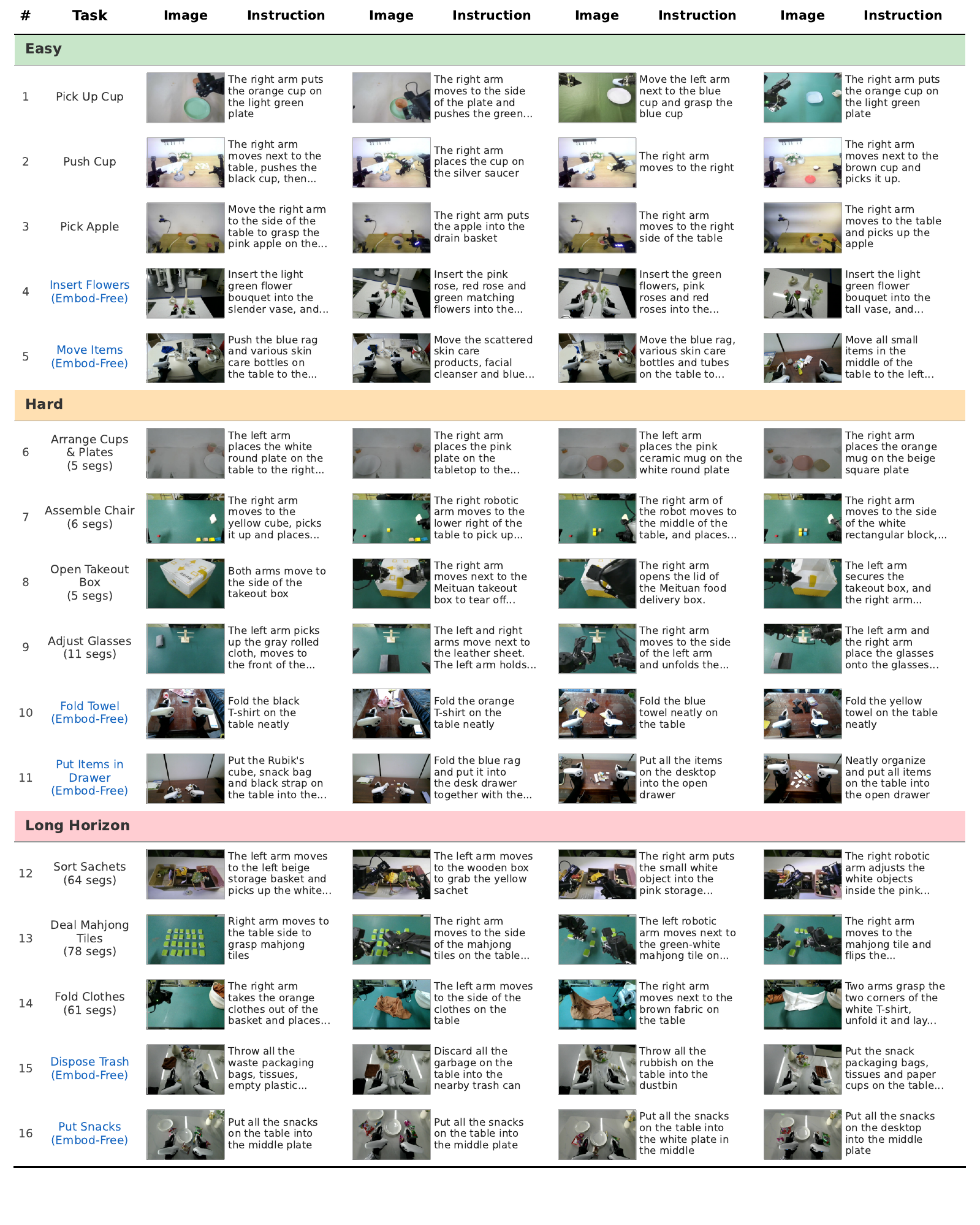}
    \caption{
        \textbf{Pre-training Tasks Overview.}
        Representative pre-training tasks with four scene or event variations.
        Rows are grouped by difficulty, and task names in
        {\color{efblue}\textbf{blue}} indicate embodiment-free examples.
        Instructions are taken from human segment-level captions.
    }
    \label{fig:pretrain-overview}
\end{figure*}

\begin{figure*}[t]
    \centering
    \includegraphics[width=\linewidth]{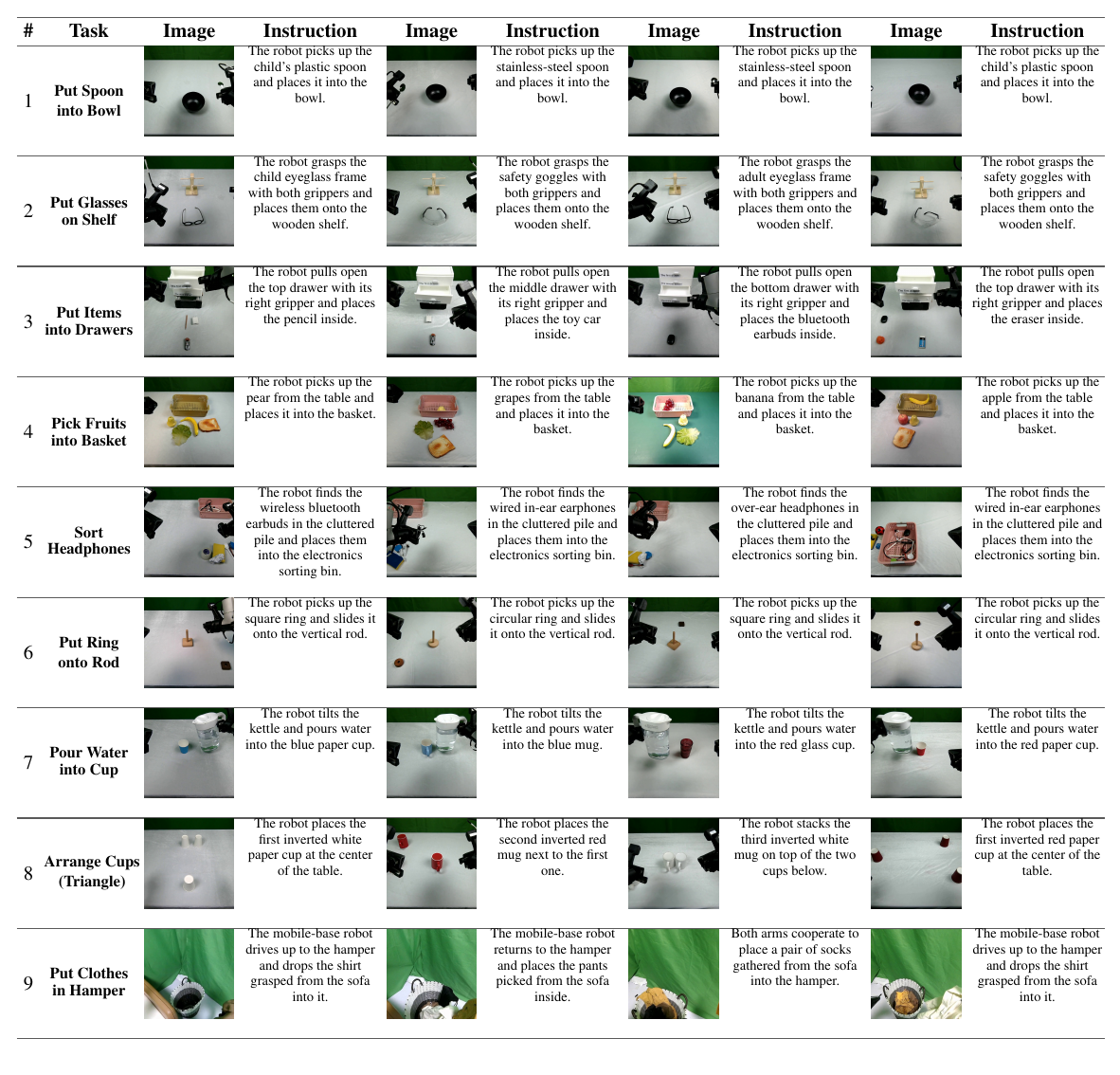}
    \caption{
        \textbf{Real-Robot Evaluation Setup.}
        Representative real-robot evaluation scenes.
        Each row shows one task with four scene variations and the
        corresponding English sub-instructions.
    }
    \label{fig:diversity-setup}
\end{figure*}

\begin{figure*}[t]
    \centering
    \includegraphics[width=\linewidth]{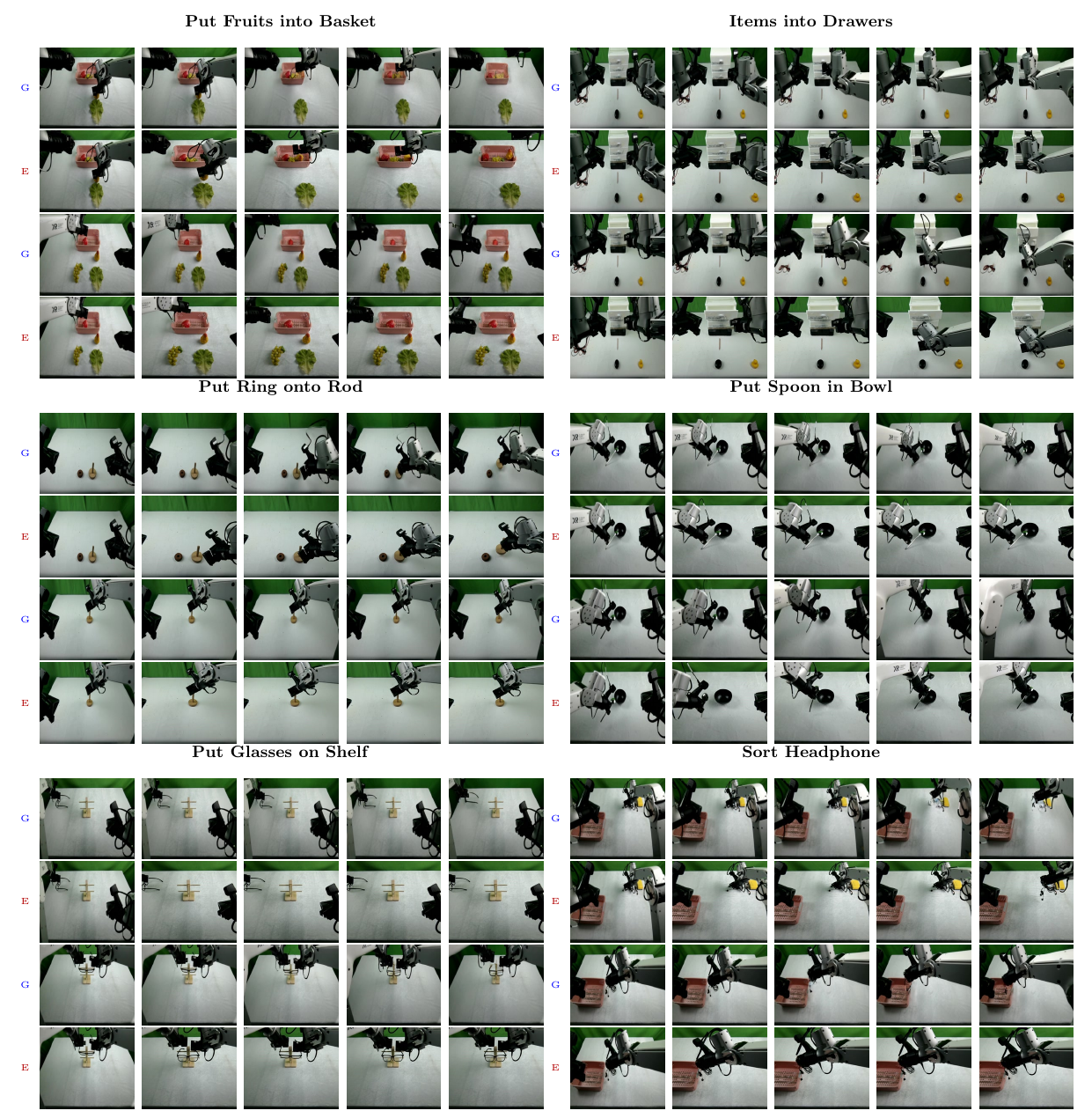}
    \caption{
        \textbf{Generated Video vs.\ Real Execution.}
        Qualitative comparison between generated future video and real-robot
        execution. \textcolor{blue}{G} denotes generated frames and
        \textcolor{red!70!black}{E} denotes execution frames.
    }
    \label{fig:gen-vs-real}
\end{figure*}

This section provides qualitative visual examples for the pre-training data, real-robot evaluation scenes, and deployment-time prediction. Figure~\ref{fig:pretrain-overview} shows representative pre-training tasks and segment-level captions; Figure~\ref{fig:diversity-setup} shows the scene diversity used in real-robot evaluation; and Figure~\ref{fig:gen-vs-real} compares generated future video with the corresponding real execution.

\begin{table}[h!]
    \centering
    \scriptsize
    \setlength{\tabcolsep}{2.2pt}
    \renewcommand{\arraystretch}{1.06}
    \resizebox{\linewidth}{!}{
    \begin{tabular}{lccccc}
        \toprule
        Task & $\pi_{0.5}$ & LingBot-VA & DreamZero & \textsc{WALL-WM}-U-Scratch & \textsc{WALL-WM}-E \\
        \midrule
        \multicolumn{6}{l}{\textbf{Diverse Manipulation}} \\
        \texttt{Arrange Cup Inverted Triangle} & 47 & 20 & 25 & 48 & 85 \\
        \texttt{Put Spoon to Bowl} & 32 & 24 & 54 & 82 & 86 \\
        \texttt{Put Glasses on Woodshelf} & 67 & 18 & 37 & 63 & 64 \\
        \texttt{Put Ring onto Rod} & 74 & 24 & 27 & 77 & 82 \\
        \texttt{Put Blocks to Color} & 83 & 34 & 27 & 90 & 92 \\
        \texttt{Pour Water from Bottle} & 19 & 12 & 12 & 21 & 34 \\
        \texttt{Pick Items into Basket} & 67.5 & 76 & 97.8 & 60 & 88 \\
        \textbf{Average} & \textbf{55.64} & \textbf{29.71} & \textbf{39.97} & \textbf{63.00} & \textbf{75.86} \\
        \midrule
        \multicolumn{6}{l}{\textbf{Reasoning Manipulation}} \\
        \texttt{Sort Headphone} & 41 & 56 & 66 & 84 & 84 \\
        \texttt{Classify Items as Shape} & 55 & 30 & 36 & 82 & 78 \\
        \texttt{Press Button in Order} & 31 & 8 & 3 & 18 & 64 \\
        \texttt{Pair Up Items} & 77 & 4 & 13.5 & 43.5 & 36 \\
        \texttt{Pick Fruits into Basket} & 78 & 60 & 45 & 70 & 96 \\
        \textbf{Average} & \textbf{56.40} & \textbf{31.60} & \textbf{32.70} & \textbf{59.50} & \textbf{71.60} \\
        \midrule
        \multicolumn{6}{l}{\textbf{Dexterous Manipulation}} \\
        \texttt{Insert Wireline} & 18 & 28 & 24 & 42 & 42 \\
        \texttt{Put Stationery in Case} & 12 & 20 & 26 & 20.5 & 22 \\
        \textbf{Average} & \textbf{15.00} & \textbf{24.00} & \textbf{25.00} & \textbf{31.25} & \textbf{32.00} \\
        \midrule
        \multicolumn{6}{l}{\textbf{Generalization}} \\
        \texttt{Place Plates into Storage Slots} & 26 & -- & 48 & 4 & 64 \\
        \texttt{Cover Pot with Lid} & 14 & -- & 24 & 32 & 26 \\
        \texttt{Push Cleaning Cloth to Table Edge} & 32 & -- & 18 & 12 & 78 \\
        \texttt{Insert Screwdriver into Cup} & 24 & -- & 24 & 26 & 47 \\
        \textbf{Average} & \textbf{24.00} & \textbf{--} & \textbf{28.50} & \textbf{18.50} & \textbf{53.75} \\
        \bottomrule
    \end{tabular}
    }
    \caption{\textbf{Detailed real-robot Task Progress scores.}
    Task-level Task Progress is reported for all real-robot evaluation suites.
    \textsc{WALL-WM}-U-Scratch and \textsc{WALL-WM}-E denote unified mode from scratch and event mode, respectively.
    A dash indicates that the corresponding baseline was not evaluated.}
    \label{tab:real-robot-detailed-scores}
\end{table}

\begin{table*}[h!]
    \centering
    \scriptsize
    \setlength{\tabcolsep}{2pt}
    \renewcommand{\arraystretch}{1.08}
    \begin{tabular}{p{0.22\textwidth}p{0.36\textwidth}p{0.36\textwidth}}
        \toprule
        Task & Description & Scoring Rule \\
        \midrule
        \multicolumn{3}{l}{\textbf{Diverse Manipulation}} \\
        \texttt{Arrange Cup Inverted Triangle} & Invert 3 cups in a triangle: 2 bottom, 1 top & Per cup: pick up (1) + position (1--3); top (3), retract (1) \\
        \texttt{Put Spoon to Bowl} & Move bowl to center, place spoon in bowl & Bowl (2), move (2), spoon (2), place in bowl (3), retract (1) \\
        \texttt{Put Glasses on Woodshelf} & Pick up glasses, adjust, place on rack & Pick up (2), center (2), adjust (2), rack (3), retract (1) \\
        \texttt{Put Ring onto Rod} & Place a ring object onto a vertical pole & Grasp ring (3), move to pole (3), stack (3), retract (1) \\
        \texttt{Put Blocks to Color} & Place RGB blocks onto matching color patches & 3 pts per correct, 3 blocks, retract (1) \\
        \texttt{Pour Water from Bottle} & Pour water from bottle into cup & Move cup (2), bottle (2), pour (3), set down (2), retract (1) \\
        \texttt{Pick Items into Basket} & Pick up 4 objects and place into basket & Per object: pick up (1) + place (1.25); 4 objects, retract (1) \\
        \midrule
        \multicolumn{3}{l}{\textbf{Reasoning Manipulation}} \\
        \texttt{Sort Headphone} & Find earphones in clutter and place them in the basket & Identify (2), pick up (3), place (4), retract (1) \\
        \texttt{Classify Items as Shape} & Sort 3 objects by shape to target positions & 3 pts per correct, 3 objects, retract (1) \\
        \texttt{Press Button in Order} & Press 3 buttons in the instructed order & 3 pts per correct press, 3 buttons, retract (1) \\
        \texttt{Pair Up Items} & Match object pairs and place them at target positions & Per pair: grasp (2) + place (2.5); 2 pairs, retract (1) \\
        \texttt{Pick Fruits into Basket} & Place specified fruits into basket & 3 pts per correct fruit, 3 fruits, retract (1) \\
        \midrule
        \multicolumn{3}{l}{\textbf{Dexterous Manipulation}} \\
        \texttt{Insert Wireline} & Bimanual: left picks cable, passes to right, inserts & Left grasp (3), right receive (3), insert (3), retract (1) \\
        \texttt{Put Stationery in Case} & Bimanual: open zipper, insert items, close & Open (3.5), 1 pt per item (3), zip (1.5), retract (1) \\
        \midrule
        \multicolumn{3}{l}{\textbf{Generalization}} \\
        \texttt{Place Plates into Storage Slots} & Pick up plates and place them into the instructed storage slots & Pick up (2), align with slot (3), place into slot (4), retract (1) \\
        \texttt{Cover Pot with Lid} & Pick up pot lid and place on pot & Approach (2), pick up lid (3), cover (4), return (1) \\
        \texttt{Push Cleaning Cloth to Table Edge} & Pick up cloth and wipe or push grease stains to the table edge & Approach (2), pick up (2), wipe all (5), return (1) \\
        \texttt{Insert Screwdriver into Cup} & Pick up the screwdriver and insert it into the cup & Identify (2), pick up (3), insert into cup (4), retract (1) \\
        \bottomrule
    \end{tabular}
    \caption{\textbf{Real-robot Task Progress rubrics.}
    Each task is scored on a 10-point Task Progress scale with partial credit assigned to observable intermediate steps.
    The final reported score is normalized to the 0--100 Task Progress values used in the main evaluation.}
    \label{tab:scoring-rubrics}
\end{table*}

\clearpage
\bibliographystyle{assets/plainnat}
\bibliography{main}

\clearpage
\appendix

\end{document}